\documentclass[lettersize,journal]{IEEEtran}
\usepackage{enumitem}

\usepackage[utf8]{inputenc}
\usepackage{arxiv}
\usepackage{cite,soul}
\usepackage{epsfig}
\usepackage{amsthm,amsmath,amssymb,amsfonts}
\usepackage{pifont,times}
\usepackage{algorithmic}
\usepackage{graphicx,graphics}
\usepackage[hidelinks]{hyperref}
\usepackage{dsfont}
\usepackage{colortbl,color}
\usepackage{subcaption}
\usepackage{makecell}
\usepackage{mathtools}

\usepackage{multirow}      % for \multirow

\usepackage[linesnumbered,ruled]{algorithm2e}

\newtheorem{theorem}{Theorem}[section]

\newtheorem{lemma}[theorem]{Lemma}

\newtheorem{remark}[theorem]{Remark}

\newcommand{\ones}{\mathds{1}}
\newcommand{\tr}{\top}
\newcommand{\Rm}{R_\textnormal{max}}
\newcommand{\cA}{\mathcal{A}}

\newcommand{\bE}{\mathbb{E}}

\newcommand{\cF}{\mathcal{F}}

\newcommand{\cM}{\mathcal{M}}

\newcommand{\bP}{\mathbb{P}}
\newcommand{\bR}{\mathbb{R}}
\newcommand{\cP}{\mathcal{P}}
\newcommand{\cR}{\mathcal{R}}
\newcommand{\cS}{\mathcal{S}}
\newcommand{\taut}{\tau_{t}}
\newcommand{\tauT}{\tau_{T}}
\newcommand{\tauk}{\tau_{k}}

\newcommand{\Wt}{W_{t+1}}

\newcommand{\bt}{\beta_{t}}
\newcommand{\bk}{\beta_{k}}
\newcommand{\bT}{\beta_{T}}
\newcommand{\Delbar}{\bar{\Delta}}

\newcommand{\gt}{\gamma_t}
\newcommand{\Cealp}{C_{E,\alpha}}
\newcommand{\CG}{C_G}

\newcommand{\KG}{K_G}
\newcommand{\Cbeta}{C_\beta}

\newcommand{\ts}{\tilde{s}}
\newcommand{\ta}{\tilde{a}}
\newcommand{\Omone}{\Omega^{(1)}}
\newcommand{\Omtwo}{\Omega^{(2)}}

\newcommand{\df}[1]{\textnormal{d}{#1}}
\newcommand{\thS}{\theta^*}
\newcommand{\rS}{r^*}

\newcommand{\Aht}{\hat{A}_t}
\newcommand{\vht}{\hat{v}_t}
\newcommand{\vhk}{\hat{v}_k}
\newcommand{\zht}{\hat{z}_t}
\newcommand{\zhk}{\hat{z}_k}
\newcommand{\tz}{\tilde{z}}
\newcommand{\Atil}{\tilde{A}_t}

\newcommand{\lin}{\textnormal{lin}}

\newcommand{\qd}{\textnormal{quad}}

\newcommand{\Cm}{C_{\textnormal{M}}}

\newcommand{\diag}{\textnormal{diag}}
\newcommand{\tstr}{t_{*}}
\newcommand{\Grho}{G^{\rho}}

\newcommand{\Cerg}{C_E}
\newcommand{\CD}{C_\Delta}
\newcommand{\CS}{C_S}

\newcommand{\Clambda}{C_\lambda}
\newcommand{\xfl}{\xi_{FL}}

\newcommand{\lmin}{\lambda_\textnormal{min}}

\newcommand{\Gdel}{G^{\Delta}}

\newcommand{\ep}{\varepsilon_p}
\newcommand{\er}{\varepsilon_r}

\newcommand{\cmark}{\checkmark} % Checkmark symbol
\newcommand{\xmark}{\ding{55}}

\DeclarePairedDelimiter\floor{\lfloor}{\rfloor}

\renewcommand{\rm}{r^\mu}

\allowdisplaybreaks        % in the preamble

\hypersetup{hidelinks=true}
\usepackage{textcomp}
\def\BibTeX{{\rm B\kern-.05em{\sc i\kern-.025em b}\kern-.08em
    T\kern-.1667em\lower.7ex\hbox{E}\kern-.125emX}}
\begin{document}

\title{Parameter-Free Federated TD Learning with Markov Noise in Heterogeneous Environments}

\author{Ankur Naskar, Gugan Thoppe, Utsav Negi, and Vijay Gupta
\thanks{An earlier version of this work, involving IID sampling, appeared in \cite{naskarCDC}. Here, we study the more realistic and challenging setup of Markovian sampling. The work was partially supported by Purdue University and DST-SERB's Overseas Visiting Doctoral Fellowship, ARO under grant W911NF2310266, and ONR under grant N000142312604. It was also supported by DST-SERB's Core Research Grant CRG/2021/008330, the Indo-French Centre for the Promotion of Advanced Research---CEFIPRA (7102-1), the Walmart Centre for Tech Excellence. A.~Naskar ({\tt\small ankurnaskar@iisc.ac.in}) and G.~Thoppe ({\tt\small gthoppe@iisc.ac.in}) are with the Dept.~of Computer Science and Automation, Indian Institute of Science, Bengaluru, India. U.~Negi ({\tt\small unegi@purdue.edu}) and V.~Gupta ({\tt\small gupta869@purdue.edu}) are with the School of Electrical and Computer Engineering, Purdue University, IN, USA}%
}

\maketitle

\begin{abstract}
    Federated learning (FL) can dramatically speed up reinforcement learning by distributing exploration and training across multiple agents. It can guarantee an optimal convergence rate that scales linearly in the number of agents, i.e., a rate  of $\tilde{O}(1/(NT)),$ where  $T$ is the iteration index and $N$ is the number of agents. However, when the training samples arise from a Markov chain, existing results on TD learning achieving this rate require the algorithm to depend on unknown problem parameters. We close this gap by proposing a two-timescale Federated Temporal Difference (FTD) learning with Polyak-Ruppert averaging. Our method provably attains the optimal $\tilde{O}(1/NT)$ rate
    in both average-reward and discounted settings---offering a parameter-free FTD approach for Markovian data. Although our results are novel even in the single-agent setting, they apply to the more realistic and challenging scenario of FL with heterogeneous environments. 
\end{abstract}

\begin{IEEEkeywords}
Federated learning, Markov processes, Reinforcement learning, Learning systems
\end{IEEEkeywords}

\IEEEpeerreviewmaketitle

\section{Introduction}
\label{sec:introduction}

Federated Learning (FL) allows multiple devices or servers to collaboratively train a machine learning model without needing to transmit their local data to a central location, thus alleviating bandwidth, energy, and privacy concerns. Much work has, thus, been done to extend FL in many directions~\cite{mcmahan2017communication,bonawitz2019towards}. We are interested in the work on Federated Reinforcement Learning (FRL)~\cite{qi2021federated,nadiger2019federated,zhuo2019federated,wang2020optimizing}. In Reinforcement Learning (RL), an agent needs to learn a strategy or a policy for sequentially manipulating the state of a system, typically modeled as a Markov Decision Process (MDP), in a way that optimizes a certain cumulative reward function \cite{sutton2018reinforcement,levine2020offline,recht2019tour,cheng2021general, kyriakis2019specification}. %An MDP defines how the probability distribution of the system's future state depends on its current state and the action performed by the agent, as well as what instantaneous reward the agent would receive at each state transition. The MDP also specifies the rule, such as average-reward or exponential discounting, to get the cumulative sum of the rewards obtained at different transitions.
FRL is a natural means to confer the advantages of the FL paradigm to RL using the same cyclic three-step process as FL. First, the edge devices train the local RL model. Next, these devices transfer the trained models to a central server, which aggregates them. Finally, the server transmits this global model to the edge devices that use it for subsequent training. With FRL, the different devices can coordinate to jointly explore the vast state and action spaces, potentially leading to a linear speedup with respect to the number of participating devices. Initial works show that this intuition is true, at least when each edge device has access to the same system model \cite{khodadadian2022federated,liu2023distributed,shen2023towards, dal2023federated}.

\begin{table*}[ht]
\centering
\begin{tabular}{!{\vrule width0.6pt}c!{\vrule width0.6pt}p{16.75em}|c|c|c|c|c|c|c!{\vrule width0.6pt}}
\Xhline{0.6pt}
& \textbf{Reference} & \textbf{Federated} & \textbf{Heterogeneous}   &  \textbf{Discounting} &  \textbf{ Asynchronous} & \textbf{\makecell{Markov \\ Noise}} & \textbf{\makecell{Optimal \\ rate}}  & \textbf{\makecell{Universal \\ stepsize}} 
\\
\Xhline{0.6pt}
\multirow{14}{0.4cm}
{
      \rotatebox[origin=c]{90}{\parbox{2.5cm}{\centering TD Learning}}
}
% \hline
%
& \cite{dalal2018td0}: Dalal et al., 2018  & \xmark & - & Exp & \cmark & \xmark & \xmark & \cmark 
\\
\cline{2-9}
&  \cite{lakshminarayanan2018linear}: Lakshminarayanan et al., 2018 & \xmark & - & Exp and Avg & \cmark & \xmark & \cmark &\xmark
\\
\cline{2-9}
&  \cite{bhandari2018finite, bhandari2021finite}: Bhandari et al., 2018, 2021  & \xmark & - & Exp & \cmark & \cmark & \cmark & \xmark  
 \\ 
\cline{2-9}
&  \cite{chen2021finite}: Chen et al., 2021 & \xmark & - & Exp & \cmark & \cmark & \cmark & \xmark 
\\
\cline{2-9}
&  \cite{patil2023finite}: Patil et al., 2023 & \xmark & - & Exp & \cmark & \xmark & \cmark & \cmark
\\
\cline{2-9}
&  \cite{patil2023finite}: Patil et al., 2023 & \xmark & - & Exp & \cmark & \cmark & \cmark & \xmark
\\
\cline{2-9}
& \cite{chen2024lyapunov}: Chen et al., 2024 & \xmark & -  & Exp & \cmark & \cmark & \cmark & \xmark
\\
\cline{2-9}
& \cite{chen2025concentration}: Chen et al., 2025 & \xmark & - & Exp & \cmark  & \xmark & \cmark & \xmark
\\
\cline{2-9}
& \cite{haquestochastic}: Haque and Maguluri, 2025 & \xmark & - & Avg & \cmark  & \cmark & \cmark & \xmark
\\
\cline{2-9}
& \cite{chen2025non}: Chen et al., 2025 & \xmark & - & Exp and Avg & \cmark & \cmark & \cmark & \xmark
\\
\cline{2-9}
&  \cite{liu2023distributed}: Liu et al., 2023 & \cmark & \xmark & Exp & \cmark & \xmark & \cmark & \xmark
\\
\cline{2-9}
&  \cite{dal2023federated}: Dal Fabbro et al., 2023  & \cmark & \xmark &  Exp & \cmark & \cmark & \cmark & \xmark
\\
\cline{2-9}
& \cite{khodadadian22a}: Khodadadian et al., 2022 & \cmark & \xmark & Exp &  \cmark & \cmark & \cmark & \xmark 
\\
\cline{2-9}

& \cite{wang2024federated}: Wang et al., 2024 & \cmark & \cmark & Exp & \cmark & \cmark & \cmark & \xmark
\\
\cline{2-9}
& \cite{naskarCDC}: Naskar et al., 2024 & \cmark & \cmark & Exp and Avg & \cmark  & \xmark & \cmark & \cmark
\\
\cline{2-9}
\rowcolor[gray]{0.75}
& \textbf{Our work} & \cmark & \cmark & Exp and Avg & \cmark & \cmark & \cmark & \cmark \\
\Xhline{0.6pt}
\multirow{6}{0.4cm}
{
      \rotatebox[origin=c]{90}{\parbox{2cm}{\centering $Q$-Learning}}
}
& \cite{even2003learning}: Even-Dar and Mansour, 2003 & \xmark  & - & Exp & \xmark & - &  \xmark & \cmark 
\\
\cline{2-9}
&  \cite{wainwright2019stochastic}: Wainwright, 2019 & \xmark & - & Exp & \xmark & - & \cmark & \xmark
\\
\cline{2-9}
&  \cite{qu2020finite}: Qu and Wierman, 2020 & \xmark & - & Exp & \cmark & \cmark & \cmark & \xmark
\\
\cline{2-9}
& \cite{zhang2021finite}: Zhang et al., 2021 & \xmark & - & Avg & \xmark & - & \cmark & \xmark
\\
\cline{2-9}
& \cite{li2023statistical}:  Li et al., 2023 & \xmark & - & Exp & \xmark & - & \cmark & \cmark
\\
\cline{2-9}
& \cite{chen2024lyapunov}: Chen et al., 2024 & \xmark & -  & Exp & \cmark & \cmark & \cmark & \xmark
\\
\cline{2-9}
& \cite{chen2025concentration}: Chen et al., 2025 & \xmark & - & Exp & \cmark  & \xmark & \cmark & \xmark
\\
\cline{2-9}
& \cite{haquestochastic}: Haque and Maguluri, 2025 & \xmark & - & Exp & \cmark  & \cmark & \cmark & \xmark
\\
\cline{2-9}
& \cite{chen2025non}: Chen et al., 2025 & \xmark & - & Exp and Avg & \cmark & \cmark & \cmark & \xmark
\\
\cline{2-9}
& \cite{chandak2025finite}: Chandak et al., 2025 & \xmark & - & Exp and Avg & \cmark & \cmark & \cmark & \xmark
\\
\cline{2-9}
& \cite{naskar2025parameter}: Naskar et al., 2025 & \cmark & \cmark & Exp & \cmark & \cmark & \cmark & \cmark
\\
\cline{2-9}
& \cite{naskar2025parameter}: Naskar et al., 2025 & \cmark & \cmark & Avg & \xmark & - & \cmark & \cmark 
\\
\Xhline{0.6pt}
\end{tabular}
\caption{\label{tab:lit.survey} Comparison of our work with the existing literature on TD-learning and $Q$-learning algorithms. In the column labeled discounting, Exp refers to exponential, while Avg refers to average reward. [24], [30]: High-probability bounds}
\end{table*}

In practice, the systems that the edge devices interact with are rarely homogeneous. For instance, when designing a controller for an autonomous car using data from multiple cars, each car may have a different environment and configuration. %Similarly, even if multiple robots are operated under similar conditions, their physical model parameters need not be identical.  
Indeed, much of the FL literature is devoted to taming such heterogeneity. In FRL, this problem is even more acute since if the MDPs at the edge devices are different, %the controller learned by the server may not be optimal, or even stabilizing, for any of them! It 
it is not clear a priori whether the data collected by multiple heterogeneous edge devices can be aggregated to find a `universal' controller that performs well across all the edge models. Even if this were possible, one could ask whether the speedup from the homogeneous model case can be achieved in the heterogeneous case to find this universal controller. 

Recent works such as \cite{wang2024federated} and \cite{zhang2024finite}, which analyze federated TD and federated SARSA under \textit{exponential discounting} demonstrate that optimal convergence rates with linear speedup are achievable even in heterogeneous settings. A key limitation of \cite{wang2024federated} and \cite{zhang2024finite}, however, is that their rates rely on stepsizes depending on unknown problem-specific quantities---specifically, the minimum eigenvalues of matrices determined by the unknown MDP transition probabilities. 

To address this issue, Polyak–Ruppert (PR) averaging \cite{polyak1992acceleration,ruppert1991stochastic} has emerged as an effective approach in both single-agent and federated settings. The key idea is to run the algorithm with a universal stepsize while maintaining a running average of the iterates, and then show that this average achieves the optimal convergence rate. For instance, in single-agent TD learning with exponential discounting and average rewards, \cite{patil2023finite} and \cite{naskarCDC}, respectively, establish that PR averaging yields the optimal rate without requiring problem-specific stepsizes. \cite{naskar2025parameter} shows the same for federated Q-learning under both exponential discounting and average-reward setups.

However, the analyses in \cite{patil2023finite} and \cite{naskarCDC} assume that the training data---comprising state, action, and reward samples---is generated in an Independent and Identically Distributed (IID) fashion. For the more realistic setting of Markovian data, \cite{patil2023finite} proposes subsampling the trajectory every $\tau$ steps---where $\tau$ is dictated by the (unknown) mixing time of the chain (see their Section 6)---which renders their approach impractical. While \cite{naskar2025parameter} avoids this limitation in the exponentially-discounted case, its results for average-reward Q-learning apply only in the synchronous setting. That is, in each iteration, the analysis assumes access to the next state and reward samples for every state–action pair, which is again impractical.

In summary, the key question of whether PR averaging can yield parameter-free optimal rates in asynchronous average-reward RL with Markovian single-trajectory data remains open, even in the single-agent setting. In the federated case, an additional open problem is whether such rates also translate into a linear speedup with the number of agents. The main difficulty in resolving these questions arises from the fact that the average-reward Bellman operator is not a contraction in the standard norm, but only in a semi-norm. 

In this work, we address the above gaps for the TD(0) algorithm for policy evaluation with linear function approximation. For completeness, we also prove an analogous results for the exponentially discounted setting. While the latter result can be inferred from the analysis in \cite{naskar2025parameter}, to the best of our knowledge, it has not been explicitly stated in the literature. 
% We provide an outline of this proof as well.

% Similarly, \cite{durmus2024finite} derives finite-time bounds for PR-averaged iterates of general linear stochastic approximation with Markovian data, but their main result (Theorem 4) also requires stepsizes that depend explicitly on the mixing time.

% Our main idea is to use iterate averaging---a known technique to remove the dependence of step sizes on unknown problem parameters in single-agent or centralized linear stochastic approximation .  In fact, \cite{patil2023finite} used this idea to derive their parameter-free single-agent TD algorithm with exponential discounting for IID training data. In a federated setup with Markovian sampling and heterogeneous MDPs, it is not clear whether a similar idea will work. In this work, we show that iterate averaging does indeed help. 

Our key contributions can be summarized as follows.  

\begin{itemize}
    \item \textbf{Parameter-Free Optimal Rates for Single-agent TD learning}: Using PR averaging, we  obtain the first parameter-free optimal convergence rate of $\tilde{O}(1/T),$ where $T$ is the iteration index, for asynchronous TD(0) with linear function approximation. Our results apply to policy evaluation with Markovian samples in both average-reward and exponentially-discounted settings. 

    \item \textbf{Federated TD-learning with Linear Speedup} Although our results are novel even in the single-agent setting, they extend to the more realistic---and more challenging---scenario of FL with heterogeneous environments. In this case, our main result shows that, up to a heterogeneity gap, the convergence rate is $\tilde{O}(1/(NT))$, where $N$ is the number of agents. Our result thus implies that the sample complexity decreases linearly with $N$.

    \item \textbf{Two-timescale Analysis}: PR-averaging naturally induces a two-timescale behavior: the original iterates evolve on the faster timescale, while their averages evolve on the slower one. In our analysis, we also estimate the average reward on the slower timescale. This contrasts existing work on average-reward TD learning where both value and average-reward estimates share the same timescale. This fact makes our approach of independent interest. 
  
    \item \textbf{Numerical Simulations}: We demonstrate the efficacy of our approach through simulations in synthetic settings. 
\end{itemize}

Table~\ref{tab:lit.survey} provides a comparison of our work to the prior literature on TD and Q-learning. 

\section{Setup and Problem Formulation}
\label{sec:setup}

We consider $N$ agents (also called clients or nodes), where each agent $i$ has access to a Markov Decision Process (MDP) $\cM_{i}:= (\cS, \cA, \cR_{i}, \cP_{i}).$ Here, $\cS$ and $\cA$ are the finite and common state and action spaces, respectively, while  $\cR_{i}:\cS\times \cA\to \bR$ and $\cP_{i}:\cS\times \cA\to \Delta (\cS)$ are the reward and probability transition functions at agent $i \in [N]$, that can potentially vary among the agents. Further, the notation $\Delta(\cS)$ stands for the set of distributions on $\cS$ and $[N] := \{1, \ldots, N\}.$ Throughout, we use $N = 1$ to denote the single-agent setting, while $N > 1$ corresponds to the federated setup.

We presume that we are provided with a stationary policy $\mu: \cS \to \Delta(\cA)$ and a feature matrix $\Phi \in \bR^{|\cS| \times d}$ for some $1 \leq d \ll |\cS|.$ Our goal then is to analyze the convergence rates of  TD algorithm with PR-averaging---under both average-reward and discounted criteria---that leverage all $N$ agents to estimate $\mu$'s value function in $\Phi$'s column space.

Under the average-reward criterion, the value or quality of the policy $\mu$ is measured using two notions: the average reward and the differential value function. For the MDP $\cM_i,$ the average reward $r^\mu_i \in \bR^{|\cS|}$ is given by
\begin{equation}\label{e: avg.reward}
    r^{\mu}_i(s) := \liminf\limits_{T\to\infty}\frac{1}{T} \bE \bigg[\sum_{t = 0}^{T - 1} \cR_i(s_{t}, a_t) \bigg| s_{0} = s \bigg], \quad s \in \cS,
\end{equation}
where the expectation is with respect to the distribution of the Markovian state-action trajectory $s_0, a_0,  \ldots, s_{T - 1}, a_{T - 1}$ with $a_t \sim \mu(\cdot|s_t)$ and $s_{t + 1} \sim \cP_i(\cdot|s_t, a_t).$
On the other hand, the differential value function $V^\mu_i$ is the fixed point of the differential Bellman operator $T^\mu_i$ given by
\begin{equation}
    \label{e:avg.Bellman.operator}
        T^\mu_i V = \cR^\mu_i - r^\mu_i + \cP^\mu_i V,
\end{equation}
where $\cR^\mu_i(s) := \sum_{a \in \cA} \mu(a|s) \cR_i(s, a),$ and $\cP^{\mu}_i(s, s') \equiv \cP^{\mu}_i(s'|s) := \sum_{a \in \cA} \mu(a|s) \cP_i(s'|s, a).$

Under exponential discounting, $\mu$'s value function is
\begin{equation}
    \hat{V}^\mu_i(s) = \bE \left[\sum_{t = 0}^\infty \gamma^t \cR_i(s_t, a_t) \middle|s_0 = s\right],
\end{equation}
where $\bE$ has the same meaning as in \eqref{e: avg.reward} and $\gamma \in [0, 1)$ is the discount factor. Alternatively, $\hat{V}^\mu_i$ is the fixed point of the Bellman operator $\hat{T}_i^\mu: \bR^{|\cS|} \to \bR^{|\cS|}$ given by 
%
%\begin{equation}
$    \hat{T}^\mu_i V = \cR^\mu_i + \gamma \cP^\mu_i V,$
%\end{equation}
%
where $\cR^\mu_i$ and $\cP^\mu_i$ are defined as for \eqref{e:avg.Bellman.operator}. 

We assume the following standard condition (\cite{dalal2018td0,zhang2021finite}).
{
\renewcommand{\theenumi}{$\mathbf{\cA_\arabic{enumi}}$}
\begin{enumerate}   
     \item\label{a: ergodic} \textbf{Ergodicity}: For each $i,$ the Markov chain $(\cS, \cP^\mu_i)$ induced by the policy $\mu$ is irreducible and aperiodic.
\end{enumerate}
}
\noindent For each $i \in [N],$ this assumption guarantees that the Markov chain $(\cS, \cP^\mu_i)$ has a unique and positive stationary distribution $d^\mu_i;$ further, this Markov chain is ergodic and, for each $s \in \cS,$ 
\begin{equation}
    r^\mu_i(s) = r_i^* := (d^\mu_i)^\tr \cR^\mu_i.
\end{equation}

% Our goal is to develop two Federated TD algorithms, one for the average reward setting and one for exponential discounting. For the average reward, the algorithm should produce a vector $\theta \in \bR^d$ and a scalar $r \in \bR$ such that $\Phi \theta$ (resp. $r$) closely approximates $V^\mu_i$ (resp. $r^*_i$) for each  $i$. For exponential discounting, the algorithm should generate a vector $\vartheta \in \bR^d$ such that $\Phi \vartheta$ is close to $\hat{V}^\mu_i$ for each $i$. Importantly, these algorithms should have an optimal convergence rate, and the corresponding sample complexity should decrease linearly with the number $N$ of agents. Additionally, the stepsize choice for achieving these results should be universal, meaning it should be independent of the unknown model parameters.

\section{Main Results}
\label{sec:algorithm}

\begin{algorithm}[t]
\caption{AvgFedTD(0) \label{AvgFedTD}}
\SetKwInOut{Input}{Input}
\SetKw{Initialize}{Initialize}
\SetKwFor{For}{for}{}{end}
\SetKwBlock{Central}{Central server}{}
\SetKwBlock{Local}{Each agent $i\in [N]$ in parallel}{}
\Input{Policy $\mu,$ step-size sequence $(\bt),$ feature vectors $\{\phi(s): s\in \cS\},$ $r_{0}\in \bR,$ $\theta_{0}\in \bR^{d}.$}

%\vspace{1ex}

\Initialize{$\bar{\theta}_{0} = \theta_{0}, r_{0}^{i}=r_{0}, \forall i\in[N].$}

%\vspace{1ex}

\For{each iteration $t= 0,1, \ldots, T-1:$}
   { 
        %\nonl
        \Local
        {
            %\State 
            Sample $a_{t}^{i}\sim\mu(\cdot|s_{t}^{i}),$ and observe $s_{t+1}^{i} \sim \cP_{i}(\cdot|s_{t}^{i}, a_{t}^{i}).$

            %\vspace{1ex}
         
            %\State 
            Compute local TD error 
            $\delta_{t+1}^{i} = (\cR_{i}(s_{t}^{i},a_{t}^{i}) - r_{t})\phi(s_{t}^{i}) + \phi(s_{t}^{i})[\phi(s_{t+1}^{i}) - \phi(s_{t}^{i})]^\tr \theta_{t}.$

            %\vspace{1ex}
            
            %\State 
            Update local average reward estimate  $r_{t+1}^{i} =  r_{t}^{i} + \frac{1}{t+1}[\cR_{i}(s_{t}^{i},a_{t}^{i}) - r_{t}^{i}].$

            %\vspace{1ex}

            %\State 
            Send $(\delta_{t+1}^{i}, r^{i}_{t+1})$ to central server.

        }

        %\nonl
        \Central
        {
            %\State 
            Update global model parameter
            %\nonl
            $\theta_{t+1} = \theta_{t} + \frac{\bt}{N}\sum_{i\in[N]}\delta_{t+1}^{i}.$

            %\vspace{1ex}
            
            %\State 
            Update Polyak-Ruppert average $\bar{\theta}_{t+1}  = \bar{\theta}_{t} +   \frac{1}{t+1}[\theta_{t} - \bar{\theta}_{t}].$ \label{alg:PR}

            %\vspace{1ex}
            
            %\State 
            Update average reward estimate
            $r_{t+1} = \frac{1}{N}\sum_{i\in[N]}r_{t+1}^{i}.$

            %\vspace{1ex}
            
            %\State 
            Send $(\theta_{t+1},r_{t+1})$ to each agent $i\in[N].$
        }
   }
\end{algorithm}
%%%%%%%%%%%%%%%%%%%%%%%%%%%%%%%%%%%%%%%%%%%%%%%%%%%%%%%%%%
\begin{algorithm}[t]
\caption{ExpFedTD(0) \label{FedTD}}
\SetKwInOut{Input}{Input}
\SetKw{Initialize}{Initialize}
\SetKwFor{For}{for}{}{end}
\SetKwBlock{Central}{Central server}{}
\SetKwBlock{Local}{Each agent $i\in [N]$ in parallel}{}
\Input{Policy $\mu,$ step-size sequence $(\bt),$ feature vectors $\{\phi(s): s\in \cS\},$ $\vartheta_0 \in \bR^d.$}

%\vspace{1ex}

\Initialize{$\bar{\vartheta}_{0} = \vartheta_{0}.$}

%\vspace{1ex}

\For{each iteration $t= 0,1, \ldots, T-1:$}
   { 
        \Local
        {
            %\State 
            Sample $a_{t}^{i}\sim\mu(\cdot|s_{t}^{i}),$ and observe $s_{t+1}^{i} \sim \cP_{i}(\cdot|s_{t}^{i}, a_{t}^{i}).$

            %\vspace{1ex}
         
            %\State 
            Compute local TD error 
            $\delta_{t+1}^{i} = \cR_i(s_t^i,a_t^i)\phi(s_t^i) + \phi(s_t^i)(\gamma\phi^{\tr}(s_{t+1}^i) - \phi^{\tr}(s_t^i))\vartheta_{t}.$

            %\vspace{1ex}

            %\State 
            Send $\delta_{t+1}^{i}$ to central server.

        }

        %\nonl
        \Central
        {
            %\State 
            Update global model parameter
            %\nonl
            $\vartheta_{t+1} = \vartheta_{t} + \frac{\bt}{N}\sum_{i\in[N]}\delta_{t+1}^{i}.$

            %\vspace{1ex}
            
            %\State 
            Update Polyak-Ruppert average $\bar{\vartheta}_{t+1}  = \bar{\vartheta}_{t} +   \frac{1}{t+1}[\vartheta_{t} - \bar{\vartheta}_{t}].$ \label{alg:PR}

            %\vspace{1ex}
            
            %\State 
            Send $\vartheta_{t+1}$ to each agent $i\in[N].$
        }
   }
\end{algorithm}

In this section, we present our main convergence-rate results for policy evaluation using TD learning with PR-averaging.

The federated TD algorithms with PR-averaging: AvgFedTD(0) for average reward  and ExpFedTD(0) for exponential discounting are presented in Algorithms~\ref{AvgFedTD} and \ref{FedTD}, respectively. In AvgFedTD(0), each iteration has three key phases. In the first phase, each client node computes the local average reward estimate $r_{t + 1}^i$ using the universal $1/(t + 1)$ stepsize and the local TD error $\delta^i_{t + 1}$ and then transmits both these quantities to the central server. In the second phase, the server uses these values from  the clients to compute the global value function approximation parameter $\theta_{t + 1}$ using the universal stepsize $\beta_t$, the global average reward estimate $r_{t + 1},$ and the running average $\bar{\theta}_{t + 1}$ of $\theta_0, \ldots, \theta_{t}.$ In the final phase, the server broadcasts $\theta_{t + 1}$ and $r_{t + 1}$ to the clients. The ExpFedTD(0) algorithm is similar to AvgFedTD(0), except that there are no average reward estimates and the TD error involves the discount factor and is computed differently.

%as
%
%\[
%    \delta_{t+1}^{i} = \cR_{i}(s_{t}^{i},a_{t}^{i}) \phi(s_{t}^{i}) + \phi(s_{t}^{i})(\gamma\phi^{\tr}(s_{t+1}^{i}) - \phi^{\tr}(s_{t}^{i}))\theta_{t}.
%\]

\begin{remark}
In the average-reward setting, distributed TD learning for policy evaluation has not been studied; existing work considers only the single-agent case \cite{zhang2021finite}. Relative to that, the \(N=1\) case of AvgFedTD(0) differs in two ways: (i) we update \(\theta_t\) and \(r_t\) on different timescales---\(\theta_t\) on the faster timescale with stepsize \(\beta_t=(t+1)^{-\beta}\), \(\beta\in(1/2,1)\), and \(r_t\) on the slower timescale with stepsize \((t+1)^{-1}\); and (ii) we apply Polyak--Ruppert averaging to \(\theta_t\), with the average again updated on the slower timescale. In the exponential-discounting case, \cite{dal2023federated} studies federated TD; our ExpFedTD(0) differs by additionally incorporating PR-averaging.
\end{remark}

% while (considering the space constraints) we provide the algorithm for the exponentially discounted setup in the appendix. \footnote{In the exponentially discounted case, we are only concerned about the model parameters $(\theta_t)$ and their running averages $(\bar{\theta}_t).$ Moreover, at each iteration, the agent $i$ computes its local TD error as $\delta_{t+1}^{i} = (\cR_{i}(s_{t}^{i},a_{t}^{i}) - r_{t})\phi(s_{t}^{i}) + \phi(s_{t}^{i})(\gamma\phi^{\tr}(s_{t+1}^{i}) - \phi^{\tr}(s_{t}^{i}))\theta_{t},$ where $\gamma\in (0,1)$ is the discount factor.}  

To obtain finite-time bounds for the two algorithms, we make the following standard assumptions~\cite{zhang2021finite, wang2024federated}, where $\|\cdot\|$ and $\|\cdot\|_{F}$ are the Euclidean and Frobenius norms, respectively.
{
\renewcommand{\theenumi}{$\mathbf{\cA_\arabic{enumi}}$}
\begin{enumerate}  

\setcounter{enumi}{1}

     % \item\label{a: ergodic} For each $i\in [N],$ the Markov chain induced by $\bP_{i}^{\mu}$ on $\cS$ is irreducible and aperiodic.
     
     \item\label{a: dynamics.hetero} \textbf{Heterogeneity bound}: $\exists \ep,\er >0$ such that, $\forall i,j \in [N]$ and $ s, s'\in \cS,$
    $\lvert \cP_{i}^{\mu}(s,s') - \cP_{j}^{\mu}(s,s') \rvert  \leq \varepsilon_{p} \cP_i^{\mu}(s,s')$ and $\| \cR_{i} - \cR_{j} \| \leq \varepsilon_{r}.$

    \item\label{a: reward.bound} \textbf{Bounded rewards}: $\exists \Rm>0$ such that $\lvert \cR_{i}(s,a)\rvert \leq \Rm,$ $\forall i \in [N], \forall s\in \cS,$ and $\forall a \in \cA.$

    \item\label{a: feature.matrix} \textbf{Conditions on the feature matrix}: The matrix $\Phi$ has full-column rank with $\|\Phi\|_{F}=1.$ Additionally, for the average-reward case, the column space of $\Phi$ does not contain the vector of all ones, i.e., $\ones\notin \{\Phi\theta :\theta \in \bR^{d}\}.$
    
 \end{enumerate}   
}
%

%\noindent All our assumptions are standard . Assumption \ref{a: dynamics.hetero} puts an upper bound on the heterogeneity among the local environments, while \ref{a: feature.matrix} is required to ensure that our algorithm converges to a unique stable fixed point.

We also introduce some notation. For all $i \in [N],$ let $D_{i}^{\mu}:= 
\diag(d_{i}^{\mu}).$ Also, let $A_{i}:= \Phi^{\tr}D_{i}^{\mu}(I - \cP_{i}^{\mu})\Phi,$ $\Upsilon_{i}:= \Phi^{\tr}D_{i}^{\mu}(I - \gamma\cP_{i}^{\mu})\Phi,$ $v_i := \Phi^\tr D_i^\mu\ones,$ and $b_{i} := \Phi^{\tr}D_{i}^{\mu}\cR_{i}^{\mu}.$ Further, let $\theta_{i}^{*} := A_{i}^{-1}(b_{i} - v_{i}r^{*}_{i})$ and $\vartheta_{i}^{*} := \Upsilon_{i}^{-1}b_{i}.$ Assumptions~\ref{a: ergodic} and \ref{a: feature.matrix} guarantee the positive definiteness of $A_i$ and $\Upsilon_i$. Next, let $A := \frac{1}{N}\sum_{i\in [N]} A_{i},$ and $b := \frac{1}{N} \sum_{i\in [N]} b_{i}$ be the average of $A_i$'s and $b_i$'s. Similarly, let $v := \frac{1}{N}\sum_{i\in [N]} v_{i},$ $r^{*}:= \frac{1}{N}\sum_{i\in[N]} r_{i}^{*},$ and $\theta^{*} := A^{-1}(b - vr^{*}).$ The positive definiteness of $A$ follows from that of the $A_i$'s. Also, let $\mathbb{Z}_{+}$ be the set of positive integers. Due to  \ref{a: ergodic}, it is well known fact that $\exists \Cerg > 0$ and $\alpha\in (0,1)$ such that, for any $t \geq \tau \geq 0,$
\begin{equation}
    \label{e:ergodicity}
    \max\limits_{i\in[N]}\left\| \bP(s_t^i= \cdot|s^i_{t-\tau})- d^{\mu}_i(\cdot)\right\|_{\text{TV}} \leq \Cerg \alpha^{\tau}.
\end{equation}
Let $\lambda$ be a fixed number in $(0, \lambda_{\min}(A + A^\tr)$ and the stepsize $\bt = 1/(t + 1)^\beta$ for $\beta \in (1/2, 1).$ Finally, let $\taut:= \min\{ \tau \in \mathbb{Z}_{+} : \alpha^{\tau} < \frac{1}{(t+1)^{2}}\},$ and $\tstr:= \max\{\tstr^{(1)}, \tstr^{(2)}, \tstr^{(3)},  t_U \},$ with $\tstr^{(1)} := \min \{ t \in \mathbb{Z}_{+} : t \geq 2\tau_{t} + 2 \},$ $\tstr^{(2)} := \min \{ t \in \mathbb{Z}_{+} : \taut^2\beta^2_{t-\taut} <1/1248 \},$ $\tstr^{(3)} := \min \{ t \in \mathbb{Z}_{+} : \beta_{s - \tau_s} <\sqrt{2}\beta_s\ \forall s \geq t\},$ and $t_U$ is as defined in Table~\ref{tab: constants}.

We are now ready to state our main results.  

\begin{theorem}[AvgFedTD(0)]
    \label{thm : finite-time convergence}
    Assume \ref{a: ergodic}---\ref{a: feature.matrix} hold. Let $(\bar{\theta}_{t}, r_{t})$ be the iterates generated by \textnormal{AvgFedTD(0)}. Then, $\forall i\in [N]$ and $T > \tstr,$
    \begin{align}
    \label{eq:3_packet}
        \bE (r_{T} - r_{i}^{*})^{2}\leq {} & \frac{C_{r,\qd}}{(T+1)^{2}} + \frac{C_{r,\lin}\tauT^2}{N(T+1)} + H_r(\ep, \er)
        \\
        \bE\| \bar{\theta}_{T} - \theta_{i}^{*}\|^{2} \leq {} & \frac{C_{\bar{\theta},\qd}\ln^2(T)}{(T+1)^{2\beta}} + \frac{C_{\bar{\theta},\lin}\tauT^2}{N(T+1)} \nonumber \\
        {} & +\ H_\theta(\ep, \er),
        \label{eq:4_packet}
    \end{align} 
where the constants $C_{r,\qd}, C_{r, \lin}, C_{\bar{\theta}, \qd}, C_{\bar{\theta}, \lin}, H_r(\ep, \er),$ and $H_\theta(\ep, \er)$ are as defined in Table~\ref{tab: constants}.  The last two constants, which capture the heterogeneity gap, go to $0$ as $\max\{\ep, \er\} \to 0.$ Also, $\tau_T = O(\ln T).$
\end{theorem}

\begin{theorem}[ExpFedTD(0)]
    \label{thm : finite-time convergence.exp}
    Assume \ref{a: ergodic}---\ref{a: feature.matrix} hold. Let $(\bar{\vartheta}_{t})$ be the iterates generated by \textnormal{ExpFedTD(0)}. Then, $\forall i\in [N]$ and $T > \tstr,$
    \[
            \bE\| \bar{\vartheta}_{T} - \vartheta_{i}^{*}\|^{2} = O\left(\frac{1}{N(T + 1)}\right) 
            % \frac{\hat{C}_{\bar{\theta}, \qd}\ln^2(T)}{(T+1)^{2\beta}} + \frac{\hat{C}_{\bar{\theta}, \lin} \tauT^2}{N(T+1)} 
            + \hat{H}(\ep, \er),
    \]  
where 
% the constants $\hat{C}_{\bar{\theta}, \qd,}, \hat{C}_{\bar{\theta}, \lin},$ and 
$\hat{H}(\ep, \er)$ is as defined in Table~\ref{tab: constants}. Further, the heterogeneity gap $\hat{H}(\ep, \er) \to 0$ as $\max\{\ep, \er\} \to 0.$ 
% Also, $\tau_T = O(\ln (T + 1)).$

\end{theorem}

%A few remarks about our results are now in order. 

\begin{remark}
    For exponential discounting, \cite{dal2023federated} and \cite{wang2024federated} establish finite-time error bounds for federated TD learning in homogeneous and heterogeneous settings, respectively. However, their results require the stepsize to depend on the smallest eigenvalues of $\Upsilon_1, \ldots, \Upsilon_N.$ This is challenging in practice as these eigenvalues are influenced by the unknown transition probabilities in $\cP_1, \ldots, \cP_N.$ Our error bounds for \textnormal{ExpFedTD(0)} are comparable to those in \cite{wang2024federated}, but we use universal stepsizes, thanks to the use of iterate averaging.
\end{remark}
    
\begin{remark}
    \label{rem:ExpFedTD(0)}
    Since our bound in Theorem~\ref{thm : finite-time convergence.exp} closely aligns with those in \cite{dal2023federated, wang2024federated}, all the benefits of running the TD algorithm in a federated learning setup, as highlighted in these works, also apply to \textnormal{ExpFedTD(0)}. Specifically, in the homogeneous case where $\ep = \er = 0$, \textnormal{ExpFedTD(0)}'s error bound decays at the optimal rate of $O(1/(NT))$, which is statistically optimal for iterative stochastic optimization algorithms. Moreover, the number of iterations it requires to achieve an $\epsilon$-close solution is $O(1/(N\epsilon^2))$, which decreases linearly with the number $N$ of agents. %In other words, the sample size needed by each agent to get an $\epsilon$-close solution decreases linearly with $N,$ clearly demonstrating the advantage of cooperation. In the case where 
    When the local MDPs differ, the heterogeneity gap $\hat{H}_\theta(\ep, \er)$ is $O((\ep + \er)^2)$. Thus, even in this scenario, collaboration enables each agent to find an $O(\ep + \er)$-approximate solution for its optimal parameter with an $N$-fold speedup, mirroring the findings in \cite{wang2024federated}.
\end{remark}

\begin{remark}
%AvgFedTD(0) is the first federated TD learning algorithm of any kind for the average reward setup.    
For the average reward setting, no existing work achieves the optimal convergence rate with universal stepsizes. Our result is the first to do so, marking a novel contribution to both single-agent and federated TD learning. 
% While a naive algorithm can be obtained by combining the federated TD idea for exponential discounting from \cite{wang2024federated} and the single agent TD method for average reward from \cite{zhang2021finite}, it would differ from AvgFedTD(0) in two ways. One, this naive algorithm wouldn't include the $\bar{\theta}_t$ iterates and, two, both $\theta_t$ and $r^i_t$ would be updated using  a $c_1/(c_2 + t)$ type stepsize.
%
As in Remark~\ref{rem:ExpFedTD(0)}, \textnormal{AvgFedTD(0)} has an optimal convergence rate with a linear speedup in $N.$ 
\end{remark}

\begin{remark}
    We emphasize that all our results apply to more challenging but realistic Markovian sampling.
\end{remark}

\begin{figure*}[htbp]    
    \includegraphics[width=0.97\linewidth]{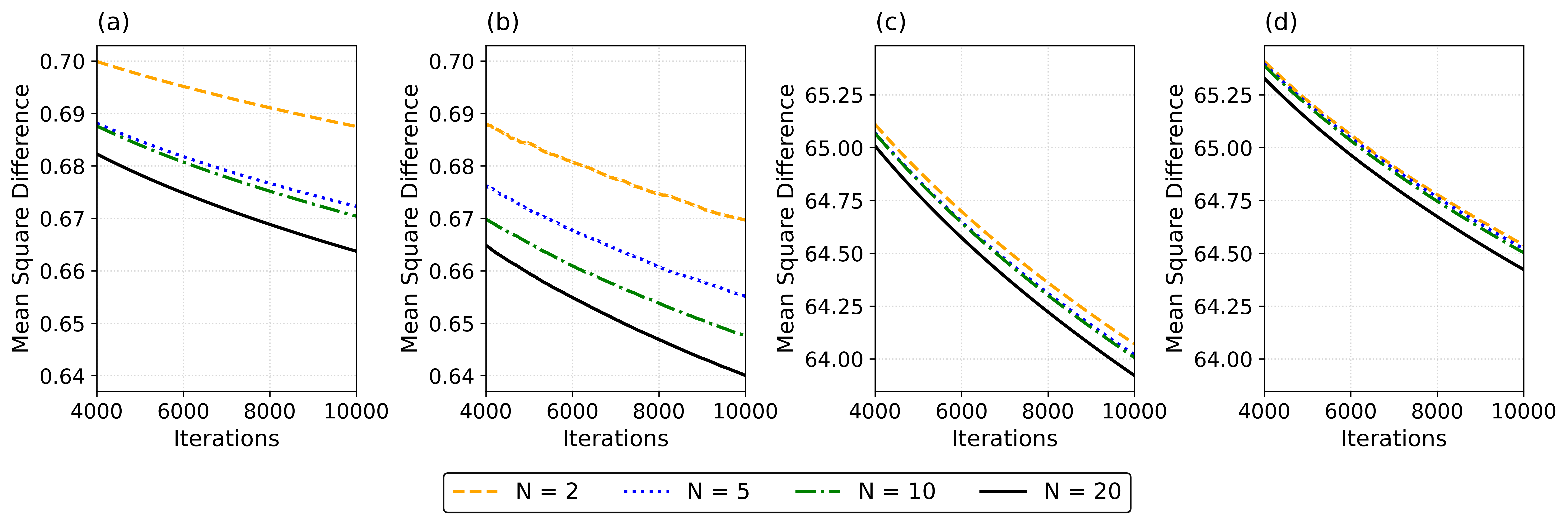}
    \caption{Evaluation of our proposed parameter-free algorithms with prior works. Specifically, for average reward, we compare AvgFedTD(0) (Fig.~a) with the federated variant of (Zhang et al., 2021) (Fig.~b). Similarly, for exponential discounting, we compare  ExpFedTD(0) (Fig.~c) to the federated TD method from~\cite{wang2024federated} (Fig.~d) for the setting described in Section~\ref{sec:experiments}. The y-axis of each plot is the mean square difference between the ideal parameters and global parameters, i.e., $\bE \|\bar{\theta}_t - \theta_{1}^{*}\|^2_{2},$ while the x-axis is the number of iterations. Clearly, our proposed parameter-free algorithms show comparable performance to the ones in the literature that depend on unknown problem parameters.}
    \label{fig:MSE.compare}
\end{figure*}

\section{Proofs}
\label{s: app.technical proofs}
In this section, we establish Theorems~\ref{thm : finite-time convergence} and \ref{thm : finite-time convergence.exp}. We begin in Section~\ref{s:Main.Proofs} by presenting the key intermediate lemmas and showing how they lead to our main results. Section~\ref{s:Intermediate.Proofs} then develops several technical results, which we use to prove these intermediate lemmas. Finally, Section~\ref{s:layer.3.proofs} provides the detailed proofs of these technical results. For clarity, all constants are summarized in Table~\ref{tab: constants}, while the remaining notations are defined in Sections~\ref{sec:setup} and \ref{sec:algorithm}.

\subsection{Proofs of Theorems~\ref{thm : finite-time convergence} and \ref{thm : finite-time convergence.exp}}
\label{s:Main.Proofs}

We begin with Theorem~\ref{thm : finite-time convergence}'s proof. From Algorithm~\ref{AvgFedTD}, it is easy to guess that $(r_t, \bar{\theta}_t)$ will converge to  $(r^*, \theta^*).$ The following result bounds the distance between $(r^*,\theta^*)$ and the solution $(r_i^*,\theta_i^*)$ that is local to agent $i$'s MDP.  

\begin{table*}[h]
    \centering
    \caption{Table of constants.}
    \begin{tabular}{|c|c|c|c|}
    \hline
    \textbf{Constants} & \textbf{Values} & \textbf{Constants} & \textbf{Values}\\
    \hline
    \hline 
        $C_{d}(\varepsilon_p)$ &  $\Big(\frac{1 + \varepsilon_p}{1 - \varepsilon_p} \Big)^{|\cS|} - 1 = 2 |\cS| \varepsilon_p + O(\varepsilon_p^2)$ 
        &
        $H_r(\varepsilon_p, \varepsilon_r)$     &   $2\Big(\varepsilon_r + \Rm C_d(\varepsilon_p)\Big)^2$
        \\[1ex]
        $A(\varepsilon_p)$ & $\varepsilon_p\sqrt{|\cS|} + C_d(\varepsilon_p) ( 1 +  \sqrt{|\cS|})$
        &
        $b(\varepsilon_p, \varepsilon_r)$   &   $\sqrt{|\cS|}\ (\varepsilon_r + C_d(\varepsilon_p)\  \Rm)$
        \\[1ex]
        $H_{\theta}(\varepsilon_p, \varepsilon_r)$ & $\max\limits_{i\in[N]}\frac{2\kappa^2(A_{i})\ \|A_{i}\|^{2}\|\theta^{*}_{i}\|^{2}}{[\|A_{i}\| - \kappa(A_{i}) A(\varepsilon_p)]^{2}}\Big[ \frac{A^2(\varepsilon_p)}{\|A_{i}\|^{2}} + \frac{b^2(\varepsilon_p, \varepsilon_r)}{\|b_{i} - v_i \rS_i\|^{2}} \Big]$ 
        & $\hat{H}(\varepsilon_p, \varepsilon_r)$ & $\max\limits_{i\in[N]}\frac{2\kappa^2(\Upsilon_{i})\ \|\Upsilon_{i}\|^{2}\|\theta^{*}_{i}\|^{2}}{[\|\Upsilon_{i}\| - \kappa(\Upsilon_{i}) \Upsilon(\varepsilon_p)]^{2}}\Big[ \frac{\Upsilon^2(\varepsilon_p)}{\|\Upsilon_{i}\|^{2}} + \frac{b^2(\varepsilon_p, \varepsilon_r)}{\|b_{i}\|^{2}} \Big]$ 
        \\[1ex]
        $C_{\rho,\text{W}}$    &   $4\Rm^2\max\{7, 2 \Cerg^{2}\}$   
        &
        $\xi_\rho$  &   $\exp\left[\sum_{t=\tstr}^{\infty} \frac{9\taut^{2}}{(t-\taut)^{2}}\right]$  
        \\[1ex]
        $C_{r,\lin}$    &   $16 \xi_{\rho} C_{\rho,\text{W}}$ 
        &
        $C_{r,\qd}$ &   $8\xi_{\rho}\Rm^{2}(\tstr+1)^{2} + \frac{64\xi_{\rho}C_{\rho,\text{W}}}{\tstr+1}$
        \\[1ex]
        $\Cm$   &   $40\cdot\max\{\Rm^2 + \|\theta^*\|^2, 6+\Cerg^2 \}$
        & $\Upsilon(\varepsilon_p)$ & $\varepsilon_p\sqrt{|\cS|} + C_d(\varepsilon_p) ( 1 +  \sqrt{|\cS|})$ 
        \\[1ex]
        $\lambda$ & \text{a fixed number in }$(0,\lmin(A+A^\top))$ & $\lambda_h, h$ & $\lambda h - \frac{\Cerg}{(1-\alpha)},$ \text{a fixed positive integer s.t.} $\lambda_h>0$
        \\[1ex]
        $\mu$ & $\lambda_{\max}(A^\tr A) - \lambda_{\min}(A + A^\tr)$ 
        &
        $t_A$ & $\max\left\{\left(\frac{\lambda_{\max}(A^\tr A)}{\lambda_{\min}(A + A^\tr) - \lambda}\right)^{1/\beta} - 1, 0\right\}$
        \\[1ex]
        $\CG$  &   $\max_{t_1<t_2<t_A} \prod_{t=t_1}^{t_2} e^{\bt(\lambda + \mu)}$
        & $\Clambda$ 
        &  $\big(\frac{2e^{\lambda/2}}{\lambda} \big)$
        \\[1ex]
        $\KG$ & $\big(\frac{e}{\lambda}\big)\CG e^{2\lambda}$  & $\CS$ & $4\CG e^\lambda\Big( 1+ \CG\sqrt{\frac{2\Cerg}{(1-\alpha)}}\Big)$
        \\[1ex]
        $\Cbeta$ & $\sum_{t=\tstr}^\infty \bt^2$ & $\CD$ & $(\|\thS\| + 2\tstr\Rm)e^{2\tstr}$
        \\[1ex]
        $\xi_M$ & $4(\|\thS\| + \Rm)$ & $B_L$ & $6\CG\Clambda(\|\thS\| + \Rm)$
        \\[1ex]
        $\xi_G$ & $\CG e^{-\frac{\lambda \tstr^{(1-\beta)}}{(1-\beta)}}\sum_{t=\tstr}^\infty e^{-\frac{\lambda t^{(1-\beta)}}{(1-\beta)}}$  & $\xi_{\Gamma}$ & $C_\Gamma e^{ \frac{\lambda_h (h + 1)^{1 - \beta}}{h (1 - \beta)}} \sum_{t = \tstr}^{\infty} e^{-\frac{\lambda_h(t + 1)^{1 - \beta}}{h(1 - \beta)} } $
        \\[1ex]
        $C_\Gamma$ & $2\lambda h^2(1+2\lambda)^h$ & $\xi_\Omega$ & $\big( \frac{2+2\beta}{e\lambda}\big)^{\frac{1}{1-\beta}}$
        \\[1ex]
        $\xfl^{(1)}$ & $2\CG\CS\xi_M(1+\Cerg)\Big[\Clambda+ \frac{\Clambda}{2(1-\beta)} + \frac{\Clambda}{2\CG} + \big(\frac{16}{\lambda} + \xi_G \big) \Big]$ & $\xfl^{(2)}$ & $\CS\Big[\Clambda \frac{C^{1/2}_{r,\qd}}{\sqrt{2}} + B_L \big(\frac{e^{\lambda\tstr/2}}{e\lambda}\big)^{\frac{2}{(1-\beta)}} \Big] + \CG^2\Clambda B_L \big(\frac{e^{\lambda\tstr}}{e\lambda}\big)^{\frac{1}{(1-\beta)}}$
        \\[1ex]
        $\xfl^{(3)}$ & $\CS\Clambda \frac{C^{1/2}_{r,\lin}}{\sqrt{2}}$ &
        $\Cealp$ & $\Cerg\Big[ \frac{\pi}{6} + \frac{2}{(1-\alpha)} + 2\frac{\sqrt{\Cbeta}}{\sqrt{1-\alpha^2}} \Big]$
        \\[1ex]
        $\xi^{(1)}_{L,\qd}$ & $2\CG\Clambda \xi_G\xi_M + \KG \frac{C^{1/2}_{r,\qd}}{\sqrt{2}} + \KG\xi_M \Cealp$ & $\xi^{(1)}_{L,\lin}$ & $[4\KG\xi_M + \sqrt{2}\KG C^{1/2}_{r,\lin}]\Big[ 1 + \frac{1}{\ln(1/\alpha)} \Big]$
        \\[1ex]
        $\xi_{FL,\qd}$ & $\Big[\xfl^{(2)} + \frac{\xfl^{(1)}}{(1-\beta)}\Big(1 + \frac{1}{\ln(1/\alpha)} \Big)\Big] \Big[1 + \frac{4C_\Gamma h}{\lambda_h} \Big]$ & $\xi_{FL,\lin}$ & $2\xfl^{(3)}\Big[1 + \frac{4C_\Gamma h}{\lambda_h} \Big]$
        \\[1ex]
        $C_{\bar{\theta}, \lin}$ & $4\Big[ \xi^{(1)}_{L,\lin} + \xi^{(1)}_{FL,\lin}\Big]^2$ &
        $C_{\bar{\theta},\qd}$ & $8\Big[ \CD(\tstr+\xi_\Gamma) + \xi^{(1)}_{L,\qd} + \xi_{FL,\qd}\Big]^2$
        \\[1ex]
    \hline 
    \end{tabular}
    \label{tab: constants}
\end{table*}

% \subsection{Proofs of heterogeneity bound}

\begin{lemma}\label{thm: perturb.bound}
    For each $i\in [N],$     $2(r^{*} - r^{*}_{i} )^{2} \leq H_r(\varepsilon_{p}, \varepsilon_{r})$ and $2\| \theta^{*} - \theta_{i}^{*}\|^{2} \leq H_\theta(\varepsilon_{p}, \varepsilon_{r})$, where $H_r(\ep, \er)$ and $H_\theta(\ep, \er)$ are as defined in Table~\ref{tab: constants}.
\end{lemma}

This claim's proof from that of \cite[Theorem~1]{wang2024federated}. 

Next, we derive the rates at which $(r_t)$ and $(\bar{\theta_t})$ converge to $\rS$ and $\thS,$ respectively. The two-timescale nature of our algorithm allows us to analyze $(r_t)$'s convergence rate independently to that of $(\theta_t).$ For all $t \geq 0$ and $i\in[N],$ let 
\begin{equation}
        \Wt^{(i)} := \cR_{i}(s_{t}^{i}, a_{t}^{i}) - r_i^* \hspace{0.5em} \text{ and } \hspace{0.5em} \Wt := \frac{1}{N}\sum_{i=1}^N W^{(i)}_{t+1}.
        \label{e:Wt.defn}
        % \Mt^{(i)} := & (\cR_{i}(s_{t}^{i}, a_{t}^{i})\phi(s_{t}^{i}) - b_{i}) - (\phi(s_{t}^{i}) - v_{i})r_t  \nonumber \\
        % & - (\phi(s_{t}^{i})(\phi^{\tr}(s_{t}^{i}) - \phi^{\tr}(s_{t+1}^{i})) - A_{i})\theta_t 
        % \label{e:Mt.defn},
\end{equation}
Further, let 
\begin{align}
    %\begin{aligned}
    \nonumber \hat{A}^i_{t}  & :=  \phi(s_{t}^{i})\left( \phi(s_{t}^{i}) - \phi(s_{t+1}^{i}) \right)^{\tr}, \\
    \hat{v}^i_{t} & := \phi(s_{t}^{i}), \\
    \nonumber \zht^i 
    & := \left[\cR_{i}(s_{t}^{i}, a_{t}^{i})\phi(s_{t}^{i}) - b_{i}\right] - \left[\hat{v}_{t}^{i} - v_{i} \right]r^{*} 
    \\
    \nonumber & \qquad +  \left[A_{i} - \hat{A}^i_t \right]\thS.
\end{align}
Also, let 
\begin{equation}
\label{e: noise.mat} 
    \Aht = \frac{1}{N}\sum_{i=1}^{N}\hat{A}^{i}_{t},\ \vht = \frac{1}{N}\sum_{i=1}^{N}\hat{v}^{i}_{t}, \text{ and } \zht = \frac{1}{N}\sum_{i=1}^{N}\hat{z}^{i}_{t}.
\end{equation}
Finally, let $\rho_{t}:= r_{t} - r^{*}, $ $\Delta_{t} := \theta_{t}-\theta^{*},$ and $\Delbar_{t} := \bar{\theta}_{t}-\theta^{*}.$ Then, we have from Algorithm~\ref{AvgFedTD} that
\begin{align*}
    \delta_{t + 1}^i = {} & \cR_{i}(s_{t}^{i},a_{t}^{i}) \phi(s_t^i) - \hat{v}^i_t r_t - \hat{A}^i_t \theta_t \nonumber \\
    %
    % = {} & \cR_{i}(s_{t}^{i},a_{t}^{i}) \phi(s_t^i)  - \hat{v}^i_t r^* - \hat{A}^i_t \thS - \hat{v}^i_t \rho_t - \hat{A}^i_t \Delta_t \nonumber \\
    %
    = {} & \hat{z}^i_t - \hat{v}^i_t \rho_t - \hat{A}^i_t \Delta_t + b_i - v_i \rS - A_i \thS.
\end{align*}
Hence, it follows that
\begin{align*}
    \frac{1}{N} \sum_{i \in [N]} \delta_{t + 1}^i = {} & \hat{z}_t - \hat{v}_t \rho_t - \hat{A}_t \Delta_t + b - v \rS - A \thS \\
    = {} & \hat{z}_t -  \hat{v}_t \rho_t - \hat{A}_t \Delta_t,
\end{align*}
where the last relation holds since $A \thS = b - v \rS.$ Finally, from Algorithm~\ref{AvgFedTD} and the above relations, it can be seen that $\rho_{t},$ $\Delta_{t},$ and $\Delbar_{t}$ satisfy
\begin{equation}\label{e: update.diff.rule}
    \begin{aligned}
        \rho_{t+1}  & =  \Big(1 - \frac{1}{t+1}  \Big)  \rho_{t}  + \frac{1}{t+1}\Wt,
        \\
        \Delta_{t+1} & = (I - \bt\Aht)\Delta_t - \bt\vht\rho_t + \bt\zht, 
        \\
        \Delbar_{t+1} & = \Big(1 - \frac{1}{t+1}\Big)\Delbar_{t} +  \frac{1}{t+1} \Delta_{t}.
    \end{aligned}
\end{equation}

Using these update rules, we obtain the convergence rates for $(r_{t})$ and $(\bar{\theta}_t),$ which are given in Lemmas \ref{thm: avg.rew.converge} and \ref{thm: avg.value.converge}, respectively, whose proofs are in Sections~\ref{s:Intermediate.Proofs}.
\begin{lemma}\label{thm: avg.rew.converge}
    For $T>\tstr,$
    \begin{equation}\label{e: avg.rew.converge}
        \bE\rho_{T}^{2} \leq \frac{C_{r,\qd}}{2(T+1)^{2}} + \frac{C_{r,\lin}\ \tauT^{2}}{2N(T+1)}, 
    \end{equation}
    where the constants $C_{r,\qd}$ and $C_{r, \lin}$ are as defined in Table~\ref{tab: constants}.
\end{lemma}

\begin{lemma}\label{thm: avg.value.converge}
    For $T>\tstr,$ 
    \[
        \bE\| \bar{\Delta}_{T}\|^2 \leq \frac{C_{\bar{\theta},\qd}\ln^2 (T)}{2(T+1)^{2\beta}} + \frac{C_{\bar{\theta},\lin}\  \tauT^2 }{2N(T+1)},
    \]
    where the constants $C_{\bar{\theta},\qd}$ and $C_{\bar{\theta},\lin}$ are as defined in Table~\ref{tab: constants}.
\end{lemma}

We now prove Theorem~\ref{thm : finite-time convergence}. 

\noindent \emph{\textbf{Proof of Theorem~\ref{thm : finite-time convergence}}}. For all $i \in [N],$ using the fact that $(a + b)^2 \leq 2 a^2 + 2b^2,$ we get
\begin{align}
\label{eq:1_packet}
     \bE(r_T - r_i^*)^2 &\leq 2\bE(r_T - r^*)^2 + 2(r^* - r_i^*)^2\\
     \bE\|\bar{\theta}_T - \theta_i^*\|^2 &\leq 2\bE\|\bar{\theta}_T - \theta^*\|^2 + 2\|\theta^* - \theta_i^*\|^2.
     \label{eq:2_packet}
\end{align}
Since $\rho_T = r_T - r^*,$ using Lemma~\ref{thm: avg.rew.converge} and Lemma~\ref{thm: perturb.bound} in~(\ref{eq:1_packet}) yields~(\ref{eq:3_packet}). Similarly, since $\bar{\Delta}_T = \bar{\theta}_T - \theta^*,$ using Lemma~\ref{thm: avg.value.converge} and Lemma~\ref{thm: perturb.bound} in~(\ref{eq:2_packet}) yields~(\ref{eq:4_packet}). \hfill $\blacksquare$

Proving Theorem~\ref{thm : finite-time convergence.exp} requires the same recipe. Similar to Lemmas~\ref{thm: perturb.bound}, \ref{thm: avg.rew.converge}, and \ref{thm: avg.value.converge}, we can show the following results.
\begin{lemma}
\label{lem:exp_1}
    For each $i\in [N],$ $2\|\vartheta^* - \vartheta^*_i\|^2 \leq \hat{H}(\varepsilon_p, \varepsilon_r).$
\end{lemma}

% Similarly, we can show that the following holds for the iterates $(\vartheta_t)$ generated by ExpFedTD(0):
%
\begin{lemma}
\label{lem:exp_2}
    For $T>\tstr,$
    \begin{align*}
    %
   %      \bE\|\vartheta_T - \vartheta^*\|^2 &= \tilde{O}\left(\frac{\beta_T}{N} \right),
   % \\
   \bE\|\bar{\vartheta}_T - \vartheta^*\|^2 &= O\left(\frac{1}{N(T + 1)}\right).
   % \frac{\hat{C}_{\bar{\theta}, \qd}\ln^2(T)}{2 (T+1)^{2\beta}} + \frac{\hat{C}_{\bar{\theta}, \lin}\  \tauT^2}{2 N(T+1)}.
    %
    \end{align*} 
\end{lemma}

\noindent \emph{\textbf{Proof of Theorem~\ref{thm : finite-time convergence}}}. For all $i \in [N],$ we have
\[
    \bE\|\bar{\vartheta}_T - \vartheta_i^*\|^2 \leq 2\bE\|\bar{\vartheta}_T - \vartheta^*\|^2 + 2\|\vartheta^* - \vartheta_i^*\|^2.
\]
The desired bound now follows from Lemmas~\ref{lem:exp_1} and~\ref{lem:exp_2}. \hfill $\blacksquare$

\subsection{Proofs of our Key Intermediate Lemmas}
\label{s:Intermediate.Proofs}

Lemmas~\ref{thm: avg.rew.converge} and \ref{thm: avg.value.converge} are needed to prove Theorem~\ref{thm : finite-time convergence}. In this section, we prove these key technical results. The proofs of Lemmas~\ref{lem:exp_1} and \ref{lem:exp_2} follow similarly; hence, we skip those. We again highlight that the definition of all our notations can be found in Sections~\ref{sec:setup} and \ref{sec:algorithm}.

We begin with Lemma~\ref{thm: avg.rew.converge}'s proof. Let $\cF_t := \sigma(\{s_k^i  : k< t, i\in [N] \})$ and $\bE_t[\cdot]$ denote $\bE[\cdot | \cF_t].$ For $t = 0,$ $\bE_t = \bE$ and $s^i_0,$ $i \in [N],$ is presumed to be sampled from some arbitrary but fixed initial distribution.

We need the following technical result  to prove Lemma~\ref{thm: avg.rew.converge}.

\begin{lemma} 
\label{lemma:AandB}
The following statements hold. 
\begin{enumerate}[label=(\roman*)]
    \item For any $t \geq \tau \geq 0,$ we have
    \begin{equation}
        \label{e:Wt.Bd}
        \lvert\bE_{t - \tau} \Wt^{(i)}\rvert\leq  \Cerg \Rm\alpha^{\tau}.
    \end{equation}
    
    \item For all $t \geq \tstr,$
    \begin{equation}
    \label{eq:bound_A}
        \bE \Wt^{2} \leq \frac{4\Rm^{2}}{N} + \frac{\Cerg^{2}\Rm^{2}}{(t + 1)^4} .
    \end{equation}

    \item For all $t \geq \tstr,$
    \begin{multline}
    \lvert 2\bE \rho_{t} \Wt\rvert \leq 
    \frac{(t + 1)}{(t - \tau_t)^2} \bigg[8 \taut^2 \bE \rho_t^2 \\
    +\ C_{\rho,\text{W}} \left(\frac{\taut^2}{N} + \frac{1}{(t- \tau_t)^{2}} \right) \bigg]. \label{eq:bound_B}
    \end{multline}

    \item Define $\xi_\rho:= \exp\left(\sum_{t=\tstr}^\infty \frac{9 \taut^{2}}{(t-\taut)^{2}}\right);$ this is finite by Cauchy's condensation test. Then, for $\tstr \leq t_{1}<t_{2},$ 
    \begin{align}
    G_{t_{1}, t_{2}}^{\rho} &:= \prod_{t=t_{1}}^{t_{2}-1}\Big(1 -\frac{2}{t+1} + \frac{9\taut^{2}}{(t-\taut)^{2}} \Big) \label{e:G.t1.t2.Defn} \\
    &\leq \xi_{\rho}\Big(\frac{t_{1}+1}{t_{2}+1}\Big)^{2} \label{eq:bound_D}.
    \end{align}
    Further, for all $T > \tstr,$
    \begin{align}
    \sum_{t=\tstr}^{T-1} \Big( & \frac{\taut^{2}}{N (t - \taut)^{2}} + \frac{1}{(t - \tau_t)^4}\Big) G_{t+1:T}^{\rho} \nonumber \\
    \leq {} &  \xi_\rho  \Big(\frac{4\tau_T^2 }{N(T + 1)} + \frac{16}{(T + 1)^2(\tstr + 1)}\Big). \label{eq:bound_E}
\end{align}
\end{enumerate}
\end{lemma}

The proof of this result is given in Section~\ref{s:layer.3.proofs}. We now use the above result to prove Lemma~\ref{thm: avg.rew.converge}.

\emph{Proof of \textbf{Lemma~\ref{thm: avg.rew.converge}}}. From \eqref{e: update.diff.rule}, we have
\begin{align*}
    & \bE \rho^2_{t + 1} \\
    \leq {} & \frac{t^2}{(t + 1)^2} \bE \rho^2_t + \frac{|2\bE \rho_t W_{t + 1}|}{t + 1} + \frac{\bE W^2_{t + 1}}{(t + 1)^2} \\
    = {} & \left[1 - \frac{2}{t + 1} + \frac{1}{(t + 1)^2}\right] \bE \rho_t^2 + \frac{|2\bE \rho_t W_{t + 1}|}{t + 1} + \frac{\bE W^2_{t + 1}}{(t + 1)^2}. 
\end{align*}
Substituting \eqref{eq:bound_A} and \eqref{eq:bound_B} in the above inequality leads to 
\begin{align}
    \bE & \rho^2_{t + 1} \nonumber \\
    \overset{(a)}{\leq} {} & \left[1 - \frac{2}{t + 1} + \frac{9 \tau_t^2}{(t - \tau_t)^2}\right] \bE \rho_t^2 + \frac{4 \Rm^2}{N(t + 1)^2} + \frac{C_E^2 \Rm^2}{(t + 1)^6} \nonumber \\
    {} & +\ \frac{1}{(t - \tau_t)^2} C_{\rho, \text{W}} \left(\frac{\tau_t^2}{N} + \frac{1}{(t - \tau_t)^2}\right) \nonumber \\
    \overset{(b)}{\leq} {} & \left[1 - \frac{2}{t + 1} + \frac{9 \tau_t^2}{(t - \tau_t)^2}\right] \bE \rho_t^2 + \frac{2C_{\rho, \text{W}} \tau_t^2}{N (t - \tau_t)^2} +  \frac{2C_{\rho, \text{W}} }{ (t - \tau_t)^4}, \nonumber
\end{align}
where (a) follows since $\taut^2 \geq 1,$ and (b) holds since $4\Rm^2\leq C_{\rho, \text{W}}.$ Now, by iterated application of the above inequality and using the definition of $G_{t_1, t_2}^\rho$ from  \eqref{e:G.t1.t2.Defn}, we get 
\begin{multline*}
    \bE \rho_T^2 \leq \Grho_{\tstr:T} \bE \rho_{\tstr}^2 \\
    + 2C_{\rho, \text{W}} \sum_{t = \tstr}^{T - 1} \Grho_{t + 1:T} \Big(\frac{\tau_t^2}{N (t - \tau_t)^2} + \frac{1}{(t - \tau_t)^4}\Big).
\end{multline*}
Finally, substituting \eqref{eq:bound_D} and \eqref{eq:bound_E} in the above inequality gives
\begin{align}
    \bE \rho_T^2 \leq {} &  \xi_\rho \Big(\frac{\tstr + 1}{T + 1}\Big)^2 \bE \rho_{\tstr}^2 \nonumber \\
    & + 2C_{\rho, \text{W}} \xi_\rho \Big(\frac{4\tau_{T}^2}{N(T + 1)} + \frac{16}{(T + 1)^2(\tstr + 1)}\Big).
    \label{e:rho.T.2.Intermediate.Bd}
\end{align}

From \eqref{e:Wt.defn}, we have $|W_{t + 1}| \leq 2\Rm$ for all $t \geq 0.$ Combining this fact with \eqref{e: update.diff.rule} then shows $|\rho_1| \leq |W_1| \leq 2 \Rm$ and
\[
    |\rho_{t + 1}| \leq \frac{t}{t + 1}|\rho_t| + \frac{2 \Rm}{t + 1}.
\]
Using induction, it is now easy to see that $|\rho_{t + 1}| \leq 2 \Rm$ for $t \geq 0;$ in particular, this shows that $\bE \rho^2_{\tstr} \leq 4 \Rm^2.$

To complete the proof of Lemma~\ref{thm: avg.rew.converge}, we substitute this last inequality in \eqref{e:rho.T.2.Intermediate.Bd} and use the definitions of $C_{r, \lin}$ and $C_{r, \qd}$ from Table~\ref{tab: constants}. \hfill \scalebox{0.8}{$\blacksquare$}

Next, we derive Lemma~\ref{thm: avg.value.converge}. Recall the definitions of $\Aht, \vht,$ and $\zht$ from \eqref{e: noise.mat}. 
Also, let $\tstr$ be as defined above Theorem~\ref{thm : finite-time convergence}.

From the update rule for $\Delta_t$ from \eqref{e: update.diff.rule}, we have
\begin{equation}\label{e: delta.decomp}
    \Delta_t = \Gamma_{0:t}\Delta_0 + \Delta^{(2)}_t,
\end{equation}
where 
\begin{equation}
\label{Delta.2.Relation}
    \Delta^{(2)}_t :=  \sum_{k=0}^{t-1}\bk\Gamma_{k+1:t}\left(-\rho_k\vhk + \zhk\right),
\end{equation}
and, for all $0 \leq t_t < t_2,$
\begin{align}
    \Gamma_{t_1:t_2} := \prod_{k=t_1}^{t_2-1}(I - \bk\hat{A}_k).
\end{align}
Consequently, for any $T > \tstr,$
\[
    \Delbar_T =  \frac{1}{T}\sum_{t=0}^{\tstr-1}\Delta_t + \frac{1}{T}\sum_{t=\tstr}^{T-1}\Gamma_{0:t}\Delta_0 + \frac{1}{T}\sum_{t=\tstr}^{T-1}\Delta^{(2)}_t.
\]

We now rewrite the above expression to enable our subsequent analysis. For any $t \geq 0,$ we have from \eqref{Delta.2.Relation} that
\[
    \Delta^{(2)}_{t+1} = (I - \bt\Aht)\Delta^{(2)}_t + \bt\left(-\rho_t \vht + \zht \right).
\]
Hence, if we let  $\Atil := \Aht - A,$  $L^{(1)}_0 = L^{(2)}_0 = L^{(3)}_0 = 0,$ and 
\begin{equation}
\label{e: L.update}
    \begin{aligned}
        L^{(1)}_{t+1} = & (I-\bt A)L^{(1)}_t - \bt\rho_t\vht + \bt\zht
        \\
        L^{(2)}_{t+1} = & (I- \bt A)L^{(2)}_t - \bt\tilde{A}_tL^{(1)}_t
        \\
        L^{(3)}_{t+1} = & (I- \bt\Aht)L^{(3)}_t - \bt\tilde{A}_tL^{(2)}_t,
    \end{aligned}
\end{equation}
then a simple inductive argument shows that, for any $t \geq 0,$ 
\[
    \Delta^{(2)}_t = L^{(1)}_t + L^{(2)}_t +L^{(3)}_t.
\]
Thus, for any $T>\tstr,$ we can rewrite the $\Delbar_T$ as
\begin{multline}\label{e: bar.delta.decomp}
    \Delbar_T =  \frac{1}{T}\sum_{t=0}^{\tstr-1}\Delta_t + \frac{1}{T}\sum_{t=\tstr}^{T-1}\Gamma_{0:t}\Delta_0 
    \\
    + \frac{1}{T}\sum_{t=\tstr}^{T-1}L^{(1)}_t +\frac{1}{T}\sum_{t=\tstr}^{T-1}\left( L^{(2)}_t + L^{(3)}_t \right).
\end{multline}

The following lemma provides bounds on each term in \eqref{e: bar.delta.decomp}. gma

\begin{lemma}\label{thm: PR.bounds}
    The following statements hold. 
    \begin{enumerate}
        \item For $t \leq \tstr,$ $
            \|\Delta_t\| \leq \CD.$
            
        \item Let $\Cerg$ and $\alpha$ be as in \eqref{e:ergodicity}, and $\lambda\in (0,\lmin(A+A^\top))$ be a fixed constant. Let $h$ be a fixed integer such that
        \[
            h > \frac{\Cerg}{\lambda(1 - \alpha)} \text{ and }   \lambda_h = \lambda h - \frac{\Cerg}{1 - \alpha}.
        \]
        Then, for $0 \leq t_1 < t_2,$ we have
        \begin{equation}\label{e: gamma_sq}
           \bE_{t_1}\left\|\Gamma_{t_1:t_2}\right\|^2 \leq C_{\Gamma}e^{\big(-2\lambda_h\sum_{i=1}^{\floor{(t_2-t_1)/h}} \beta_{t_1+ih}\big)}.
        \end{equation}
        
        \item For any $T >\tstr,$
        \begin{equation}
        \label{e: sum.L1}
            \bE^{1/2}\Big\|\sum_{t=\tstr}^{T-1} L^{(1)}_t\Big\|^2 \nonumber
            \leq 
            \xi^{(1)}_{L,\qd}\ln(T)
            + \frac{\xi^{(1)}_{L,\lin}}{\big[ 1 + \frac{1}{\ln(1/\alpha)} \big]}\frac{\tauT\sqrt{T}}{\sqrt{N}}.
        \end{equation}
                
        \item Let $\xi_{FL}$ be defined as in Table~\ref{tab: constants}. For any $t \geq \tstr,$
        \begin{align}
            \bE^{1/2}\|L^{(2)}_t\|^2 & \leq f_{L}(t), \label{e: order.L2}
            \\
            \bE^{1/2}\|L^{(3)}_t\|^2 & \leq \Big(\frac{4C_\Gamma h}{\lambda_h}\Big) f_L(t), \label{e: order.L3}
        \end{align}
        where
        \begin{equation}\label{e: fl(t)}
            f_L(t) := \xfl^{(1)} \taut\bt + \frac{\xfl^{(2)}}{t} + \frac{\xfl^{(3)}}{\sqrt{N(t+1)}}.
        \end{equation}
    \end{enumerate} 
\end{lemma}

The proof of this result is in Section~\ref{s:layer.3.proofs}.  Assuming this result to be true, we now prove Lemma~\ref{thm: avg.value.converge}.

\emph{Proof of \textbf{Lemma~\ref{thm: avg.value.converge}}}. Using triangle inequality and the fact $\|\Delta_0\| \leq \CD$, it follows from \eqref{e: bar.delta.decomp} that, for any $T > \tstr,$
\begin{multline*}
        \bE^{1/2}\|\Delbar_T\|^2 
        \leq \frac{1}{T} \sum_{t = 0}^{\tstr - 1} \bE^{1/2} \|\Delta_t\|^2 + \frac{\CD}{T}\sum_{t=\tstr}^{T-1}\bE^{1/2}\|\Gamma_{0:t}\|^2 
        \\
        +\ \frac{1}{T}\bE^{1/2}\bigg\|\sum_{t=\tstr}^{T-1}L^{(1)}_t\bigg\|^2 + \frac{1}{T}\sum_{t=\tstr}^{T-1}\left[ \bE^{1/2}\|L^{(2)}_t\|^2 + \bE^{1/2}\|L^{(3)}_t\|^2 \right].
\end{multline*}
We now bound the four terms on the RHS. \\

\noindent \textbf{Term 1}: From Statement~1 in Lemma~\ref{thm: PR.bounds}, we get
\begin{equation}
    \frac{1}{T} \sum_{t = 0}^{\tstr - 1} \bE^{1/2} \|\Delta_t\|^2 \leq \frac{\CD \tstr}{T} \leq \frac{\CD \tstr}{T^\beta},
\end{equation}
which bounds the first term. \\

\noindent \textbf{Term 2}: Next, from Statement~2 of Lemma~\ref{thm: PR.bounds}, we get
\[
    \bE^{1/2} \|\Gamma_{0:t}\|^2 \leq \sqrt{C_\Gamma} e^{-\lambda_h \sum_{i = 1}^{\lfloor t/h \rfloor} \beta_{ih}}.
\]
Now, 
\begin{align*}
    &\sum_{i = 1}^{\lfloor t/h \rfloor} \beta_{ih} 
    = \sum_{i = 1}^{\lfloor t/h \rfloor} \frac{1}{(ih + 1)^\beta} 
    \geq \int\limits_1^{\lfloor t/h\rfloor+1}\frac{\df{x}}{(hx+1)^\beta}
    \\
    & \geq \frac{(h(\lfloor t/h\rfloor+1)+ 1)^{(1 - \beta)} - (h+1)^{(1-\beta)}}{h(1 - \beta)} \\
    & \geq \frac{(t + 1)^{1 - \beta} - (h + 1)^{1 - \beta}}{h (1 - \beta)}.
\end{align*}
Therefore, 
\begin{align*}
    \sum_{t = \tstr}^{T - 1} \bE^{1/2} \|\Gamma_{0:t}\|^2 & \leq C_\Gamma e^{\frac{\lambda_h(h+1)^{1-\beta}}{h(1-\beta)}} \sum_{t=\tstr}^{T-1} e^{-\frac{\lambda_h}{h(1-\beta)} t^{(1-\beta)}}
    \\
    & \leq \xi_\Gamma,
\end{align*}
where $\xi_\Gamma$ is as in Table~\ref{tab: constants}. Hence, 
\begin{equation}
    \frac{\CD}{T} \sum_{t = \tstr}^{T - 1} \bE^{1/2} \|\Gamma_{0:T}\|^2 \leq \frac{\CD \xi_{\Gamma}}{T} \leq \frac{\CD \xi_{\Gamma}}{T^\beta}.
\end{equation}
%
%
% where (a) follows since $t \geq \tstr,$ while (b) follows since $\sum_{t=1}^\infty\frac{1}{t^2}= \frac{\pi^2}{6}.$

% $t^2\cdot e^{-\frac{\lambda_h}{h(1-\beta)}t^{(1-\beta)}} \leq ( 2h/e\lambda_h)^{2/(1-\beta)}.$

\noindent \textbf{Term 3}: From Statement 3 of Lemma~\ref{thm: PR.bounds}, we have
\begin{align*}
    \bE^{1/2}\Big\|\sum_{t=\tstr}^{T-1} L^{(1)}_t\Big\|^2
    \leq \xi^{(1)}_{L,\qd}\ln(T)
            + \frac{\xi^{(1)}_{L,\lin}}{\big[ 1 + \frac{1}{\ln(1/\alpha)} \big]}\frac{\tauT\sqrt{T}}{\sqrt{N}}.
\end{align*}
Using $1\leq \ln(T),$ $\tauT \leq \big[ 1 + \frac{1}{\ln(1/\alpha)} \big]\ln(T),$ and $1/T\leq 1/T^\beta$ gives the desired bound.

\noindent \textbf{Term 4}: From Statement 4 of Lemma~\ref{thm: PR.bounds}, we get
\begin{multline*}
    \frac{1}{T}\sum_{t=\tstr}^{T-1}\Big[ \bE^{1/2}\|L^{(2)}_t\|^2 + \bE^{1/2}\|L^{(3)}_t\|^2\Big]
    \\
    \leq \bigg[ 1+ \Big(\frac{4C_\Gamma h}{\lambda_h}\Big) \bigg]\frac{1}{T}\sum_{t=\tstr}^{T-1}f_L(t).
\end{multline*}
We now bound $\sum_{t = \tstr}^{T - 1} f_L(t).$ Using its definition and the fact that $\taut \leq \tauT,$ we get
\begin{align*}
    & \sum_{t=\tstr}^{T-1} f_L(t) 
    \\
    & = \xfl^{(1)}\sum_{t=\tstr}^{T-1} \taut\bt + \xfl^{(2)}\sum_{t=\tstr}^{T-1}\frac{1}{t} 
    + \xfl^{(3)}\sum_{t=\tstr}^{T-1}\frac{1}{\sqrt{N(t+1)}}
    \\
    & \leq \tauT\xfl^{(1)}\sum_{t=\tstr}^{T-1} \frac{1}{(t+1)^\beta} + \xfl^{(2)}\sum_{t=\tstr}^{T-1}\frac{1}{t} 
    + \frac{\xfl^{(3)}}{\sqrt{N}}\sum_{t=\tstr}^{T-1}\frac{1}{\sqrt{(t+1)}}
    \\
    & \leq \frac{\tauT\xfl^{(1)}}{(1-\beta)}T^{(1-\beta)} + \xfl^{(2)}\ln(T) + 2\xfl^{(3)}\frac{\sqrt{T}}{\sqrt{N}}.
\end{align*}
Therefore,
\begin{align*}
    &\frac{1}{T}\sum_{t=\tstr}^{T-1} f_L(t) \leq \frac{\tauT\xfl^{(1)}}{(1-\beta)T^{\beta}} + \xfl^{(2)}\frac{\ln(T)}{T} + \frac{2\xfl^{(3)}}{\sqrt{NT}}
    \\
    & \overset{(a)}{\leq} \bigg[\frac{\xfl^{(1)}}{(1-\beta)}\Big[1 + \frac{1}{\ln(1/\alpha)}\Big] + \xfl^{(2)}\bigg]\frac{\ln(T)}{T^\beta} + \frac{2\xfl^{(3)}\ln(T)}{\sqrt{NT}},
\end{align*}
where $(a)$ uses $\tauT\leq \Big[1 + \frac{1}{\ln(1/\alpha)} \Big]\ln(T),$ $1/T\leq 1/T^\beta,$ and $1\leq \ln(T).$

Thus, letting $\xi_{FL,\qd}$ and $\xi_{FL,\lin}$ be defined as in Table~\ref{tab: constants} gives us 
\begin{multline*}
    \frac{1}{T}\sum_{t=\tstr}^{T-1}\Big[ \bE^{1/2}\|L^{(2)}_t\|^2 + \bE^{1/2}\|L^{(3)}_t\|^2\Big]
    \\
    \leq \xi_{FL,\qd}\frac{\ln(T)}{T^\beta} + \xi_{FL,\lin}\frac{\ln(T)}{\sqrt{NT}}.
\end{multline*}

At last, we combine the bounds on \textbf{Terms} \textbf{1}, \textbf{2}, \textbf{3}, \textbf{4}. We then use $1\leq \ln(T)$ the fact that $1/T \leq 2/(T+1),$ to obtain 
\begin{align*}
    \bE^{1/2}\|\Delbar_T\|^2 &\leq \sqrt{2}\Big[\xi^{(1)}_{L,\lin} + \xi_{FL,\lin} \Big]\frac{\ln(T)}{\sqrt{N(T+1)}} 
    \\
    & + 2^\beta\Big[ \CD(\tstr+ \xi_\Gamma) + \xi^{(1)}_{L,\qd} + \xi_{FL,\qd} \Big]\frac{\ln(T)}{(T+1)^\beta}.
\end{align*}

Finally, squaring both sides in the above expression gives the desired bound in Lemma~\ref{thm: avg.value.converge}. \hfill \scalebox{0.8}{$\blacksquare$}

\subsection{Proofs of Remaining Technical Results}
\label{s:layer.3.proofs}

 Lemmas~\ref{lemma:AandB} and \ref{thm: PR.bounds} are derived here. 

\emph{Proof of \textbf{Lemma~\ref{lemma:AandB}}}.
We prove each statement individually.

(i). \textbf{The bound in \eqref{e:Wt.Bd}} holds since
\begin{align*}
    \lvert\bE_{t - \tau} \Wt^{(i)}\rvert \overset{(a)}{\leq} {} & \sum_{s, a} |\bP(s_t^i = s|s^i_{t - \tau})   - d^{\mu}_i(s)|\ \mu(a|s)  |\cR_i(s, a)| \\
    \overset{(b)}{\leq} {} & \Rm \sum_{s, a} |\bP(s_t^i = s|s^i_{t - \tau})   - d^{\mu}_i(s)|\ \mu(a|s) \\
    \overset{(c)}{\leq} {} & \Rm \sum_{s} |\bP(s_t^i = s|s^i_{t - \tau})   - d^{\mu}_i(s)|   \\
    \overset{(d)}{\leq} {} & \Cerg \Rm \alpha^\tau 
\end{align*}
where (a) follows from $\Wt^{(i)}$ and $r_i^*$'s definition in \eqref{e:Wt.defn} and Section~\ref{sec:setup}, respectively; (b) follows from Assumption~\ref{a: reward.bound};  (c) holds since $\sum_{a} \mu(a|s) = 1$; while (d) follows from \eqref{e:ergodicity}.

(ii). Consider \textbf{the bound in \eqref{eq:bound_A}}. For all $t > 0$, we have
    \begin{align*}
        \bE \Wt^{2} & = \bE\bigg(\frac{1}{N}\sum_{i=1}^{N} \Wt^{(i)}\bigg)^{2} \\
        & = \frac{1}{N^{2}}\sum_{i=1}^{N} \bE(\Wt^{(i)})^{2} + \frac{2}{N^{2}}\sum_{i<j} \bE \Wt^{(i)} \Wt^{(j)}
        \\
        & \overset{(a)}{\leq} \frac{1}{N^{2}}\sum_{i=1}^{N} \bE(\Wt^{(i)})^{2} + \frac{2}{N^{2}}\sum_{i<j} \lvert\bE \Wt^{(i)}\rvert \lvert\bE \Wt^{(j)}\rvert  
        \\
        & \overset{(b)}{\leq} \frac{1}{N^{2}}\sum_{i=1}^{N} \bE(\Wt^{(i)})^{2} + \Cerg^{2}\Rm^{2}\alpha^{2t} \\
        & \overset{(c)}{\leq} \frac{4\Rm^{2}}{N} + \Cerg^{2}\Rm^{2}\alpha^{2t},
    \end{align*}
where (a) holds since $W^{(i)}_{t+1}$ and $W^{(j)}_{t+1}$ are independent $\forall i\neq j \in [N],$ (b) holds due to \eqref{e:Wt.Bd}, while (c) is true since, from \eqref{e:Wt.defn}, we have the trivial bound $|\Wt^{(i)}| \leq 2 \Rm.$

Now, for $t \geq \tstr,$ we have $t > 2 \tau_t$ and $\alpha^{\tau_t}  \leq 1/(t + 1)^2.$ Hence, $\alpha^{4t} \leq \alpha^{2 \tau_t} \leq 1/(t + 1)^4.$ The desired result follows. 

(iii). To obtain \textbf{the bound in~\eqref{eq:bound_B}}, note that, $\forall 0 \leq \tau \leq t,$
\begin{align}
        \lvert 2\bE  \rho_{t} &\Wt\rvert \nonumber \\
        \leq {} & \lvert2\bE\rho_{t-\tau}\Wt\rvert + \lvert2\bE(\rho_{t} - \rho_{t-\tau})\Wt \rvert  \nonumber \\
        \overset{(a)}{=} {} & \lvert2\bE\rho_{t-\tau}\bE_{t-\tau}\Wt\rvert + \lvert2\bE(\rho_{t} - \rho_{t-\tau})\Wt \rvert  \nonumber \nonumber  \\
        \leq {} & 2\bE\lvert\rho_{t-\tau}\rvert \lvert\bE_{t-\tau}\Wt\rvert + 2\bE\lvert\rho_{t} - \rho_{t-\tau}\lvert \lvert \Wt \rvert,
        \label{eq:term_B}
\end{align}
where (a) uses the iterated expectation law and $\rho_t \in \cF_t.$

Next, we bound the two terms in \eqref{eq:term_B}. Observe that
\begin{align}
        2\bE\lvert\rho_{t-\tau}\rvert  & \lvert\bE_{t-\tau}\Wt\rvert \nonumber \\
        \overset{(a)}{\leq} & \frac{1}{t+1}\bE\rho_{t-\tau}^{2} + (t+1)\bE\lvert\bE_{t-\tau}\Wt\rvert^{2} \nonumber
        \\
        \overset{(b)}{\leq} & \frac{1}{t+1}\bE\rho_{t-\tau}^{2} +(t+1)\Cerg^{2}\Rm^{2}\alpha^{2 \tau}, \label{eq:temp_bound_1}
\end{align}
where (a) holds due to the Cauchy-Schwarz inequality, while (b) follows from \eqref{e:Wt.Bd}. Similarly, 
\begin{align}
        2\bE\lvert & \rho_{t} -  \rho_{t-\tau}\lvert \lvert \Wt \rvert \nonumber \\
        \leq {} & (t+1)\bE(\rho_{t} - \rho_{t-\tau})^{2} + \frac{1}{t+1}\bE \Wt^{2} 
        \nonumber \\
        \leq {} & (t+1)\bE(\rho_{t} - \rho_{t-\tau})^{2} + \frac{4\Rm^{2}}{N(t+1)} + \frac{\Cerg^{2}\Rm^{2}}{(t+1)^5},
        \label{eq:temp_bound_2}
\end{align}
where the last inequality follows from \eqref{eq:bound_A}. Substituting \eqref{eq:temp_bound_1} and \eqref{eq:temp_bound_2} in \eqref{eq:term_B} and noting that $\bE \rho_{t-\tau}^2 \leq 2\bE \rho_t^2 + 2 \bE(\rho_t - \rho_{t-\tau})^2$ and $(t + 1) + 2/(t + 1) \leq t + 3$ for $t \geq 0$ then gives
\begin{multline*}
        \lvert 2\bE \rho_{t}\Wt\rvert \leq \frac{2}{t+1}\bE\rho_{t}^{2} + (t+3)\bE(\rho_{t} - \rho_{t-\tau})^{2} \\ + \frac{4\Rm^{2}}{N(t+1)} + \frac{\Cerg^{2}\Rm^{2}}{(t + 1)^5} + (t+1)\Cerg^{2}\Rm^{2}\alpha^{2\tau}.
\end{multline*}
Now we choose $\tau=\tau_{t}$ so that $\alpha^{2\tau_t} \leq 1/(t + 1)^4.$ Separately, we have $(t+3)\leq 3(t+1)$ for $t \geq 0.$ Consequently, $\forall t \geq \tstr,$ 
\begin{multline}
    \lvert 2\bE\rho_{t}\Wt\rvert \leq \frac{2}{t + 1}\bE\rho_{t}^{2} + 3(t+1)\bE(\rho_{t} - \rho_{t -\taut})^{2} 
    \\
    + \frac{4\Rm^{2}}{N(t+1)} + 
    % \frac{\Cerg^{2}\Rm^{2}}{t+1}\alpha^{2t} + 
    \frac{2\Cerg^{2}\Rm^{2}}{(t+1)^{3}}. 
    \label{eq:temp_cross}
\end{multline}

We now bound  $\bE(\rho_{t} - \rho_{t -\taut})^{2}$ for $t \geq \tstr.$ From \eqref{e: update.diff.rule}, a simple induction argument shows that 
\[
    \rho_t - \rho_{t - \tau_t} = \frac{-\tau_t}{t} \rho_{t - \tau_t} + \frac{1}{t} \sum_{j = t -\tau_t}^{t - 1} W_{j + 1}.
\]
Using $|\rho_{t - \tau_t}| \leq |\rho_t| + |\rho_t - \rho_{t - \tau_t}|,$ we then get
\[
    |\rho_t - \rho_{t - \tau_t}| \leq \frac{\tau_t}{t - \tau_t} |\rho_t| + \frac{1}{t - \tau_t} \sum_{j = t -\tau_t}^{t - 1} |W_{j + 1}|.
\]
Now, by squaring, taking expectation, and using $(\sum_{i = 1}^m a_i)^2 \leq m \sum_{i = 1}^m a_i^2$ for any $a_1, \ldots, a_m \in \bR$ and $m \geq 1,$ we have
\[
    \bE (\rho_t - \rho_{t - \tau_t})^2     \leq \frac{2 \tau_t^2}{(t - \tau_t)^2} \bE \rho^2_t + \frac{2 \tau_t}{(t - \tau_t)^2} \sum_{j = t -\tau_t}^{t - 1} \bE W^2_{j + 1}.
\]
For all $t \geq \tstr,$ substituting \eqref{eq:bound_A} in the above inequality and noting that $\sup_{t - \tau_t \leq j \leq t - 1} (j + 1)^{-4} \leq (t - \tau_t)^{-4}$ then shows 
\begin{multline}
\label{e:rho.t -rho.t-tau.t.sq.Bd}
     \bE (\rho_t - \rho_{t - \tau_t})^2\\
    \leq \frac{2 \tau_t^2}{(t - \tau_t)^2} \bE \rho^2_t 
    + \frac{8 \Rm^2 \tau_t^2}{N (t - \tau_t)^2} + \frac{2  \Cerg^2 \Rm^2 \tau_t^2}{(t - \tau_t)^6}. 
\end{multline}

% \begin{align}
%     \bE (\rho_t & - \rho_{t - \tau_t})^2 \nonumber \\
%     %
%     \overset{(a)}{\leq} {} & \frac{2 \tau_t^2}{(t - \tau_t)^2} \bE \rho^2_t 
%     %
%     + \frac{8 \Rm^2 \tau_t^2}{N (t - \tau_t)^2} + \frac{2  \Cerg^2 \Rm^2 \tau_t^2}{(t - \tau_t)^6} \nonumber \\
%     %
%     \overset{(b)}{\leq} {} & \frac{2 \tau_t^2}{(t - \tau_t)^2} \bE \rho^2_t + \frac{8 \Rm^2 \tau_t}{N (t - \tau_t)^2} + \frac{2 \tau_t \Cerg^2 \Rm^2 \alpha^{2\tau_t}}{(t - \tau_t)^2} \nonumber\\
%     %
%     \overset{(c)}{\leq} {} &  \frac{2 \tau_t^2}{(t - \tau_t)^2} \bE \rho^2_t + \frac{8 \Rm^2 \tau_t}{N (t - \tau_t)^2} + \frac{2 \tau_t\Cerg^2 \Rm^2}{(t - \tau_t)^6} , \label{e:rho.t -rho.t-tau.t.sq.Bd}
% \end{align}
%

Finally, we have that
\begin{align}
    \lvert 2 \bE \rho_{t} & \Wt\rvert \nonumber \\
    %
    % \leq {} & \frac{2}{(t+1)}\bE\rho_{t}^{2} + 3(t+1)\bE(\rho_{t} - \rho_{t -\taut})^{2} \nonumber 
    % \\
    % {} & + \frac{4\Rm^{2}}{N(t+1)} + \frac{2 \Cerg^{2}\Rm^{2}}{(t+1)^3} \nonumber \\
    %
    \overset{(a)}{\leq} {} &  \frac{2}{(t+1)}\bE\rho_{t}^{2} + \frac{6 (t + 1)\tau_t^2}{(t - \tau_t)^2} \bE \rho^2_t \nonumber  \\
    {} & + \frac{24 \Rm^2 (t + 1) \tau_t^2}{N (t - \tau_t)^2} +  \frac{4\Rm^{2}}{N(t+1)} \nonumber  \\
    {} & + \frac{6  \Cerg^2 \Rm^2 (t + 1) \tau_t^2}{(t - \tau_t)^6} +  \frac{2\Cerg^{2}\Rm^{2}}{(t+1)^{3}}  \nonumber  \\
    \overset{(b)}{\leq} {} & \frac{(t + 1)}{(t - \tau_t)^2} \left[8 \taut^2 \bE \rho_t^2 + \frac{28 \taut^2 \Rm^2 }{N} + \frac{8\Cerg^{2}\Rm^{2}}{(t- \tau_t)^{2}} \right] \nonumber \\
    \overset{(c)}{\leq} {} & \frac{(t + 1) }{(t - \tau_t)^2} \left[8 \taut^2 \bE \rho_t^2 + C_{\rho,\text{W}} \left(\frac{\taut^2}{N} + \frac{1}{(t- \tau_t)^{2}} \right) \right],
\end{align}
where (a) follows by substituting \eqref{e:rho.t -rho.t-tau.t.sq.Bd} in \eqref{eq:temp_cross}, (b) follows by combining similar terms and using the inequalities $\tau_t/(t - \tau_t) \leq 1,$ $t \geq \tstr^{(1)},$ and $1 \leq \tau_t^2,$ $\forall t \geq 0$ (which also implies $t + 1 \geq t - \tau_t$); while (c) follows using  $C_{\rho,\text{W}}$'s definition from Table~\ref{tab: constants} and since $t - \taut \geq 1$\ $\forall t > \tstr$. 

The desired relation in \eqref{eq:bound_B} now follows.

(iv). Consider \textbf{the bound in~\eqref{eq:bound_D}}. Since $1 - x \leq e^{-x}$ for any $x \in \bR,$ it follows that, for any $\tstr\leq t_{1}<t_{2}$ 
\begin{align*}         
    \Grho_{t_{1}:t_{2}} 
    \leq {} & \exp\Big( \sum_{t=t_{1}}^{t_{2}-1} \frac{9\taut^{2}}{(t-\taut)^{2}}  \Big)\cdot \exp\Big( -\sum_{t=t_{1}}^{t_{2}-1} \frac{2}{t+1} \Big) \\
    \leq {} & \xi_{\rho} \cdot\exp\Big( 2\ln\Big[ \frac{t_{1}+1}{t_{2}+1}\Big]\Big) \\
    = {} & \xi_{\rho}\Big(\frac{t_{1}+1}{t_{2}+1}\Big)^{2}.
    \end{align*} 

Next consider \textbf{the bound in~\eqref{eq:bound_E}}. For all $t \geq \tstr,$ we have
\begin{align*}
    \sum_{t=\tstr}^{T-1} & \Big( \frac{\taut^{2}}{N (t - \taut)^{2}} + \frac{1}{(t - \tau_t)^4}\Big) G_{t+1:T}^{\rho} \\
    \overset{(a)}{\leq} {} & \xi_\rho \sum_{t=\tstr}^{T-1} \Big( \frac{\tau_T^2}{N (t - \taut)^{2}} + \frac{1}{(t - \tau_t)^4}\Big) \frac{(t + 2)^2}{(T + 1)^2} \\
    \overset{(b)}{\leq} {} & \xi_\rho \sum_{t=\tstr}^{T-1} \Big( \frac{4\tau_T^2}{N (t + 2)^{2}} + \frac{16}{(t + 2)^4}\Big) \frac{(t + 2)^2}{(T + 1)^2} \\
    \overset{(c)}{\leq} {} &  \xi_\rho  \Big(\frac{4\tau_T^2}{N(T + 1)} + \frac{16}{(T + 1)^2(\tstr + 1)}\Big),
\end{align*}
where (a) follows since $\tau_t^2 \leq \tau_T^2,$ (b) holds since $t - \tau_t \geq (t + 2)/2$ for $t \geq \tstr^{(1)},$ while (c) holds since $\sum_{t = \tstr}^{T - 1} \frac{1}{(t + 2)^2} \leq \int_{\tstr - 1}^{T - 1} \frac{1}{(x + 2)^2} \textnormal{d}x \leq \frac{1}{\tstr + 1}.$ \hfill \scalebox{0.8}{$\blacksquare$}
\\

It remains to prove \textbf{Lemma~\ref{thm: PR.bounds}}, for which we make use of Lemma~\ref{thm: mat.prod.bound} and \ref{thm: technical.bounds}.  For every $0\leq t_1<t_2,$ let
\begin{align*}
    &\Gdel_{t_1:t_2} := \prod_{t=t_1}^{t_2-1}(I-\bt A), \quad M_{t_1: t_2} := \beta_{t_1}\sum_{t=t_1+1}^{t_2-1}\Gdel_{t_1+1:t},
    \\
    & \quad\qquad\qquad S_{t_1:t_2} := \sum_{t=t_1}^{t_2-1}\bt\Gdel_{t+1:t_2}\tilde{A}_t\Gdel_{t_1:t}.
\end{align*}

% $\Gdel_{t_{1},t_{2}} := \prod_{t=t_1}^{t_2-1}(I-\bt A),$ $M_{t_1, t_2} := \beta_{t_1}\sum_{t=t_1+1}^{t_2-1}\Gdel_{t_1+1:t},$ and $S_{t_1:t_2} := \sum_{t=t_1}^{t_2-1}\bt\Gdel_{t+1:t_2}\tilde{A}_t\Gdel_{t_1:t}.$ 

\begin{lemma}\label{thm: mat.prod.bound} 
    Choose and fix a $\lambda \in (0, \lmin(A + A^\top)).$ Let the constants $\CG,\Clambda, \KG,$ and $\CS$ be as defined in Table~\ref{tab: constants}. Then, the following holds for every $0\leq k<T,$
    \begin{enumerate}
        \item $\|\Gdel_{k:T}\|\leq \CG \exp{\big(-\lambda\sum_{t=k}^{T-1}\bt \big)};$
        \item $\sum_{t=k}^{T-1}\bt\gt\exp{\big(-\lambda\sum_{s=t+1}^{T-1}\beta_s\big)} \leq \Clambda \gamma_{T-1};$
        \item $\|M_{k:T}\| \leq \KG;$
        \item $\bE\|S_{k:T}\omega\|^2\leq$ $\CS^2 \exp{\big(-2\lambda\sum_{s=k}^{T-1}\beta_s\big)}\bE\|\omega\|^2\bk^2(T-k),$
    \end{enumerate}
    where $\omega$ is any $\cF_k$-measurable random variable, and $\gt:=\bt^n/(t+1)^m,$ for $n,m\geq 0.$
\end{lemma}

\begin{lemma}\label{thm: technical.bounds}
    For every $0\leq t_1<t_2$ and $T>\tstr,$ the following statements hold. 
    \begin{align}
        \text{(i). } & \sum_{t=\tstr}^{T-1}\bE^{1/2}\|\Gdel_{\tstr:t}L^{(1)}_{\tstr}\|^2 \leq 2\CG\Clambda\xi_G\xi_M
        \label{e: sum.L1.0}
        \\\nonumber
        \text{(ii). } & \bE^{1/2}\bigg\|\sum_{k=\tstr}^{T-2}\rho_kM_{k: T-1}\vhk\bigg\|^2
        \\
        & \qquad \quad \leq  \frac{\KG}{\sqrt{2}}\left[ 2C^{1/2}_{r,\lin}\frac{\tauT\sqrt{T}}{\sqrt{N}} + C^{1/2}_{r,\qd}\ln{(T)} \right]
        \label{e: sum.L1.1}
        \\
        \text{(iii). } & \bE^{1/2}\bigg\|\sum_{k=\tstr}^{T-2}M_{k: T-1} \zhk\bigg\|^2 
        \nonumber\\
        & \leq 4\KG\xi_M\sqrt{\frac{\tauT T}{N}} + \KG\xi_M\Cealp
        \label{e: sum.L1.2}
        \\
        \text{(iv). } & \bE^{1/2}\bigg\|\sum_{k=\tstr}^{T-2}\bk \rho_k S_{k+1:T}\vhk\bigg\|^2 
        \nonumber\\
        & \qquad \quad \leq \frac{\CS\Clambda}{\sqrt{2}}\bigg[ \frac{\tauT C^{1/2}_{r,\lin}}{\sqrt{N(T+1)}} + \frac{C^{1/2}_{r,\qd}}{T+1} \bigg]
        \label{e: sum.rho}
        \\
        \text{(v). } & \bE^{1/2}\bigg\|\sum_{k=\tstr}^{T-2}\bk S_{k+1:T}\zhk\bigg\|^2 
        \leq \xfl^{(1)} \tauT\bT. \label{e: sum.S.z}
    \end{align}
    The constants above 
    are as defined in Table~\ref{tab: constants}.
\end{lemma}

With the above two lemmas, we are ready to prove Lemma~\ref{thm: PR.bounds}
\\

\emph{Proof of \textbf{Lemma~\ref{thm: PR.bounds}}}. To prove \textbf{Statement 1} in Lemma~\ref{thm: PR.bounds}, notice from Step~\ref{alg:PR} of Algorithm~\ref{AvgFedTD} that
\begin{align*}
    \|\theta_t\| \leq {} & 
    \|\theta_0\| + \sum_{k = 0}^{t-1} \frac{\beta_k}{N} \sum_{i \in [N]} \|\delta_{k + 1}^i\| \\
    \leq {} & \|\theta_0\| + \sum_{k = 0}^{t-1} \beta_k [2\Rm + 2 \|\theta_k\|].
\end{align*}
Then, applying the discrete Gronwall's inequality \cite[Appendix B]{borkar2009} and using $\beta_k \leq 1$ shows that
\[
    \|\theta_t\|\leq (\|\theta_0\| + 2t\Rm) e^{2 t}.
\]
Hence, $\forall t\leq \tstr,$
\begin{align*}
    \|\Delta_t\| = {} & \|\theta_t - \thS\| 
    \\
    \leq {} & \|\thS\| + (\|\theta_0\| + 2t\Rm) e^{2 t}.
    \\
    \leq {} & \|\thS\| + (\|\theta_0\| + 2\tstr\Rm) e^{2 \tstr} =: \CD.
\end{align*}

\textbf{Statement 2} can be obtained following arguments similar to \cite[Lemma 7]{durmus2024finite}.
\\

To prove \textbf{Statement 3}, note that, for all $t > \tstr,$
\begin{equation}\label{e: L1.form.star}
        L^{(1)}_t =  \Gdel_{\tstr:t}L^{(1)}_{\tstr} + \sum_{k=\tstr}^{t-1}\bk\Gdel_{k+1:t}\left(-\rho_k\vhk + \zhk \right).
\end{equation}
Thus, we have
\begin{align*}
    & \sum_{t=\tstr}^{T-1}L^{(1)}_t 
    \\
    & \leq  \sum_{t=\tstr}^{T-1}\Gdel_{\tstr:t}L^{(1)}_{\tstr} 
    + \sum_{t=\tstr}^{T-1}\sum_{k=\tstr}^{t-1}\bk\Gdel_{k+1:t}\left(-\rho_k\vhk + \zhk \right)
    \\
    & \overset{(a)}{=} \sum_{t=\tstr}^{T-1}\Gdel_{\tstr:t}L^{(1)}_{\tstr} 
    - \sum_{k=\tstr}^{T-2}\rho_k M_{k: T-1}\vhk
    + \sum_{k=\tstr}^{T-2}M_{k:T-1} \zhk,
\end{align*}
where $(a)$ is obtained by interchanging the order of the double summation and using the definition of $M_{k:T-1}.$ The desired bound now follows as each term on the r.h.s is bounded as in \eqref{e: sum.L1.0}, \eqref{e: sum.L1.1}, and \eqref{e: sum.L1.2}.
\\

Finally, we prove \textbf{Statement 4}. Expanding the update rule for $(L^{(2)}_t)$ in \eqref{e: L.update} and substituting \eqref{e: L1.form.star} gives us
    \begin{align}\label{e: Ltwo.decomp}
        & L^{(2)}_T =\Gdel_{\tstr:T}L^{(2)}_{\tstr} -\sum_{t=\tstr}^{T-1}\bt\Gdel_{t+1:T}\tilde{A}_tL^{(1)}_t 
        \nonumber \\
        & = \Gdel_{\tstr:T}L^{(2)}_{\tstr} - S_{\tstr:T}L^{(1)}_{\tstr}
        \nonumber \\
        & \qquad \qquad + \sum_{k=\tstr}^{T-2}\bk \rho_kS_{k+1:T}\vhk - \sum_{k=\tstr}^{T-2}\bk S_{k+1:T}\zhk.
    \end{align}
     To derive the bound in \eqref{e: order.L2} we need to bound the four terms on the r.h.s of \eqref{e: Ltwo.decomp}. The last two terms are bounded using \eqref{e: sum.rho} and \eqref{e: sum.S.z}, respectively. For the first term in \eqref{e: Ltwo.decomp}, we take the update rule for $(L^{2}_T)$ in~\eqref{e: L.update} to get for $T > 0,$
    \begin{align}\label{e: Ltwo.bound.uniform}
        & \|L^{(2)}_T\| 
        \leq  \sum_{t=0}^{T-1}\bt\|\Gdel_{t+1:T} \|\|L^{(1)}_t \| 
        \nonumber\\
        & \overset{(a)}{\leq} \CG\sum_{t=0}^{T-1}\bt\Big( e^{-\lambda\sum_{s=t+1}^{T-1}\beta_s} \Big)\|L^{(1)}_t \| 
        \nonumber\\
        & \overset{(b)}{\leq} \CG B_L\sum_{t=0}^{T-1}\bt\Big( e^{-\lambda\sum_{s=t+1}^{T-1}\beta_s} \Big) \overset{(c)}{\leq} \CG\Clambda B_L,
    \end{align}
    where $(a)$ and $(c)$ are obtained from Lemma~\ref{thm: mat.prod.bound}, whereas $(b)$ follows from \eqref{e: L1.star.bound}. Therefore, the second term in \eqref{e: Ltwo.decomp} is bounded as follows:
    \begin{align}
        \bE^{1/2}\|&\Gdel_{\tstr:T} L^{(2)}_{\tstr}\|^2\nonumber \\
        & \overset{(a)}{\leq} \CG \Big(e^{-\lambda\sum_{s=\tstr}^{T-1}\beta_s}\Big)\bE^{1/2}\|L^{(2)}_{\tstr}\|^2
        \nonumber \\
        & \overset{(b)}{\leq}  \CG^2\Clambda B_L \Big(e^{-\lambda\sum_{s=\tstr}^{T-1}\beta_s}\Big)
        \nonumber \\
        & \overset{(c)}{\leq} \CG^2\Clambda B_L e^{\frac{\lambda\tstr}{(1-\beta)}}\Big( e^{-\frac{\lambda}{(1-\beta)}T^{(1-\beta)}}\Big)
        \nonumber\\
        & \overset{(d)}{\leq}  \bigg[e^{\frac{\lambda\tstr}{(1-\beta)}}\Big(T e^{-\frac{\lambda}{(1-\beta)}T^{(1-\beta)}}\Big)\bigg]\frac{\CG^2\Clambda B_L}{T}
        \nonumber \\
        & \overset{(e)}{\leq} \bigg[\Big(\frac{e^{\lambda\tstr}}{e\lambda}\Big)^{1/(1-\beta)}\bigg]\frac{\CG^2\Clambda B_L}{T},
    \end{align}
    where $(a)$ follows from Lemma~\ref{thm: mat.prod.bound}, $(b)$ uses \eqref{e: Ltwo.bound.uniform}, $(c)$ is obtained by taking $\sum_{s=\tstr}^{T-1}\beta_s\geq \frac{1}{(1-\beta)}\big[T^{(1-\beta)} - \tstr^{(1-\beta)} \big],$ $(d)$ is obtained by multiplying and dividing by $T,$ and $(e)$ follows since, using calculus, we can show that $x^n\exp{\big(-\lambda x^{(1-\beta)}/(1-\beta)}\big)\leq \big( n/e\lambda\big)^{n/(1-\beta)}.$
    For the second term in \eqref{e: Ltwo.decomp}, we proceed as follows: use Lemma~\ref{thm: mat.prod.bound} and  to obtain
    \begin{align}
        \bE^{1/2}\|S_{\tstr:T}L^{(1)}_{\tstr}\|^2 
        & \overset{(a)}{\leq} \CS \beta_{\tstr} \sqrt{T-\tstr}\Big(  e^{-\sum_{s=\tstr}^{T-1}\beta_s}\Big) \bE\|L^{(1)}_{\tstr}\|^2
        \nonumber\\
        & \overset{(b)}{\leq} \CS B_L\Big(T e^{-\sum_{s=\tstr}^{T-1}\beta_s}\Big)
        \nonumber\\
        & \overset{(c)}{\leq} \bigg[e^{\frac{\lambda \tstr}{(1-\beta)}}\Big(T^2 e^{-\frac{\lambda}{(1-\beta)}T^{(1-\beta)}}\Big)\bigg]\frac{\CS B_L }{T}
        \nonumber\\
        & \leq \bigg[\Big(\frac{2e^{\frac{\lambda\tstr}{2}}}{e\lambda} \Big)^{2/(1-\beta)}\bigg]\frac{\CS B_L }{T},
    \end{align}
    where $(a)$ uses Lemma~\ref{thm: mat.prod.bound}, $(b)$ combines \eqref{e: L1.star.bound} and the fact that $\beta_{\tstr}\leq 1$ and $\sqrt{T-\tstr}< T,$ whereas $(c)$ is obtained by multiplying and dividing by $T,$ and lower bounding the $\sum_{s=\tstr}^{T-1}\beta_{s}$ by $\frac{1}{(1-\beta)}\big[ T^{(1-\beta)} - \tstr^{(1-\beta)} \big].$

Finally, we bound $\bE^{1/2}\|L^{(3)}_T\|^2$ given in \eqref{e: order.L3}.  Note that
\begin{align}\label{e: Lthree.norm.bound}
    \bE&\|\Gamma_{t+1:T}L^{(2)}_t\|^2 
    = \bE\Big[[L^{(2)}_t ]^\top\big[ \Gamma_{t+1:T}^\top\Gamma_{t+1:T}\big][L^{(2)}_t]\Big] 
    \nonumber\\
    &= \bE\Big[[L^{(2)}_t ]^\top\bE_t\big[ \Gamma_{t+1:T}^\top\Gamma_{t+1:T}\big][L^{(2)}_t]\Big] 
    \nonumber \\
    &\leq \bE\|L^{(2)}_t\|^2\bE_t\|\Gamma_{t+1:T}\|^2
\end{align}
The following expression follows from the update rule for $(L^{(3)}_T)$ in \eqref{e: L.update}:
    \[
        L^{(3)}_T = \sum_{t=0}^{T-1}\Gamma_{t+1:T}\bt\tilde{A}_t L^{(2)}_t.
    \]
    Now applying the triangle inequality, we have
    \begin{align*}
        & \bE^{1/2}\| L^{(3)}_T\|^2 \overset{(a)}{\leq} 2\sum_{t=0}^{T-1}\bt \bE^{1/2}\|\Gamma_{t+1:T}L^{(2)}_t\|^2
        \\
        & \overset{(b)}{\leq} 4\sum_{t=0}^{T-1}\bt \bE^{1/2}\big[ \|L^{(2)}_t\|^2 \bE_t\|\Gamma_{t+1:T}\|^2\big] 
        \\
        & \overset{(c)}{\leq} 4C_\Gamma\sum_{t=0}^{T-1}\bt\Big(e^{-\lambda_h\sum_{i=0}^{\floor{(T-t)/h}}\beta_{t+ih}} \Big) \bE^{1/2}\|L^{(2)}_t\|^2
        \\
        & \overset{(d)}{\leq} 4C_\Gamma\sum_{t=0}^{T-1}\bt\Big(e^{-\lambda_h\sum_{i=0}^{\floor{(T-t)/h}}\beta_{t+ih}} \Big) f_L(t)
        \\
        & \leq 4C_\Gamma \bigg[\sup_{0\leq t<T} f_L(t)\Big(e^{-\frac{\lambda_h}{2}\sum_{i=0}^{\floor{(T-t)/h}}\beta_{t+ih}} \Big) \bigg]
        \\
        & \qquad \qquad \qquad \times \sum_{t=0}^{T-1}\bt\Big(e^{-\frac{\lambda_h}{2}\sum_{i=0}^{\floor{(T-t)/h}}\beta_{t+ih}} \Big) 
        \\
        & \overset{(e)}{\leq} 4C_\Gamma f_L(T) \sum_{t=0}^{T-1}\bt\Big(e^{-\frac{\lambda_h}{2}\sum_{i=0}^{\floor{(T-t)/h}}\beta_{t+ih}} \Big), 
    \end{align*}
    where $(a)$ uses $\|\Atil\|\leq 4,$ $(b)$ uses \eqref{e: Lthree.norm.bound}, $(c)$ follows from Statement 1 proved earlier, and $(d)$ follows since $\bE\|L^{(2)}_T\|^2\leq f_L(t).$ Lastly, $(e)$ follows since $f_L(t)\exp{\big(-\frac{\lambda_h}{2}\sum_{i=0}^{\floor{(T-t)/h}}\beta_{t+ih}\big)}$ is increasing in $t.$ Further, using a Riemann sum-based argument as in Lemma~\cite[Lemma~4.3]{dalal2018finite}, we can show that
    \[
    \sum_{t=0}^{T-1}\bt\Big(e^{-\frac{\lambda_h}{2}\sum_{i=0}^{\floor{(T-t)/h}}\beta_{t+ih}} \Big) \leq \frac{2h}{\lambda_h}.
    \]
    This completes the proof of Lemma~\ref{thm: PR.bounds}. 
    \hfill \scalebox{0.8}{$\blacksquare$}
    \\
%%%%%%%%%%%%%%%%%%%%%%%%%%%%%%%%%%%%%%%%%%%%%%%%%%%%%%

At last, we provide the proofs for the intermediate technical lemmas~\ref{thm: technical.bounds} and \ref{thm: mat.prod.bound}.

\emph{Proof of \textbf{Lemma~\ref{thm: mat.prod.bound}}}
    \textbf{Statement 1} follows directly from~\cite [Lemma~4.1]{dalal2018finite}. Further, \textbf{statement 2} follows form~\cite[Lemma~4.3]{dalal2018finite}, which shows that
    \[
        \sum_{t = k}^{T - 1} \bt e^{-\frac{\lambda}{2} \sum_{s = t + 1}^{T - 1}  \beta_t } \leq \Big(\frac{2e^{\lambda/2}}{\lambda}\Big).
    \]
    
    To prove \textbf{statement 3}, we use the following decomposition:
    \begin{multline*}
        \sum_{t = k}^{T - 1} \bt\gt e^{-\lambda \sum_{s = t + 1}^{T - 1}  \beta_j } 
        \\
        \leq \bigg(\max_{k\leq t<T} \gt e^{-\frac{\lambda}{2}\sum_{s=t+1}^{T-1}\beta_s} \bigg)\bigg(\sum_{t = k}^{T - 1} \bt e^{-\frac{\lambda}{2} \sum_{s = t + 1}^{T - 1}  \beta_t } \bigg).
    \end{multline*}
    Since $\gt e^{\frac{\lambda}{2}\sum_{s=t+1}^{T-1}\beta_s}$ is increasing in $t,$ the first factor in the above expression is $\gamma_{T-1}.$ 
    
    For proving \textbf{statement 4}, note that
    \[
        \sum_{s=k+1}^{t-1}\beta_s \geq \int_{k+1}^{t}\frac{dx}{(1+x)^\beta} \geq \frac{t^{(1-\beta)} - (k+2)^{(1-\beta)}}{(1-\beta)}
    \]
    and hence
    \begin{align*}
        \sum_{t=k+1}^{T-1} & e^{-\lambda\sum_{s=k+1}^{t-1}\beta_s} 
        \leq e^{\frac{\lambda}{(1-\beta)}(k+2)^{(1-\beta)}}\sum_{t=k+1}^{T-1}e^{-\frac{\lambda}{(1-\beta)}t^{(1-\beta)}}
        \\
        & \leq e^{\frac{\lambda}{(1-\beta)}(k+2)^{(1-\beta)}}\int_{k}^{T-1} e^{-\frac{\lambda}{(1-\beta)}x^{(1-\beta)}}dx
        \\
        & \leq e^{\frac{\lambda}{(1-\beta)}(k+2)^{(1-\beta)}}\int_k^\infty e^{-\frac{\lambda}{(1-\beta)}x^{(1-\beta)}}dx.
    \end{align*}
    Next, we apply the change of variable $u=\frac{\lambda}{(1-\beta)} x^{(1-\beta)}$ to the above integral and get
    \[
        \int_k^\infty e^{-\frac{\lambda}{(1-\beta)}x^{(1-\beta)}}dx = \frac{a}{\eta^a}\int_{\eta k^{1/a}}^\infty e^{-u}u^{a-1}du,
    \]
    where $\eta=\frac{\lambda}{(1-\beta)}$ and $a:= \frac{1}{(1-\beta)}.$ We use the following result for $v\geq 0:$
    \[
        \int_v^\infty e^{-u}\ u^{a-1}\ du \leq e^{-v}v^{a-1}\ e.
    \]
    Setting $v = \eta k^{1/a},$ we have
    \begin{align*}
        \int_k^\infty e^{-\frac{\lambda}{(1-\beta)}x^{(1-\beta)}}dx 
        & \leq \frac{ea}{\eta^a}e^{-\eta k^{1/a}}\eta^{(a-1)}k^{(1-1/a)}
        \\
        & = \frac{ek^\beta}{\lambda}e^{-\frac{\lambda}{(1-\beta)}k^{(1-\beta)}}.
    \end{align*}
    Therefore, 
    \begin{align*}
        \bk\sum_{t=k+1}^{T-1}&  e^{-\lambda\sum_{s=k+1}^{t-1}\beta_s} 
        \leq \Big[\frac{e k^\beta\bk}{\lambda}\Big] \ e^{\frac{\lambda}{(1-\beta)}\left[ (k+2)^{(1-\beta)} - k^{(1-\beta)} \right]}.
    \end{align*}
    To obtain the desired bound, we use $k^\beta\bk<1$ and the fact that $\left[ (k+2)^{(1-\beta)}- k^{(1-\beta)} \right] \leq 2(1-\beta)(k+2)^{-\beta}< 2(1-\beta).$

    Finally, we prove \textbf{statement 5}. It follows from the definition of $S_{k:T},$ that
    \begin{multline*}
        \bE\|S_{k:T} \omega\|^2 
        = \sum_{t=k}^{T-1}\bt^2\bE\|\Gdel_{t+1:T}\Atil\Gdel_{k:t}\ \omega\|^2
        \\
        + 2\sum_{k\leq s<t}^{T-1}\beta_s\bt\bE\Big\langle \Gdel_{s+1:T}\tilde{A}_s\Gdel_{k:s}\ \omega, \Gdel_{t+1:T}\Atil\Gdel_{k:t}\ \omega\Big\rangle.
    \end{multline*}
    The first term in the above expression is bounded as follows: 
    \begin{align}\label{e: S.omega.diag}
        \sum_{t=k}^{T-1}&\bt^2\|\Gdel_{t+1:T}\Atil\Gdel_{k:t} \ \omega\|^2 
        \nonumber\\
        & \overset{(a)}{\leq} 16\CG^2\sum_{t=k}^{T-1}\bt^2 \Big(e^{-2\lambda\sum_{s=k}^{t-1}\beta_s} \Big)\Big(e^{-2\lambda\sum_{s=t+1}^{T-1}\beta_s}\Big)\bE\|\omega\|^2
        \nonumber\\
        & \overset{(b)}{=} 16\CG^2\bE\|\omega\|^2\sum_{t=k}^{T-1}\bt^2 e^{2\lambda\bt}\Big(e^{-2\lambda\sum_{s=k}^{T-1}\beta_s} \Big)
        \nonumber\\
        & \overset{(c)}{\leq} 16e^{2\lambda}\CG^2\Big(e^{-2\lambda\sum_{s=k}^{T-1}\beta_s} \Big)\bE\|\omega\|^2\sum_{t=k}^{T-1}\bt^2
        \nonumber\\
        & \overset{(d)}{\leq} 16e^{2\lambda}\CG^2\Big(e^{-2\lambda\sum_{s=k}^{T-1}\beta_s} \Big)\bE\|\omega\|^2(T-k)\bk^2
    \end{align}
    where $(a)$ follows from statement 1 and the fact that $\|\Atil\|\leq \|\Aht\|+\|A\|\leq 4,$ $(b)$ follows by multiplying and dividing the summand by $e^{2\lambda\bt},$ $(c)$ and $(d)$ follow since $(\bt)$ is decreasing in $t$ with $\beta_0=1,$ and $(e)$ follows from the definition of $\CS$ in Table~\ref{tab: constants}.

    For the second term, note that
    \begin{align*}
        &\bE\Big\langle \Gdel_{s+1:T}\tilde{A}_s\Gdel_{k:s}\ \omega, \Gdel_{t+1:T}\Atil\Gdel_{k:t} \omega\Big\rangle
        \\
        & \overset{(a)}{=} \bE\Big\langle \Gdel_{s+1:T}\tilde{A}_s\Gdel_{k:s} \omega, \Gdel_{t+1:T}\bE_s[\Atil]\Gdel_{k:t}\ \omega\Big\rangle
        \\
        & \overset{(b)}{\leq} \Big(\|\Gdel_{k:s}\|\|\Gdel_{s+1:T}\|\|\Gdel_{k:t}\|\|\Gdel_{t+1:T}\|\Big)(16\Cerg\alpha^{(t-s)}) \bE\|\omega\|^2
        \\
        & \overset{(c)}{\leq} 16\Cerg\CG^4\Big(e^{-\lambda\sum_{i=k}^{s-1}\beta_i}\Big)\Big(e^{-\lambda\sum_{i=s+1}^{T-1}\beta_i} \Big)
        \\
        & \quad\qquad \times \Big(e^{-\lambda\sum_{j=k}^{t-1}\beta_j}\Big)\Big(e^{-\lambda\sum_{j=t+1}^{T-1}\beta_i} \Big)\alpha^{(t-s)}\bE\|\omega\|^2
        \\
        & \overset{(d)}{\leq} 16e^{2\lambda}\Cerg\CG^4 \Big( e^{-2\lambda\sum_{i=k}^{T-1}\beta_i} \Big)\alpha^{(t-s)}\bE\|\omega\|^2,
    \end{align*}
    where $(a)$ follows since $\omega$ is $\cF_k$-measurable and $\cF_k\subseteq \cF_s,$ for $s\geq k,$ $(b)$ follows since $\|\tilde{A}_s\|\leq 4$ and $\|\bE_s[\Atil]\|\leq 4\Cerg\alpha^{t-s},$ $(c)$ follows from statement 1, and $(d)$ follows by multiplying and dividing by $e^{\lambda(\beta_s+\bt)}$ and using $\beta_s+\bt< 2.$ 
    Hence, we have 
    \begin{align*}
        &2\sum_{k\leq s<t}^{T-1}\beta_s\bt\bE\Big\langle \Gdel_{s+1:T}\tilde{A}_s\Gdel_{k:s}\ \omega, \Gdel_{t+1:T}\Atil\Gdel_{k:t}\ \omega\Big\rangle
        \\
        & \leq 32e^{2\lambda}\Cerg\CG^4 \Big( e^{-2\lambda\sum_{i=k}^{T-1}\beta_i} \Big)\bE\|\omega\|^2 \sum_{k\leq s<t}^{T-1}\beta_s\beta_t\alpha^{(t-s)}
        \\
        & \overset{(a)}{\leq} 32e^{2\lambda}\Cerg\CG^4 \Big( e^{-2\lambda\sum_{i=k}^{T-1}\beta_i} \Big)\bE\|\omega\|^2 \sum_{k\leq s<t}^{T-1}\beta^2_s\alpha^{(t-s)} 
        \\
        & \overset{(b)}{\leq} \Big(\frac{32e^{2\lambda}}{1-\alpha}\Big)\Cerg\CG^4 \Big( e^{-2\lambda\sum_{i=k}^{T-1}\beta_i} \Big)\bE\|\omega\|^2 \sum_{s=k}^{T-1}\beta^2_s
        \\
        & \overset{(c)}{\leq} \Big(\frac{32e^{2\lambda}}{1-\alpha}\Big)\Cerg\CG^4 \Big( e^{-2\lambda\sum_{i=k}^{T-1}\beta_i} \Big)\bE\|\omega\|^2(T-k)\bk^2
        \\
        & \overset{(e)}{=} (\CS^2 - 16e^{2\lambda}\CG^2)\Big( e^{-2\lambda\sum_{i=k}^{T-1}\beta_i} \Big)\bE\|\omega\|^2(T-k)\bk^2
    \end{align*}
    where $(a)$ since $\bt\leq \beta_s$ for $s\leq t,$ $(b)$ is obtained by summing $\alpha^{t-s}$ over $t,$ and $(c)$ follows since $(\bt)$ is decreasing in $t.$ This completes the proof.
\hfill \scalebox{0.8}{$\blacksquare$}

%%%%%%%%%%%%%%%%%%%%%%%%%%%%%%%%%%%%%%%%%%%%%%%%%%

\emph{Proof of \textbf{Lemma~\ref{thm: technical.bounds}}}
    \noindent (i). To prove the \textbf{bound in (\ref{e: sum.L1.0})}, note that repeatedly applying \eqref{e: L.update} gives $\forall T>0,$
    \[       
        L^{(1)}_T = \sum_{k=0}^{T-1}\bk\Gdel_{k+1:T}(-\rho_k\vhk + \zhk).
    \]
    Using $\lvert\rho_k\rvert\leq 2\Rm, \|\vhk\|\leq 1$ and $\|\zhk\|\leq 4(\Rm +\|\thS\|),$ and applying Lemma~\ref{thm: mat.prod.bound} 
    \begin{align}\label{e: L1.star.bound}
        \|L^{(1)}_{T}\| &\leq 6(\Rm+\|\thS\|)\sum_{k=0}^{T-1}\bk\|\Gdel_{k+1:T}\|
        \nonumber \\
        & \leq 6(\Rm+\|\thS\|)\CG\Clambda =: B_L.
    \end{align}
    The claim follows as 
    \begin{align*}
        \sum_{t=\tstr}^{T-1}\|\Gdel_{\tstr:t}\| 
        & \leq \CG\sum_{t=\tstr}^{T-1}e^{-\lambda\sum_{s=\tstr}^{t-1}\beta_s} 
        \\
        & \leq \CG e^{\frac{\lambda}{(1-\beta)}\tstr^{(1-\beta)}}\sum_{t=\tstr}^{T-1}e^{-\frac{\lambda}{(1-\beta)}t^{(1-\beta)}} 
        \leq \xi_G.
    \end{align*}

    \noindent (ii). To get the \textbf{bound in (\ref{e: sum.L1.1})}, we proceed as follows:
    \begin{align*}
        \bE^{1/2}\bigg\|& \sum_{k=\tstr}^{T-2} \rho_k M_{k: T-1}\vhk\bigg\|^2 
        \overset{(a)}{\leq} \sum_{k=\tstr}^{T-2}\bE^{1/2}\|\rho_k M_{k: T-1}\vhk\|^2
        \\
        & \overset{(b)}{\leq} \KG \sum_{k=\tstr}^{T-2}\bE^{1/2}\|\rho_k\vhk\|^2
        \overset{(c)}{\leq} \KG \sum_{k=\tstr}^{T-2}\bE^{1/2}\rho_k^2
        \\
        & \overset{(d)}{\leq} \frac{\KG}{\sqrt{2}}\sum_{k=\tstr}^{T-2}\left[ \frac{\tauk C^{1/2}_{r,\lin}}{\sqrt{N(k+1)}} + \frac{C^{1/2}_{r,\qd}}{(k+1)} \right]
        \\
        & \leq \frac{\KG}{\sqrt{2}}\left[ 2C^{1/2}_{r,\lin}\frac{\tauT\sqrt{T}}{\sqrt{N}} + C^{1/2}_{r,\qd}\ln{(T)} \right],
    \end{align*}
    where $(a)$ uses triangle inequality, $(b)$ follows from Lemma~\ref{thm: mat.prod.bound}, $(c)$ follows since $\|\vhk\|\leq 1,$ and  $(d)$ follows from Lemma~\ref{thm: avg.rew.converge}. 
    \\

    \noindent (iii). We now prove the \textbf{bound in (\ref{e: sum.L1.2})}. Note that
    \begin{multline}\label{e: M.z.decomp}
            \bE\bigg\| \sum_{k=\tstr}^{T-2}M_{k:T-1}\zhk\bigg\|^2 = \sum_{k=\tstr}^{T-2}\bE\|M_{k:T-1}\zhk\|^2
            \\
            + 2\sum_{t=\tstr+1}^{T-2}\sum_{s=\tstr}^{t-1}\bE\langle M_{s:T-1}\hat{z}_s, M_{t:T-1}\zht\rangle.
    \end{multline}
    To bound these two sums on the r.h.s, we first bound $\|\zht\|^2.$ Recall from that $\zht = \frac{1}{N}\sum_{i=1}^N\zht^i$ where, for each agent $i,$
    \begin{align*}
        \zht^i := \big[\cR_{i}(s_{t}^{i}, a_{t}^{i})\phi(s_{t}^{i}) - b_{i}\big] 
        - \rS\big[\hat{v}_{t}^{i} - v_{i} \big]
        +  \big[A_{i} - \hat{A}^i_t \big]\thS.
    \end{align*}
    Further note that for $0\leq \tau<t$ and for every agent $i,$
    \begin{equation}\label{e: noise.geometric.bound}
        \|\bE_{t-\tau} \zht^i\| \leq 4(\Rm+\|\thS\|)\Cerg \alpha^\tau = \xi_M\Cerg\alpha^\tau.
    \end{equation}
    Hence, for distinct agents $i$ and $j,$ we have
    \begin{multline}\label{e: noise.inner.prod}
        \bE\big\langle \zht^i, \zht^j \big\rangle \overset{(a)}{=} \bE\bE_0\big\langle \zht^i, \zht^j \big\rangle \overset{(b)}{=} \bE\big\langle \bE_0\zht^i, \bE_0\zht^j  \big\rangle 
        \\
        \overset{(c)}{\leq} \bE\|\bE_0\zht^i\| \|\bE_0\zht^j \| \overset{(d)}{\leq} 16(\Rm+\|\thS\|)^2\Cerg^2 \alpha^{2t},
    \end{multline}
    where $(a)$ uses the tower property of expectation, $(b)$ follows from the fact that each agent's local trajectory is independent of the others, $(c)$ uses the Cauchy-Schwarz inequality, and $(d)$ uses \eqref{e: noise.geometric.bound} with $\tau=0.$
    Therefore, combining \eqref{e: noise.inner.prod} and $\|\zht^i\|^2\leq 16(\Rm+\|\thS\|)^2,$ we have
    \begin{align}\label{e: noise.bound.square}
        \bE\|\zht\|^2 & = \frac{1}{N^2}\sum_{i=1}^N \bE\|\zht^i\|^2 + \frac{2}{N^2}\sum_{i<j} \bE\big\langle \zht^i, \zht^j\big\rangle
        \nonumber\\
        & \leq 16(\Rm+\|\thS\|)^2\bigg[ \frac{1}{N^2}\sum_{i=1}^N 1 + \frac{2}{N^2}\sum_{i<j} \Cerg^2\alpha^{2t} \bigg] 
        \nonumber\\
        & \leq 16(\Rm+\|\thS\|)^2\bigg( \frac{1}{N} + \Cerg^2\alpha^{2t} \bigg)
        \nonumber\\
        & = \xi^2_M\bigg( \frac{1}{N} + \Cerg^2\alpha^{2t} \bigg).
    \end{align}
    Now, we focus on each term on the r.h.s of \eqref{e: M.z.decomp}. The first term is bounded as follows:
    \begin{align*}
        \sum_{k=\tstr}^{T-2}\bE&\|M_{k:T-1}\zhk\|^2 
        \overset{(a)}{\leq} \KG^2 \sum_{k=\tstr}^{T-2} \bE\|\zhk\|^2
        \\
        & \overset{(b)}{\leq} \KG^2\xi_M^2\sum_{k=\tstr}^{T-2}\bigg( \frac{1}{N} + \Cerg^2\alpha^{2t} \bigg)
        \\
        & \leq \KG^2\xi_M^2\bigg(\frac{T}{N} + \frac{\Cerg^2}{(1-\alpha^2)} \bigg)
        \\
        & \overset{(c)}{\leq} \KG^2\xi_M^2\bigg(\frac{T \tauT }{N} + \frac{\Cerg^2}{(1-\alpha^2)} \bigg),
    \end{align*}
    where $(a)$ follows from Lemma~\ref{thm: mat.prod.bound}, $(b)$ uses \eqref{e: noise.bound.square}, and $(c)$ uses $1\leq \tauT.$
    
    For the second term, we split the double summation into two cases: one where for each $\tstr<t<T-1,$ we have $t-2\taut\leq s<t,$ and the other where $s< t-2\taut.$ This gives the following decompositions:
    \begin{align*}
        \sum_{\tstr\leq s<t}^{T-1} 2\bE& \Big\langle M_{s: T-1}\hat{z}_s, M_{t: T-1}\zht\Big\rangle
        \\
        & = \sum_{t=\tstr+1}^{T-2}\sum_{s=t-2\taut}^{t-1} 2\bE\Big\langle M_{s: T-1} \hat{z}_s, M_{t: T-1}\zht\Big\rangle 
        \\
        & + \sum_{t=\tstr+1}^{T-2}\sum_{s=\tstr}^{t-2\taut-1} 2\bE\Big\langle M_{s: T-1} \hat{z}_s, M_{t: T-1}\zht\Big\rangle.
    \end{align*}
    We bound each case one by one. For the former case, i.e.,  when $t-2\taut \leq s<t,$ note that
    \begin{align*}
        2\bE\Big\langle M_{s: T-1} &\hat{z}_s, M_{t: T-1}\zht\Big\rangle
        \\
        \overset{(a)}{\leq} & \Big(\bE\|M_{s: T-1}\hat{z}_s\|^2 +  \bE\|M_{t: T-1}\zht\|^2\Big)
        \\
        \overset{(b)}{\leq} & \KG^2\Big(\bE\|\hat{z}_s\|^2 +  \bE\|\zht\|^2\Big)
        \\
        \overset{(c)}{\leq} & \KG^2\xi_M^2\bigg( \frac{2}{N} + \Cerg^2\alpha^{2s} + \Cerg^2\alpha^{2t} \bigg)
        \\
        \overset{(d)}{\leq} & \KG^2\xi_M^2\bigg( \frac{2}{N} + \Cerg^2\alpha^s\alpha^{t-2\taut} + \Cerg^2\alpha^{2t} \bigg)
        \\
        \overset{(e)}{\leq} & \KG^2\xi_M^2\bigg( \frac{2}{N} + \Cerg^2\alpha^s\alpha^{t-2\taut} + \Cerg^2\alpha^s\alpha^t \bigg),
    \end{align*}
    where $(a)$ uses the Cauchy-Schwarz inequality, $(b)$ uses Lemma~\ref{thm: mat.prod.bound}, $(c)$ uses \eqref{e: noise.bound.square}. Lastly, $(d)$ and $(e)$ use the fact that for $t-2\taut\leq s< t,$ $\alpha^{2s} \leq \alpha^s\alpha^{t-2\taut},$ and $\alpha^{2t} \leq \alpha^s\alpha^t,$ respectively. Hence, the double sum in this case is bounded as follows:
    \begin{align*}
        \sum_{t=\tstr+1}^{T-2}& \sum_{s=t-2\taut}^{t-1} 2\bE\Big\langle M_{s: T-1} \hat{z}_s, M_{t: T-1}\zht\Big\rangle
        \\
        & \leq \KG^2\xi_M^2 \sum_{t=\tstr+1}^{T-2} \sum_{s=t-2\taut}^{t-1} \bigg[ \frac{2}{N} + \Cerg^2(\alpha^s\alpha^{t-2\taut} + \alpha^s\alpha^t) \bigg]
        \\
        & \overset{(a)}{\leq} 2\KG^2\xi_M^2  \sum_{t=\tstr+1}^{T-2}\bigg[ \frac{2\taut}{N} + \frac{\Cerg^2}{(1-\alpha)}(\alpha^{t-2\taut} + \alpha^t ) \bigg]
        \\
        & \overset{(b)}{\leq} 2\KG^2\xi_M^2\tauT\bigg[ \frac{2T}{N} + \frac{2\Cerg^2}{(1-\alpha)}\sum_{t=0}^{\infty}\alpha^t \bigg ]
        \\
        & \leq  4\KG^2\xi_M^2\bigg[\frac{T\tauT}{N} + \frac{\Cerg^2}{(1-\alpha)^2} \bigg],
    \end{align*}
    where $(a)$ uses $\sum_{s=t-2\taut}^{t-1}\alpha^s\leq \sum_{s=t-2\taut}^{t-1}1\leq 2\taut$ for the first term and the geometric series formula for the rest of the terms. Whereas $(b)$ uses the fact that $\sum_{t=\tstr+1}^{T-2}\alpha^{t-2\tstr} \leq \sum_{t=0}^{\infty}\alpha^t.$

    Now, for the latter case, i.e., when $s<t-2\taut,$ we proceed as follows:
    \begin{align*}
        2\bE\Big\langle& M_{s: T-1}\hat{z}_s, M_{t:T-1}\zht\Big\rangle 
        \\
        & \overset{(a)}{=} 2\bE\bE_{t-2\taut}\Big\langle M_{s: T-1}\hat{z}_s, M_{t:T-1}\zht\Big\rangle
        \\
        & \overset{(b)}{=} 2\bE\Big\langle M_{s: T-1}\hat{z}_s, M_{t:T-1}\bE_{t-2\taut}\zht\Big\rangle
        \\
        & \overset{(c)}{\leq} \bt^2\bE \|M_{s: T-1}\hat{z}_s\|^2 + \bt^{-2}\bE\|M_{t:T-1}\bE_{t-2\taut}\zht\|^2
        \\
        & \overset{(d)}{\leq} \KG^2\Big[\bt^2\bE \|\hat{z}_s\|^2 + \bt^{-2}\bE\|\bE_{t-2\taut}\zht\|^2\Big]
        \\
        & \overset{(e)}{\leq} \KG^2\xi_M^2\bigg[ \bt^2\Big(\frac{1}{N} + \Cerg^2\alpha^{2t} \Big) + \bt^{-2}\Cerg^2\alpha^{4\taut} \bigg]
        \\
        & \overset{(f)}{\leq} \KG^2\xi_M^2\bigg[ \bt^2\Big(\frac{1}{N} + \Cerg^2\alpha^{2t} \Big) + \frac{\bt^{-2}\Cerg^2}{(t+1)^8} \bigg]
        \\
        & \overset{(g)}{\leq} \KG^2\xi_M^2\bigg[ \bt^2\Big(\frac{1}{N} + \Cerg^2\alpha^{2t} \Big) + \frac{\Cerg^2}{(t+1)^6} \bigg]
        \\
        & \overset{(h)}{\leq} \KG^2\xi_M^2\bigg[ \frac{\beta_s^2}{N} + \Cerg^2\bt^2\alpha^{2s} + \frac{\Cerg^2}{(t+1)^3(s+1)^3} \bigg],
    \end{align*}
    where $(a)$ uses the tower property of expectation, $(b)$ uses the fact that $\hat{z}_s$ is $\cF_{t-2\taut}$-measurable for $s<t-2\taut,$ $(c)$ uses the fact that for any $\alpha>0,$ and vectors $x,y,$ we have $2|<x,y>|\leq (\alpha\|x\|^2 + \alpha^{-1}\|y\|^2),$ $(d)$ uses Lemma~\ref{thm: mat.prod.bound}, $(e)$ follows from \eqref{e: noise.geometric.bound} and \eqref{e: noise.bound.square}, $(f)$ follows from the definition of $\taut,$ $(g)$ follows since $1/(t+1)^2\leq \bt^2,$ and $(h)$ follows since $\alpha^t< \alpha^s,$ $1/(t+1)\leq 1/(s+1),$ and $\bt\leq \beta_s,$ for $t>t-2\taut>s.$
    This gives us
    \begin{align*}
        & \sum_{t=\tstr+1}^{T-2}\sum_{s=\tstr}^{t-2\taut-1}\bE\langle M_{s: T-1}\hat{z}_s, M_{t:T-1}\zht\rangle 
        \\
        & \leq \KG^2\xi_M^2 \sum_{t=\tstr+1}^{T-2}\sum_{s=\tstr}^{t-2\taut-1}\bigg[ \frac{\beta_s^2}{N} + \Cerg^2\bt^2\alpha^{2s}
        \\
        & \qquad \qquad \qquad \qquad + \frac{\Cerg^2}{(t+1)(s+1)^3} \bigg]
        \\
        & \overset{(a)}{\leq} \KG^2\xi_M^2 \sum_{t=\tstr+1}^{T-2} \bigg[ \frac{\Cbeta}{N} + \frac{\Cerg^2\bt^2}{(1-\alpha^2)} + \frac{\pi^2\Cerg^2}{6(t+1)^3} \bigg]
        \\
        & \overset{(b)}{\leq} \KG^2\xi_M^2\bigg[ \Cbeta\frac{T}{N} + \frac{\Cerg^2\Cbeta}{(1-\alpha^2)} + \frac{\pi^4\Cerg^2}{36}\bigg]
        \\
        & \overset{(c)}{\leq} \KG^2\xi_M^2\bigg[ \Cbeta\frac{T\tauT}{N} + \frac{\Cerg^2\Cbeta}{(1-\alpha^2)} + \frac{\pi^4\Cerg^2}{36}\bigg], 
    \end{align*}
    where both $(a)$ and $(b)$ follow in a straightforward manner using the definition of $\Cbeta$ from Table~\ref{tab: constants}, the geometric series formula, and the fact that $\sum_{t>1} 1/t^3 \leq \sum_{t>1} 1/t^2 < \pi^2/6.$ Lastly, $(d)$ uses $1\leq \tauT.$  The claim follows.
    \\

    \noindent (iv). The \textbf{bound in (\ref{e: sum.rho})} is obtained follows:
    \begin{align*}
        & \bE^{1/2}\bigg\| \sum_{k=\tstr}^{T-1}\bk\rho_k S_{k+1:T}\hat{v}_k \bigg\|^2 \overset{(a)}{\leq} \sum_{k=0}^{T-1}\bk\bE^{1/2}\|\rho_k S_{k+1:T}\vhk\|^2
        \\
        & \overset{(b)}{\leq} \CS\sum_{k=\tstr}^{T-1}\bk^2\Big(e^{-\lambda\sum_{s=k+1}^{T-1}\beta_s} \Big)\sqrt{T-k-1}\ \bE^{1/2}\|\rho_k\vhk\|^2
        \\
        & \overset{(c)}{\leq} \CS\sqrt{T}\sum_{k=\tstr}^{T-1}\bk^2\Big(e^{-\lambda\sum_{s=k+1}^{T-1}\beta_s} \Big) \bE^{1/2}\rho_k^2
        \\
        & \overset{(d)}{\leq} \CS\sqrt{T}\sum_{k=\tstr}^{T-1}\bk^2\Big(e^{-\lambda\sum_{s=k+1}^{T-1}\beta_s} \Big)
        \\
        & \qquad \qquad \qquad \qquad \times \frac{1}{\sqrt{2}}\left[ \frac{C^{1/2}_{r,\lin}\tauT}{\sqrt{N(k+1)}} + \frac{C^{1/2}_{r,\qd}}{(k+1)} \right]
        \\
        & \overset{(e)}{\leq} \big(\bT\sqrt{T}\big)\  \frac{\CS\Clambda}{\sqrt{2}}\left[ \frac{\tauT C^{1/2}_{r,\lin}}{\sqrt{N(T+1)}} + \frac{C^{1/2}_{r,\qd}}{T+1} \right]
        \\
        & \overset{(f)}{\leq} \frac{\CS\Clambda}{\sqrt{2}}\left[ \frac{\tauT C^{1/2}_{r,\lin}}{\sqrt{N(T+1)}} + \frac{C^{1/2}_{r,\qd}}{T+1} \right],
    \end{align*}
    where $(a)$ uses triangle inequality, $(b)$ follows from Lemma~\ref{thm: mat.prod.bound}, $(c)$ uses $\|\vhk\|\leq 1$ and $\sqrt{T-k-1} < \sqrt{T},$ $(d)$ follows from Lemma~\ref{thm: avg.rew.converge} and $(e)$ follows from arguments in the proof of Lemma~\ref{thm: mat.prod.bound}. Lastly, $(f)$ follows since $\beta>1/2$ implies $\bT\sqrt{T}<1.$
    \\

    \noindent (v). Finally, we prove the \textbf{bound in (\ref{e: sum.S.z})}. We begin by decomposing the sum into two cases as follows:
    \[
        \sum_{k=\tstr}^{T-2} \bk S_{k+1:T}\zhk = X_T + Y_T,
    \]
    where
    \[
        X_T := \sum_{k=\tstr}^{T-\tauT-1} \bk S_{k+1:T}\zhk,
        \quad Y_T := \sum_{k=T-\tauT}^{T-2} \bk S_{k+1:T}\zhk.
    \]
    We now handle these two cases separately. We begin with the case when $T-\tauT\leq k<T-1.$ In this case, we have
    \begin{align*}
        & \bE^{1/2}\big\|Y_T\big\|^2 \overset{(a)}{\leq} \sum_{k=T-\tauT}^{T-2} \bk \bE^{1/2}\|S_{k+1:T}\zhk\|^2
        \\
        & \overset{(b)}{\leq} \CS \sum_{k=T-\tauT}^{T-1}\bk^2\sqrt{T-k}\Big(e^{-\lambda\sum_{s=k+1}^{T-1}\beta_s} \Big)\bE\|\zhk\|^2
        \\
        & \overset{(c)}{\leq} \CS\sum_{k=T-\tauT}^{T-1}\bk^2\sqrt{\tauT}\Big(e^{-\lambda\sum_{s=k+1}^{T-1}\beta_s} \Big)\bE^{1/2}\|\zhk\|^2
        \\
        & \overset{(d)}{\leq} \CS\xi_M\sqrt{\tauT}\sum_{k=T-\tauT}^{T-1}\bk^2\Big(e^{-\lambda\sum_{s=k+1}^{T-1}\beta_s} \Big)\bigg[\frac{1}{\sqrt{N}} + \Cerg\alpha^k\bigg]
        \\
        & \overset{(e)}{\leq} \CS\xi_M\sqrt{\tauT}\sum_{k=T-\tauT}^{T-1}\bk^2\Big(e^{-\lambda\sum_{s=k+1}^{T-1}\beta_s} \Big)\bigg[\frac{1}{\sqrt{N}} + \frac{\Cerg}{(t+1)^4}\bigg]
        \\
        & \overset{(f)}{\leq} \CS\Clambda\xi_M\sqrt{\tauT}\bT\bigg[ \frac{1}{\sqrt{N}} + \frac{\Cerg}{(T+1)^4} \bigg] 
        \\
        & \overset{(g)}{\leq} \CS\Clambda\xi_M\bigg[ \frac{\sqrt{\tauT}\bT}{\sqrt{N}} + \frac{\Cerg}{(T+1)^4} \bigg]
        \\
        & \overset{(h)}{\leq} \CS\Clambda(1+\Cerg)\xi_M\  \tauT\bT, 
    \end{align*}
    where $(a)$ uses triangle inequality, $(b)$ uses Lemma~\ref{thm: mat.prod.bound}, $(c)$ follows since $k\geq T-\tauT,$ $(d)$ follows from \eqref{e: noise.bound.square}, $(e)$ follows since for $k\geq \tstr,$ we have $k\geq 2\tauk$ and hence $\alpha^k\leq \alpha^{2\tauk}\leq 1/(k+1)^4,$ $(e)$ follows form arguments in the proof of Lemma~\ref{thm: mat.prod.bound} and $(g)$ uses $\sqrt{\tauT}\bT\leq 1$ for the second term. Finally, $(h)$ uses $1\leq \sqrt{\tauT}\leq \tauT$ and $1/(T+1)^4\leq \bT.$

    Now, we handle the second case, i.e., when $k+\tauT<T.$ For this case, note that 
    \[
        S_{k+1:T} = \Gdel_{k+\tauT:T}S_{k+1:k+\tauT} 
        + S_{k+\tauT:T}\Gdel_{k+1:k+\tauT},
    \]
    and therefore we have 
    \begin{equation}\label{e: S.z.decomp}
        X_T 
        = X^{(1)}_T + X^{(2)}_T.
    \end{equation}
    where 
    \[
        X^{(1)}_T := \sum_{k=\tstr}^{T-\tauT-1}\bk\Gdel_{k+\tauT:T}S_{k+1:k+\tauT}\zhk 
    \]
    and 
    \[
        X^{(2)}_T := \sum_{k=\tstr}^{T-\tauT-1} \bk S_{k+\tauT:T}\Gdel_{k+1:k+\tauT}\zhk.
    \]
    To finish the proof, we bound these two sums $X^{(1)}_T$ and $X^{(2)}_T$. Note that $X^{(1)}_T$ can be handled in a straightforward manner as follows:
    \begin{align*}
        & \bE^{1/2}\Big\|X^{(1)}_T\Big\|^2 \overset{(a)}{\leq} \sum_{k=\tstr}^{T-\tauT-1} \bk \bE^{1/2}\|\Gdel_{k+\tauT:T}S_{k+1:k+\tauT}\zhk\|^2
        \\
        & \leq \sum_{k=\tstr}^{T-\tauT-1} \bk \|\Gdel_{k+\tauT:T}\|\ \bE^{1/2}\|S_{k+1:k+\tauT}\zhk\|^2
        \\
        & \overset{(b)}{\leq} \CG\CS\sum_{k=\tstr}^{T-\tauT-1}\bk\sqrt{\tauT}\Big( e^{-\lambda\sum_{s=k+\tauT}^{T-1}\beta_s} \Big)
        \\
        & \qquad \qquad \qquad \times \beta_{k+1}\Big( e^{-\lambda\sum_{s=k+1}^{k+\tauT-1}\beta_s} \Big)\bE^{1/2}\|\zhk\|^2
        \\
        & \overset{(c)}{\leq} \CG\CS\sqrt{\tauT}\sum_{k=\tstr}^{T-\tauT-1}\bk^2\Big( e^{-\lambda\sum_{s=k+1}^{T-1}\beta_s} \Big)\bE^{1/2} \|\zhk\|^2
        \\
        & \overset{(d)}{\leq} \CG\CS\xi_M\sqrt{\tauT}\sum_{k=\tstr}^{T-\tauT-1}\bk^2\Big( e^{-\lambda\sum_{s=k+1}^{T-1}\beta_s} \Big)\bigg[ \frac{1}{\sqrt{N}} + \Cerg\alpha^k \bigg]
        \\
        & \overset{(e)}{\leq} \CG\CS\xi_M\sqrt{\tauT}\sum_{k=\tstr}^{T-1}\bk^2\Big( e^{-\lambda\sum_{s=k+1}^{T-1}\beta_s} \Big)\bigg[ \frac{1}{\sqrt{N}} + \Cerg\alpha^k \bigg]
        \\
        & \overset{(f)}{\leq} \CG\CS\xi_M\sqrt{\tauT}\sum_{k=\tstr}^{T-1}\bk^2\Big( e^{-\lambda\sum_{s=k+1}^{T-1}\beta_s} \Big)\bigg[ \frac{1}{\sqrt{N}} + \frac{\Cerg}{(k+1)^4} \bigg]
        \\
        & \overset{(g)}{\leq} \CG\CS\Clambda\xi_M\sqrt{\tauT}\bT\bigg[ \frac{1}{\sqrt{N}} + \frac{\Cerg}{(T+1)^4} \bigg]
        \\
        & \overset{(h)}{\leq} \CG\CS\Clambda\xi_M\bigg[ \frac{\sqrt{\tauT}\bT}{\sqrt{N}} + \frac{\Cerg}{(T+1)^4} \bigg]
        \\
        & \leq \CG\CS\Clambda(1+\Cerg)\xi_M\ \tauT\bT,
    \end{align*}
    where $(a)$ uses triangle inequality, $(b)$ uses Lemma~\ref{thm: mat.prod.bound}, $(c)$ follows from \eqref{e: noise.bound.square}, $(d)$ uses $\beta_{k+1}<\bk,$ $(e)$ follows since the summands are non-negative, $(f)$ follows since $k>2\tauk$ for $k\geq \tstr$ and hence $\alpha^k \leq \alpha^{2\tauk}< 1/(k+1)^4,$ $(g)$ follows from the arguments in Lemma~\ref{thm: mat.prod.bound}, and $(h)$ is obtained by using $\sqrt{\tauT}\bT<1$ on the second term.

    Finally, we conclude by bounding $X^{(2)}_T$. To do so, we need to decompose $X^{(2)}_T$ suitably. To do so, we introduce the following notation.
    
    Let $(s_k, a_k)$ denote $((s_k^i,a^i_k): i\in [N])\in (\cS\times \cA)^N$ and refer to the following lemma:
    
    \begin{lemma}[\textup{\cite[Lemma 3]{durmus2024finite}}]
         Given a fixed $\tau>0,$ there exists a random process $(\ts_k, \ta_k)$ such that the following holds for every $k\geq 0:$
        \begin{enumerate}
            \item $(\ts_k,\ta_k)$ is independent of $(s_\ell, a_\ell),$ for every $\ell\geq k+\tau;$
            \item $\bP((\ts_k, \ta_k) \neq (s_k, a_k))\leq \Cerg\alpha^\tau;$
            \item $(\ts_k, \ta_k)$ has the same distribution as $(s_k,a_k).$
        \end{enumerate}
    \end{lemma}

    To exploit the above lemma, we choose $\tau = \tauT$ and define for every $k\geq 0,$
    \begin{multline*}
        \tz_k := \frac{1}{N}\sum_{i=1}^{N}\bigg(\big[\cR_{i}(\ts_k^{i}, \ta_k^{i})\phi(\ts_k^{i}) - b_i \big] - r^*\big[\phi(\ts_k^{i}) - v_i \big]
        \\
        - \Big[\phi(\ts_k^{i})(\phi(\ts_k^{i}) - \phi(\ts_{k+1}^i))^{\top} - A_i \Big]\theta^* \bigg).
    \end{multline*}
    Now we decompose $X^{(1)}_T$ as follows:
    \[
        X^{(2)}_T = X^{(21)}_T + X^{(22)}_T + X^{(23)}_T,
    \]
    where
    \begin{align*}
        X^{(21)}_T & := \sum_{k=\tstr}^{T-\tauT-1}\bk S_{k+\tauT:T}\Gdel_{k+1:k+\tauT}\bE\tz_k,
        \\
        X^{(22)}_T & := \sum_{k=\tstr}^{T-\tauT-1}\bk S_{k+\tauT:T}\Gdel_{k+1:k+\tauT}(\tz_k - \bE\tz_k),
        \\
        X^{(23)}_T & := \sum_{k=\tstr}^{T-\tauT-1}\bk S_{k+\tauT:T}\Gdel_{k+1:k+\tauT}(z_k - \tz_k).
    \end{align*}
    Further, define for $k\geq 0,$
    \[
        \cF^{k} := \sigma\Big( \big\{ s_\ell,a_\ell : \ell \geq k+\tauT  \big\}\Big).
    \]
    With this we bound $X^{(21)}_T$ as follows:
    \begin{align*}
        & \bE^{1/2}\Big\| X^{(21)}_T\Big\|^2 \overset{(a)}{\leq} \sum_{k=\tstr}^{T-\tauT-1}\bk \bE^{1/2}\|S_{k+\tauT:T}\Gdel_{k+1:k+\tauT}\bE\tz_k\|^2
        \\
        & \leq \sum_{k=\tstr}^{T-\tauT-1}\bk \|\Gdel_{k+1:k+\tauT}\|\bE^{1/2}\|S_{k+\tauT:T}\bE\tz_k\|^2
        \\
        & \overset{(b)}{\leq} \CG\CS\sum_{k=\tstr}^{T-\tauT-1}\bk\Big(e^{-\lambda\sum_{s=k+1}^{k+\tauT-1}\beta_s} \Big)
        \\
        & \qquad \qquad\times \beta_{k+\tauT}\sqrt{T-k-\tauT+1}\Big(e^{-\lambda\sum_{s=k+\tauT}^{T-1}} \Big)\|\bE\tz_k\|
        \\
        & \overset{(c)}{\leq} \CG\CS\sqrt{T}\sum_{k=\tstr}^{T-\tauT-1}\bk^2\Big(e^{-\lambda\sum_{s=k+1}^{T-1}\beta_s} \Big)\|\bE\tz_k\|
        \\
        & \overset{(d)}{\leq} \CG\CS\sqrt{T}\sum_{k=\tstr}^{T-\tauT-1}\bk^2\Big(e^{-\lambda\sum_{s=k+1}^{T-1}\beta_s} \Big)\|\bE z_k\|
        \\
        & \overset{(e)}{\leq} \CG\CS\Cerg\xi_M\sqrt{T}\sum_{k=\tstr}^{T-\tauT-1}\bk^2\Big(e^{-\lambda\sum_{s=k+1}^{T-1}\beta_s} \Big)\alpha^k
        \\
        & \overset{(f)}{\leq} \CG\CS\Cerg\xi_M\sqrt{T}\sum_{k=\tstr}^{T-\tauT-1}\bk\Big(e^{-\lambda\sum_{s=k+1}^{T-1}\beta_s} \Big)\frac{\bk}{(k+1)^4}
        \\
        & \overset{(g)}{\leq} \CG\CS\Cerg\xi_M\sqrt{T}\sum_{k=\tstr}^{T-1}\bk\Big(e^{-\lambda\sum_{s=k+1}^{T-1}\beta_s} \Big)\frac{\bk}{(k+1)^4}
        \\
        & \overset{(h)}{\leq} \CG\CS\Cerg\Clambda\xi_M\bigg[ \frac{\sqrt{T}\bT}{(T+1)^4}\bigg]
        \\
        & \overset{(i)}{\leq} \leq \CG\CS\Cerg\Clambda\xi_M\ \tauT\bT,
    \end{align*}
    where $(a)$ uses triangle inequality, $(b)$ Lemma~\ref{thm: mat.prod.bound}, $(c)$ follows since $\beta_{k+\tauT}< \bk$ and $T-(k+\tauT-1)\leq T,$ $(d)$ follows since $(\ts^i_k,\ta^i_k)$ and $(s^i_k,a^i_k)$ have the same distribution for each $i,$ $(e)$ follows since $\|\bE z_k\|= \|\bE\bE_0 z_k\|\leq \bE\|\bE_0 z_k\|$ and \eqref{e: noise.geometric.bound} implies that $\|\bE_0 z_k\|\leq \xi_M\Cerg\alpha^k,$ $(f)$ follows since for $k>\tstr,$ we have $k>2\tauk$ and $\alpha_k< \alpha^{2\tauk}<1/(k+1)^4,$ $(g)$ follows since the summands are non-negative, and $(h)$ follows from arguments in Lemma~\ref{thm: mat.prod.bound}. Finally, $(h)$ uses $\sqrt{T}\bT<1,$ and $(i)$ uses $\sqrt{T}/ (T+1)^4\leq 1\leq \tauT.$ 

    Next, we bound $X^{(23)}_T$ as follows: From the definitions of $\zhk$ and $\tz_k,$ we know $\|\zhk - \tz_k\| \leq 2\xi_M\ \ones_{\{ \zhk \neq \tz_k \}}.$ Consequently,
    \begin{equation}\label{e: noise.diff.exp}
        \bE^{1/2}\|\zhk - \tz_k\|^2 \leq 2\xi_M\ \bP(\zhk \neq \tz_k) \leq 2\xi_M\Cerg\alpha^{\tauT},
    \end{equation}
    where the second inequality follows from Lemma. Along with this, we need the following crude bound:
    \begin{align}\label{e: S.crude.bound}
        & \|S_{k+\tauT:T}\| \leq \sum_{s=k+\tauT}^{T-1}\beta_s \|\Gdel_{s+1:T}\|\|\tilde{A}_s\|\|\Gdel_{k+\tauT:s}\|
        \nonumber\\
        & \overset{(a)}{\leq} 2\CG^2\sum_{s=k+\tauT}^{T-1}\beta_s \Big( e^{-\lambda\sum_{r=s+1}^{T-1}\beta_r} \Big)\Big( e^{-\lambda\sum_{r=k+\tauT}^{s-1}\beta_r} \Big)
        \nonumber\\
        & \overset{(b)}{\leq}  2\CG^2\sum_{s=k+\tauT}^{T-1}\beta_s e^{\lambda\beta_s}\Big( e^{-\lambda\sum_{r=k+\tauT}^{T-1}\beta_r} \Big)
        \nonumber\\
        & \overset{(c)}{\leq}  2e^\lambda\CG^2\  \Big( e^{-\lambda\sum_{r=k+\tauT}^{T-1}\beta_r} \Big)\sum_{s=k+\tauT}^{T-1}\beta_s
        \nonumber\\
        & \overset{(d)}{\leq}  \frac{2e^\lambda\CG^2}{(1-\beta)}  \Big( e^{-\lambda\sum_{r=k+\tauT}^{T-1}\beta_r} \Big)T^{(1-\beta)},
    \end{align}
    where $(a)$ follows from Lemma~\ref{thm: mat.prod.bound}, $(b)$ is obtained by multiplying and dividing by $e^{\lambda\beta_s},$ $(c)$ follows as $e^{\lambda\beta_s}\leq e^{\lambda},$ and $(d)$ follows as $\sum_{s=k+\tauT}^{T-1}1/(s+1)^\beta\leq \int_0^{T-1}dx/(x+1)^\beta< T^{(1-\beta)}/(1-\beta).$
    With this, we bound $X^{(23)}_T$ as follows:
    \begin{align*}
        & \bE^{1/2}\Big\|X^{(23)}_T\Big\|^2 
        \\
        & \leq \sum_{k=\tstr}^{T-\tauT-1} \bk\|\Gdel_{k+1:k+\tauT}\|\bE^{1/2}\|S_{k+\tauT:T}(\zhk - \tz_k)\|^2
        \\
        & \overset{(a)}{\leq} \frac{\CG\CS}{(1-\beta)}\sum_{k=\tstr}^{T-\tauT-1} \bk\Big(e^{-\lambda\sum_{s=k+1}^{k+\tauT-1} \beta_s } \Big)
        \\
        & \qquad \qquad \qquad\times \Big(e^{-\lambda\sum_{s=k+\tauT}^{T-1}\beta_s} \Big) T^{(1-\beta)}\  \bE^{1/2}\|\zhk - \tz_k\|^2
        \\
        & \overset{(b)}{\leq} \frac{2\CG\CS\Cerg\xi_M}{(1-\beta)} \alpha^{\tauT} T^{(1-\beta )}\sum_{k=\tstr}^{T-1} \bk\Big(e^{-\lambda\sum_{s=k+1}^{T-1} \beta_s } \Big)
        \\
        & \overset{(c)}{\leq} \frac{2\CG\CS\Cerg\Clambda\xi_M}{(1-\beta)} \alpha^{\tauT} T^{(1-\beta )}
        \\
        & \overset{(d)}{\leq} \frac{2\CG\CS\Cerg\Clambda\xi_M}{(1-\beta)}\frac{\bT}{(T+1)}
        \\
        & \overset{(e)}{\leq} \frac{2\CG\CS\Cerg\Clambda\xi_M}{(1-\beta)}\ \tauT\bT,
    \end{align*}
    where $(a)$ is obtained by combining Lemma~\ref{thm: mat.prod.bound} and \eqref{e: S.crude.bound}, $(b)$ uses \eqref{e: noise.diff.exp}, and $(c)$ uses Lemma~\ref{thm: mat.prod.bound}. Lastly, $(d)$ follows since $\alpha^{\tauT}<1/(T+1)^2$ and $T^{(1-\beta)}< (T+1)\bT,$ and $(i)$ uses $1/(T+1)< 1\leq \tauT.$

    At last, we are ready to bound $X^{(22)}_T$ and finish the proof of Lemma~\ref{thm: technical.bounds}. 
    \[
        \Big\|X^{(22)}_T\Big\|^2 = \sum_{k=\tstr}^{T-\tauT-1}\|\Omega_k\|^2 + 2\sum_{\tstr\leq  s<t}^{T-\tauT-1}\big\langle \Omega_s, \Omega_t\big\rangle,
    \]
    where $\Omega_k := \bk S_{k+\tauT:T}\Gdel_{k+1:k+\tauT}(\tz_k - \bE\tz_k).$ Further, note that
    \begin{multline*}
        \bE\big\langle \Omega_s, \Omega_t\big\rangle 
        = \bE\bE\Big[\big\langle \Omega_s, \Omega_t\big\rangle\big| \cF^s\Big]
        \\
        \overset{(a)}{=} \bE\Big\langle \beta_s S_{s+\tauT:T}\Gdel_{s+1:s+\tauT}\bE[\tz_s - \bE\tz_s], \Omega_t \Big\rangle =0,
    \end{multline*}
    where $(a)$ follows since $S_{s+\tauT:T}$ and $\Omega_t$ are $\cF^s$-measurable, whereas $(\tz_s-\bE\tz_s)$ is independent of $\cF^s.$
    Thus, we have
    \begin{align}\label{e: bound.Xtt}
        & \bE\Big\|X^{(22)}_T\Big\|^2 \leq  \sum_{k=\tstr}^{T-\tauT-1}\bE\|\Omega_k\|^2
        \nonumber\\
        & \leq  \sum_{k=\tstr}^{T-\tauT-1}\bk^2\|\Gdel_{k+1:k+\tauT}\|^2\bE\|S_{k+\tauT:T}(\tz_k - \bE\tz_k)\|^2
        \nonumber\\
        & \overset{(a)}{\leq} 4\xi_M^2\sum_{k=\tstr}^{T-\tauT-1}\bk^2\|\Gdel_{k+1:k+\tauT}\|^2\bE^{1/2}\|S_{k+\tauT:T}\|^2
        \nonumber\\
        & \overset{(b)}{\leq} 4\CG^2\CS^2\xi^2_M\sum_{k=\tstr}^{T-\tauT-1}\bk^2\Big( e^{-2\lambda\sum_{s=k+1}^{k+\tauT-1}\beta_s} \Big)
        \nonumber\\
        & \qquad \qquad \qquad \times \bk^2(T-k-\tauT)\Big( e^{-2\lambda\sum_{s=k+\tauT}^{T-1}\beta_s} \Big)
        \nonumber\\
        & \leq 4\CG^2\CS^2\xi^2_M \sum_{k=\tstr}^{T-\tauT-1}\bk^4(T-\tauT-k)\Big( e^{-2\lambda\sum_{s=k+1}^{T-1}\beta_s} \Big)
        \nonumber\\
        & \overset{(c)}{\leq} 4\CG^2\CS^2\xi^2_M \sum_{k=\tstr}^{T-1}\bk^4(T-k)\Big( e^{-2\lambda\sum_{s=k+1}^{T-1}\beta_s} \Big)
        \nonumber\\
        & \overset{(d)}{\leq} 4\CG^2\CS^2\xi^2_M \sum_{m=1}^{T-\tstr}\beta_{T-m}^4 m\Big( e^{-2\lambda\sum_{\ell=1}^{m-1}\beta_{T-\ell}} \Big)
        \nonumber\\
        & \leq 4\CG^2\CS^2\xi^2_M\Big( \Omone_T + \Omtwo_T \Big),
    \end{align}
    where 
    \begin{align*}
        &\Omone_T := \sum_{m=1}^{\floor{T/2}}\beta_{T-m}^4 m\Big( e^{-2\lambda\sum_{\ell=1}^{m-1}\beta_{T-\ell}} \Big)
        \\
        &\Omtwo_T := \sum_{m=\floor{T/2}+1}^{T-\tstr}\beta_{T-m}^4 m\Big( e^{-2\lambda\sum_{\ell=1}^{m-1}\beta_{T-\ell}} \Big).
    \end{align*}
    We claim that $\Omone_T, \Omtwo_T = O(\bT^2).$ To bound $\Omone_T,$ note that
    \begin{align}\label{e: bound.Omegao}
        \Omone_T & \overset{(a)}{\leq} 256\bT^4 \sum_{m=1}^{\floor{T/2}}m\Big( e^{-2\lambda\sum_{\ell=1}^{m-1}\beta_{T-\ell}} \Big)
        \nonumber\\
        & \overset{(b)}{\leq} 256\bT^4 \sum_{m=1}^{\floor{T/2}}m\Big( e^{-2\lambda(m-1)\bT} \Big)
        \nonumber\\
        & \leq 256\bT^4 \sum_{m=1}^{\infty }m\Big( e^{-2\lambda(m-1)\bT} \Big)
        \nonumber\\
        & \overset{(c)}{\leq} \frac{256\bT^4 }{(1- e^{-2\lambda\bT} )^2} 
        \nonumber\\
        & \overset{(d)}{\leq} \bigg(\frac{256}{\lambda^2}\bigg)\frac{\bT^4}{\bT^2} \overset{(e)}{\leq} \bigg(\frac{256}{\lambda^2}\bigg)\tauT^2\bT^2,
    \end{align}
    where $(a)$ follows since for $m<T/2,$ we have $T-m>T/2$ and $\beta_{T-m}< \beta_{T/2} = 2^\beta/T^\beta<4\bT,$ $(b)$ uses $\sum_{\ell=1}^{\floor{T/2}}\beta_{T-\ell}\geq (m-1)\beta_{T-m+1} > (m-1)\bT,$ and $(c)$ uses the fact that $\sum_{m=1}^{\infty}m\ r^{m-1} = 1/(1-r)^2,$ for $r:= \exp{(-2\lambda\bT)}<1.$ Lastly, $(d)$ uses the inequality $(1-e^{-y})> y/2,$ with $y:= 2\lambda\bT,$ and $(e)$ uses $1\leq \tauT^2.$

    Likewise, $\Omtwo_T$ is bounded as follows:
    \begin{align}\label{e: bound.Omegat}
        \Omtwo_T & \overset{(a)}{\leq} \sum_{m=\floor{T/2}+1}^{T-\tstr}m \Big( e^{-2\lambda\sum_{\ell=1}^{m-1}\beta_{T-\ell}} \Big)
        \nonumber\\
        & \overset{(b)}{\leq} \sum_{m=\floor{T/2}+1}^{T-\tstr}m e^{-\frac{6\lambda}{10(1-\beta)}T^{(1-\beta)}}
        \nonumber\\
        & \overset{(c)}{\leq} T^2\ e^{-\frac{6\lambda}{10(1-\beta)}T^{(1-\beta)}} \overset{(d)}{\leq} \xi^2_\Omega\ \tauT^2\bT^2,
    \end{align}
    where $(a)$ uses $\beta_{T-m}\leq1,$ $(b)$ follows since $\sum_{\ell=1}^{m-1}\beta_{T-\ell}\geq \big[T^{(1-\beta)} - (T/2)^{(1-\beta)} \big]/(1-\beta) > 3T^{(1-\beta)}/10(1-\beta),$ $(c)$ follows since $\sum_{m= \floor{T/2}+1}^{T-\tstr} m \leq \sum_{m = \floor{T/2}}^{T}m \leq T^2/2 + T,$ and for $T>1,$ $T\leq T^2/2.$ Finally, $(d)$ uses the definition of $\xi_\Omega$ from Table~\ref{tab: constants}. Combining \eqref{e: bound.Xtt}, \eqref{e: bound.Omegao}, and \eqref{e: bound.Omegat} gives
    \begin{align*}
        & \bE^{1/2}\Big\|X^{(22)}_T\Big\|^2 \leq 2\CG\CS\xi_M\Big( \Omone_T + \Omtwo_T \Big)^{1/2}
        \\
        & \leq 2\CG\CS\xi_M\bigg[ \frac{16}{\lambda} + \xi_\Omega \bigg]\tauT\bT.
    \end{align*}
    This completes the proof of \eqref{e: sum.S.z}) and Lemma~\ref{thm: technical.bounds}.
\hfill \scalebox{0.8}{$\blacksquare$}
%
% \[
%     \frac{1}{T}\sum_{t=\tstr}^{T-1}t^{-\beta} \leq \frac{T^{-\beta}}{(1-\beta)} 
% \]
% %
% and
% %
% \[
%     \sum_{t=\tstr}^{T-1} e^{-\frac{\lambda T^{(1-\beta)}}{(1-\beta)}} \leq \left(\frac{2}{e\lambda}\right)^{\frac{2}{1-\beta}}\sum_{t=\tstr}^{T-1}\frac{1}{t^2} \leq e^{\frac{\pi^2}{6}}\left(\frac{2}{e\lambda}\right)^{\frac{2}{1-\beta}} .
% \]
% %

%
%

%%%%%%%%%%%%%%%%%%%%%%%%%%%%%%%%%%%%%%%%%%%%%%%%%%%%%%%%%%%%
\section{Experiments}
\label{sec:experiments}

This section discusses the performance of AvgFedTD(0) and ExpFedTD(0) under different parameter choices. We wrote the code for both AvgFedTD(0) and ExpFedTD(0) in Julia and Python, using Visual Studio Code Editor. 

% The code is divided into four parts: Imports and Constants, Value Generating Functions, Algorithm Execution, and Plot Generation. In the first two of these parts, additional packages are imported, including Random, LinearAlgebra, PyCall, StatsBase, ProgressMeter, Plots, Plots.PlotMeasures, and LaTeXStrings. 

% Also, the following hyperparameters are initialized in these sections: number of runs, states, actions, agents, and iterations between server and agents, $d, \er, \ep,\beta,\Rm$, and sampling type, i.e., ``Markov'' and ``IID''. In the third part, Python functions use the NumPy library to create policy, feature vectors, step sizes for each iteration, rewards, transition probabilities, policy-induced rewards and transition probabilities, and the corresponding stationary distributions for each agent. Using these, $\theta_1^*$ is obtained, which is used as a benchmark for the performance of the algorithms. The seed for the NumPy random value generator is set to $5,$ and the Python code is integrated into Julia using the PyCall library. In the fourth section, the values from the previous section are used to simulate the algorithm. In each iteration, the square difference between $\bar{\theta}$ and $\theta_1^*$ is calculated, stored, and then later averaged across all the runs. Lastly, plots are generated using the averaged mean square difference between $\bar{\theta}$ and $\theta_1^*$ in the last part. The $X$-axis represents the number of iterations, while the $Y$-axis is the mean square difference calculated during each iteration averaged across multiple runs.

In all of our experiments, we consider  $|\cS|=|\cA|=100,$ $d=21,$ and $N \in \{2, 5, 10, 20\}$. Each experiment consists of $300$ runs and each run has $10000$ iterations. We conduct our experiments on single process of Intel $i7-11800H$ and the running time was around $10$ minutes, on average.

% For a simulation running on a while setting the number of runs to $300$, the number of states and number of actions to $100$, $d$ to $21$, the number of observable agents to $2, 5, 10,$ and $20,$ and the number of iterations to $10000$, the running time of implementing Algorithm AvgFedTD(0) is around $10$ minutes and Algorithm ExpFedTD(0) is around $11$ minutes.

In the first experiment, we set $\varepsilon_r$ = $\varepsilon_p=0.5,$ $\beta=0.6,$ and $\Rm=1.$  For each algorithm, i.e., AvgFedTD(0) and ExpFedTD(0), we then randomly generated a policy $\mu$ and a feature matrix $\Phi$ while ensuring Assumption~$\cA_4.$ Next, we randomly generated the transition probability matrices and the reward functions for the $N$ agents. We keep all the above quantities fixed across all our runs. We use the MDP of the first agent to determine $\theta_{1}^{*}$, which we use as a reference to calculate the error $\|\bar{\theta}_{t} - \theta_{1}^{*}\|^2_{2}$ at each iteration $t$. Figure~\ref{fig:MSE.compare} plots this error, averaged over $300$ runs, against $t$ for different $N.$ Figure~\ref{fig:MSE.compare}(a) provides the result for AvgFedTD(0), Figure~\ref{fig:MSE.compare}(b) for the average-reward federated variant of the algorithm proposed in \cite{zhang2021finite},  Figure~\ref{fig:MSE.compare}(c) for ExpFedTD(0), and  Figure~\ref{fig:MSE.compare}(d) for the algorithm proposed in~\cite{wang2024federated}, under the same problem setting. For Figure~\ref{fig:MSE.compare}(c) and Figure~\ref{fig:MSE.compare}(d), the discount factor $\gamma$ is set to $0.3.$  We observe that  our proposed algorithms show the desired convergence rate of $O(\frac{1}{NT})$. This rate is the same as in~\cite{wang2024federated} even though our algorithms are parameter-free, while the other algorithms depend on unknown problem parameters. 

\begin{figure}
    \includegraphics[width=\columnwidth,keepaspectratio]{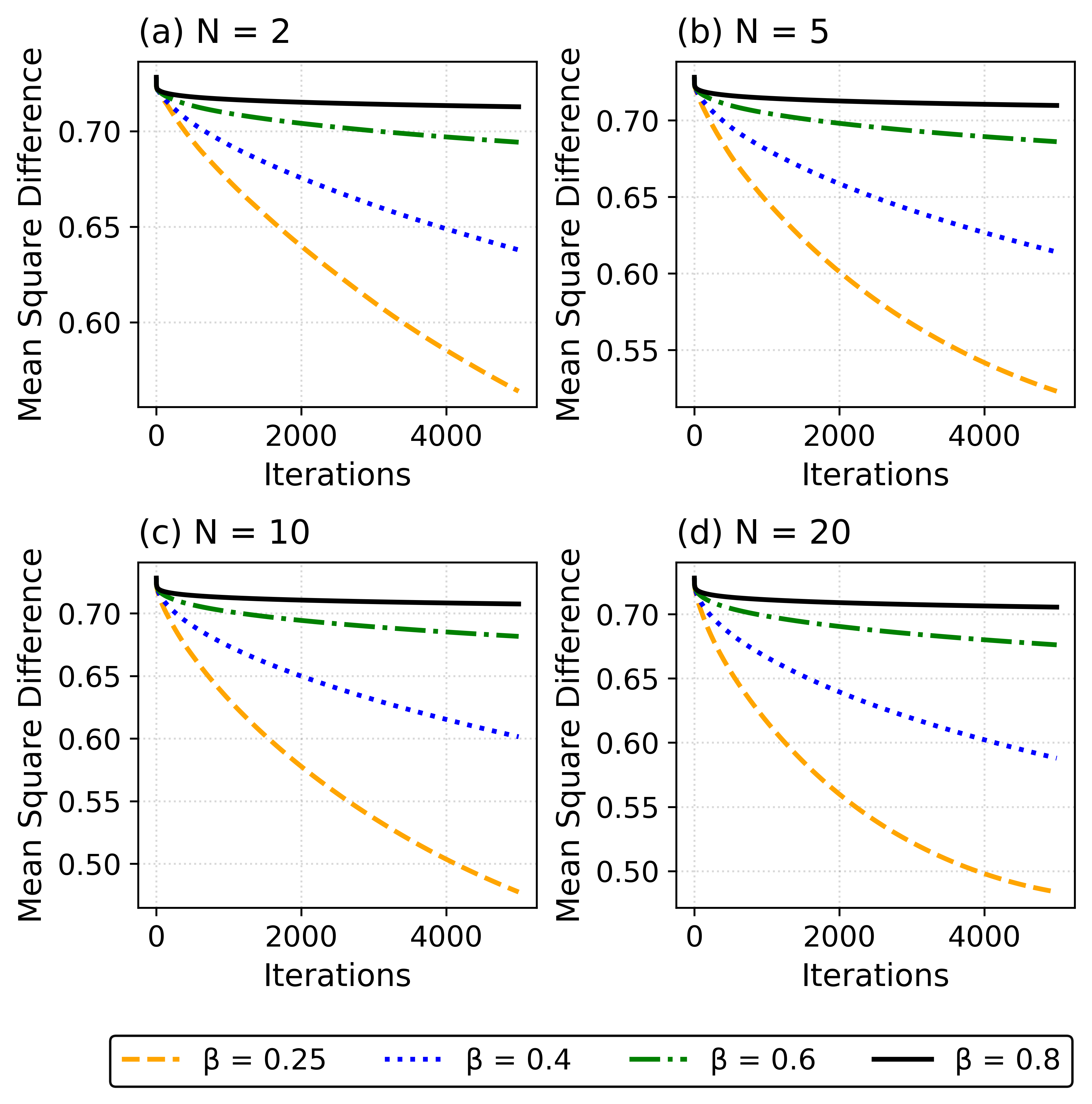}
    \caption{Comparison of different $\beta$ values across the same number of agents executing Algorithm 1 in a heterogeneous Markovian setting.}
    \label{fig:beta.avg}
\end{figure}

\begin{figure}
    \includegraphics[width=\columnwidth]{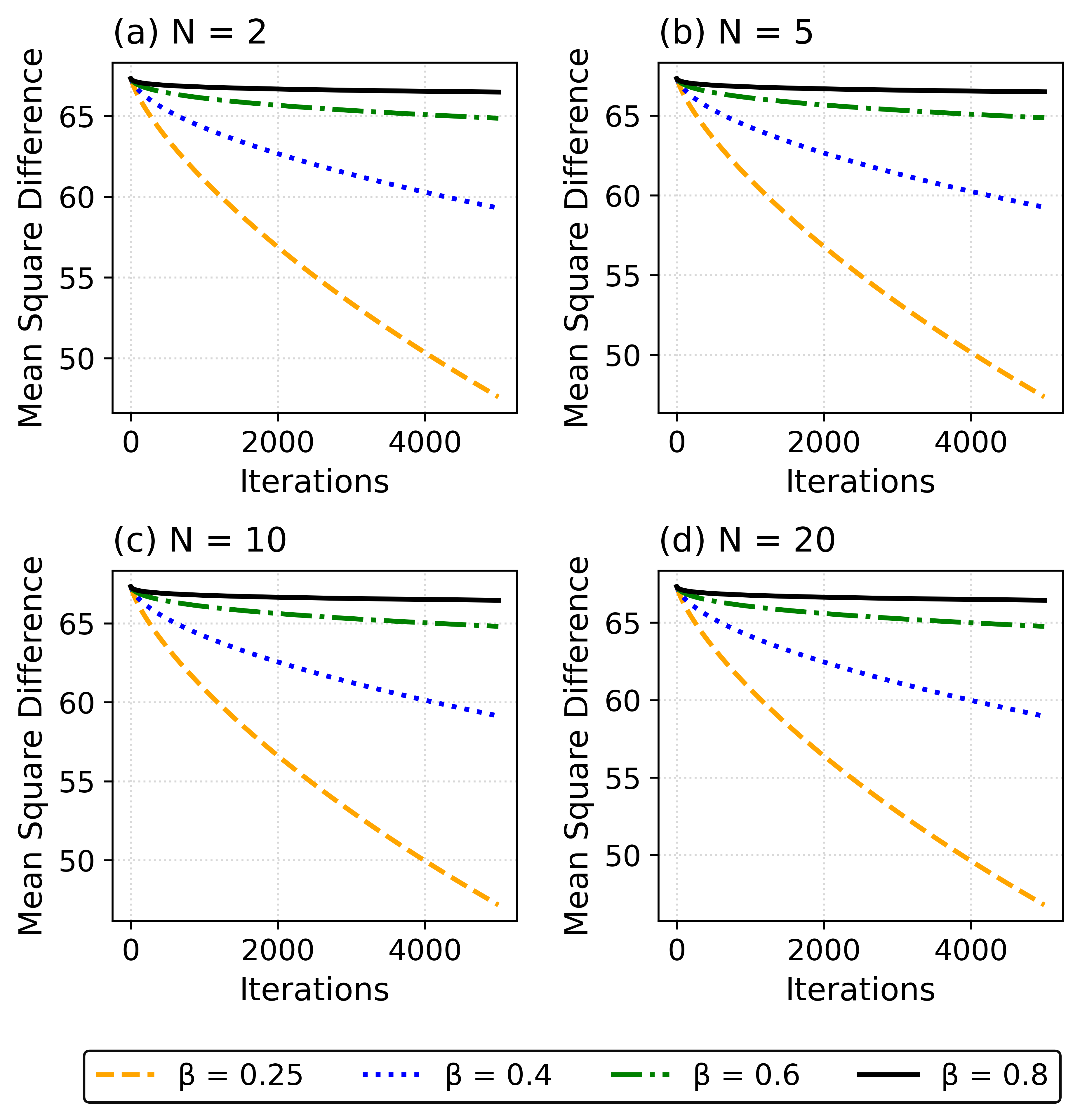}
    \caption{Comparison for different $\beta$ values with a fixed set of agents executing Algorithm 2 in a heterogeneous Markovian setting.}
    \label{fig:beta.exp}
\end{figure}
The second experiment shows the trend of how the error decays for Algorithms 1 and 2 when the stepsize parameter $\beta$ is set to $0.2, 0.4, 0.6,$ and $0.8.$ Other parameters are as in the previous experiment. The plot in Figure~\ref{fig:beta.avg} shows the results for AvgFedTD(0) and that in Figure~\ref{fig:beta.exp} for ExpFedTD(0). Clearly, the convergence rate decreases with an increase in $\beta.$ Based on these plots, we conjecture that the optimal convergence rate---in terms of the constants in the $O(\frac{1}{NT})$ bound---is achieved at $\beta = 0;$ i.e., when $\beta_t$ is held constant (as a function of $t$). However, it is unclear if this choice of constant will be parameter free. 

\begin{figure}
    \centering
    \setlength{\tabcolsep}{4pt}
    \begin{tabular}{cc}
        \includegraphics[width=0.49\columnwidth, keepaspectratio]{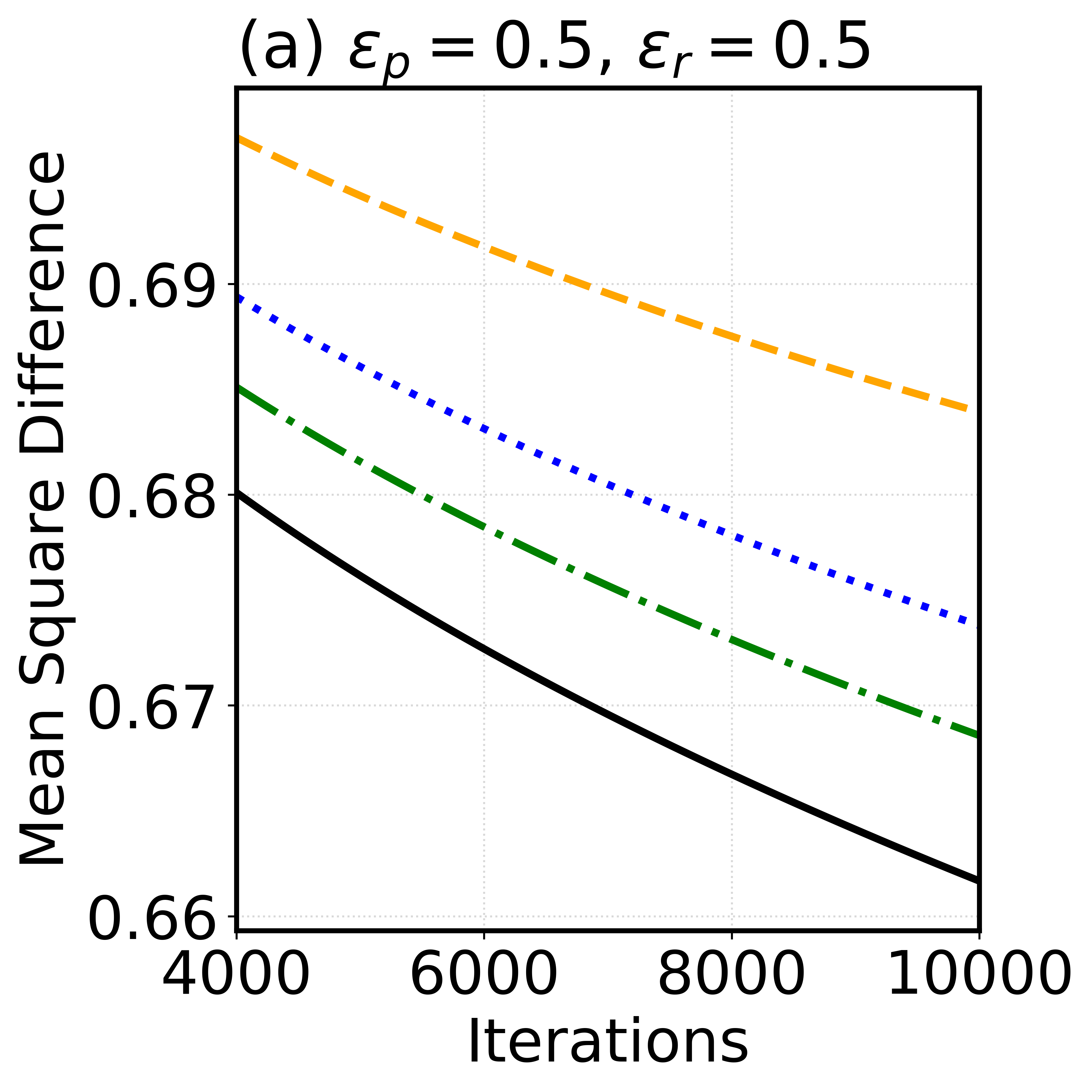} &
        \includegraphics[width=0.49\columnwidth, keepaspectratio]{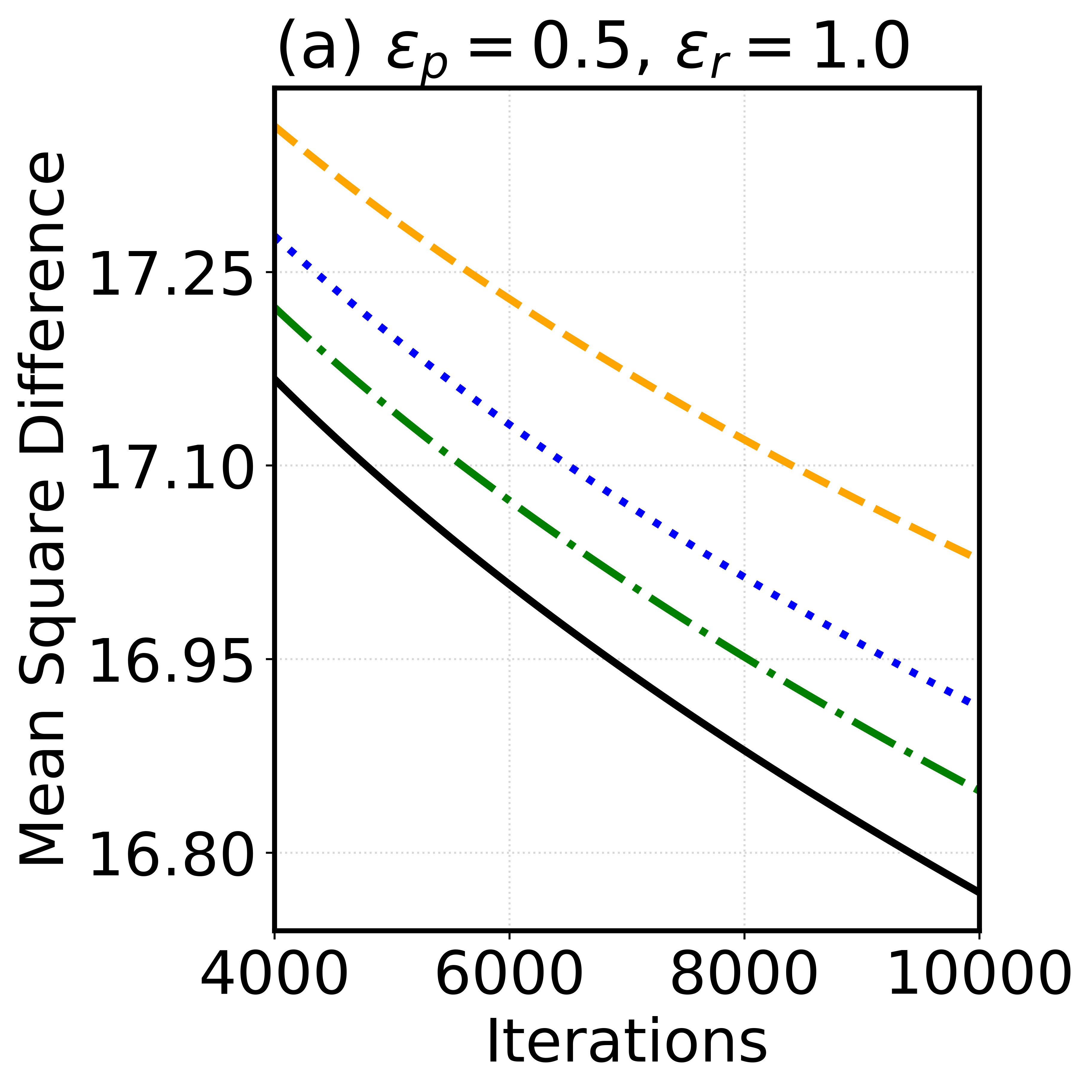} \\[0.25cm]
        
        \includegraphics[width=0.49\columnwidth, keepaspectratio]{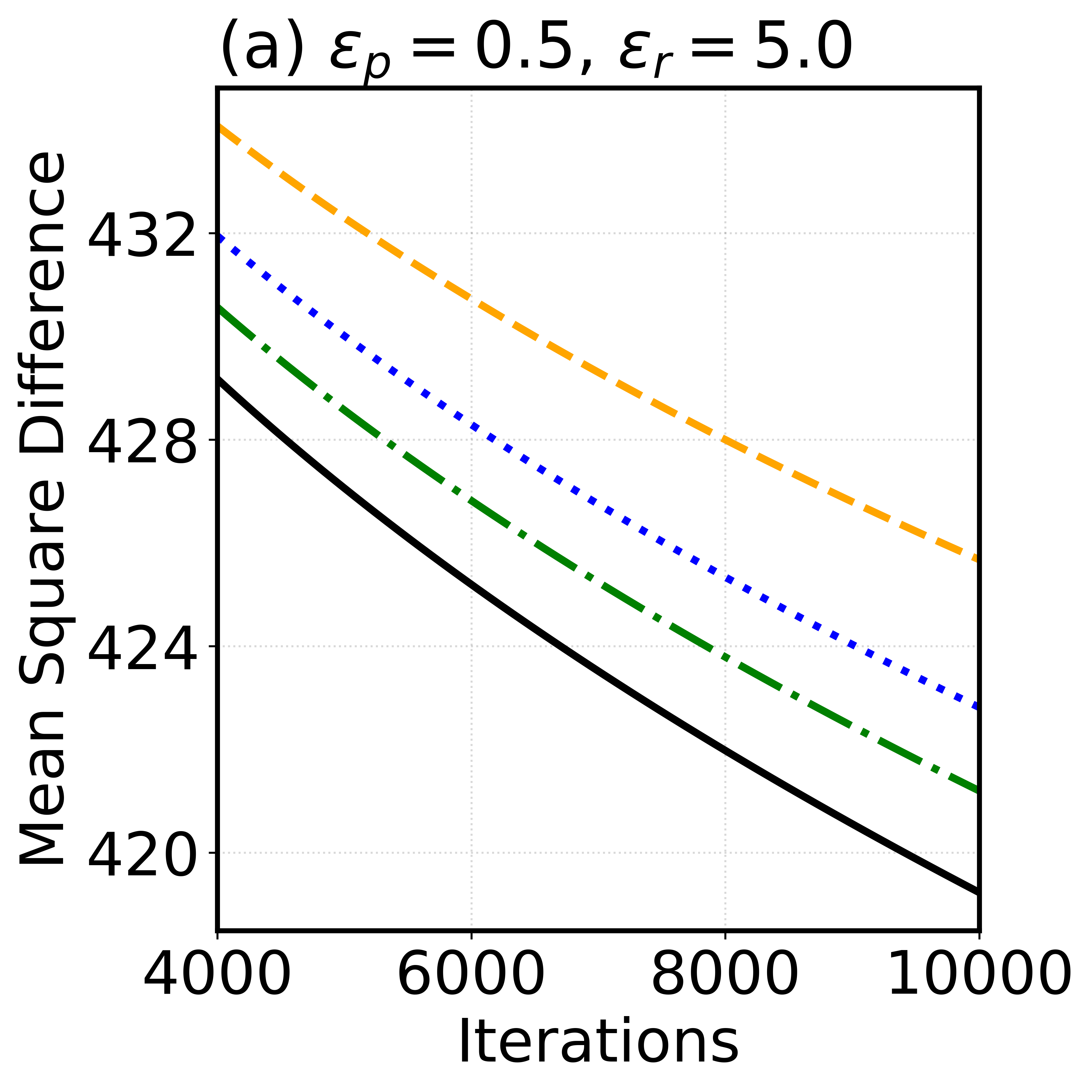} &
        \includegraphics[width=0.49\columnwidth, keepaspectratio]{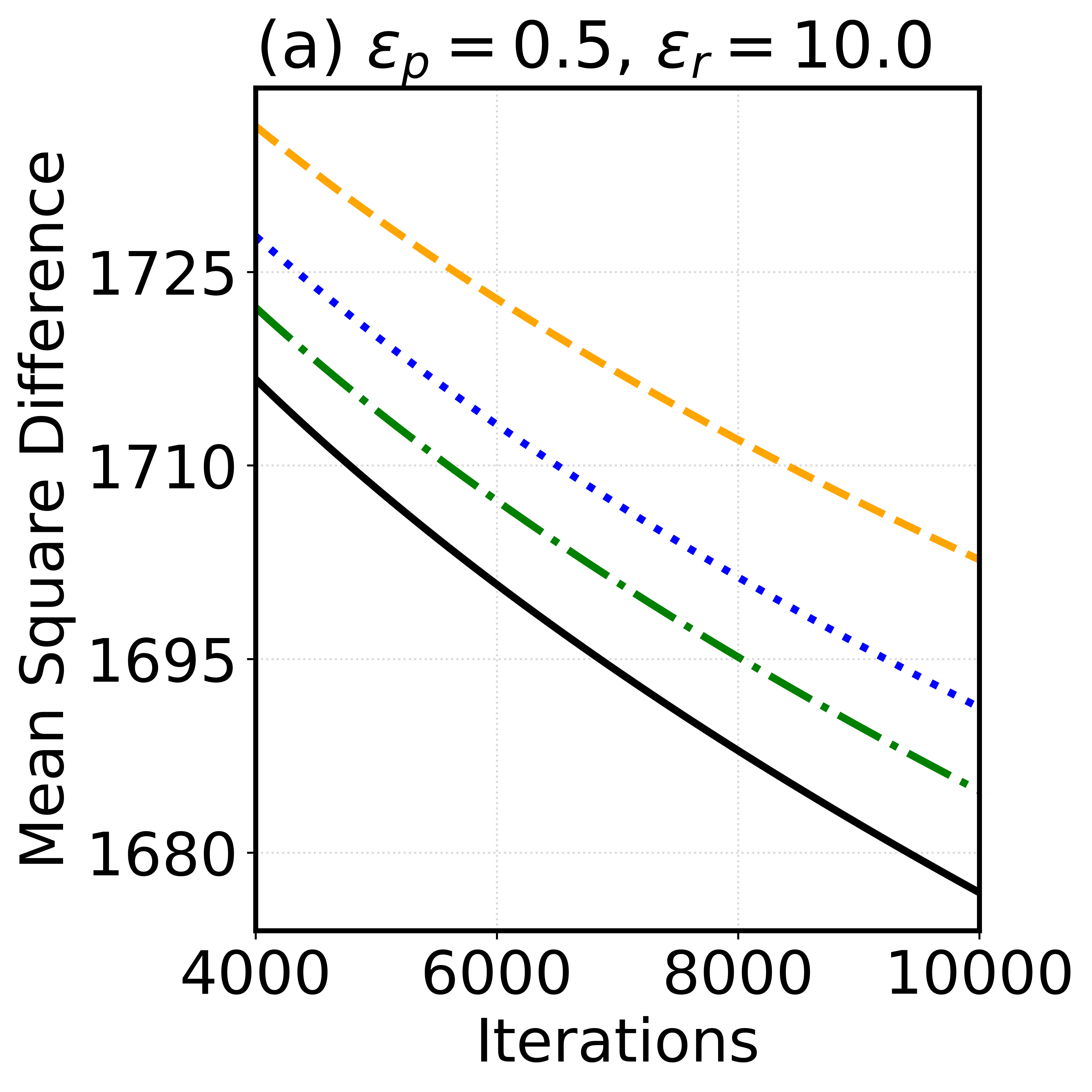} \\
    \end{tabular}

    \vspace{0.1cm}
    \includegraphics[width=\columnwidth]{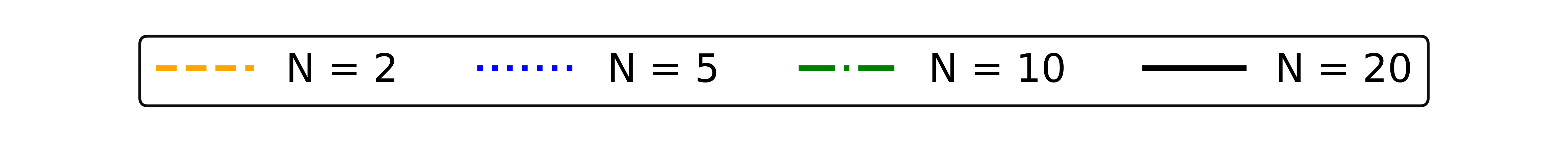}
    
    \caption{Comparison of simulation results executing Algorithm 1 with different values of $\er$ in a heterogeneous Markovian setting.}
    \label{fig:er.avg}
\end{figure}

\begin{figure}
    \centering
    \setlength{\tabcolsep}{4pt}
    \begin{tabular}{cc}
        \includegraphics[width=0.49\columnwidth, keepaspectratio]{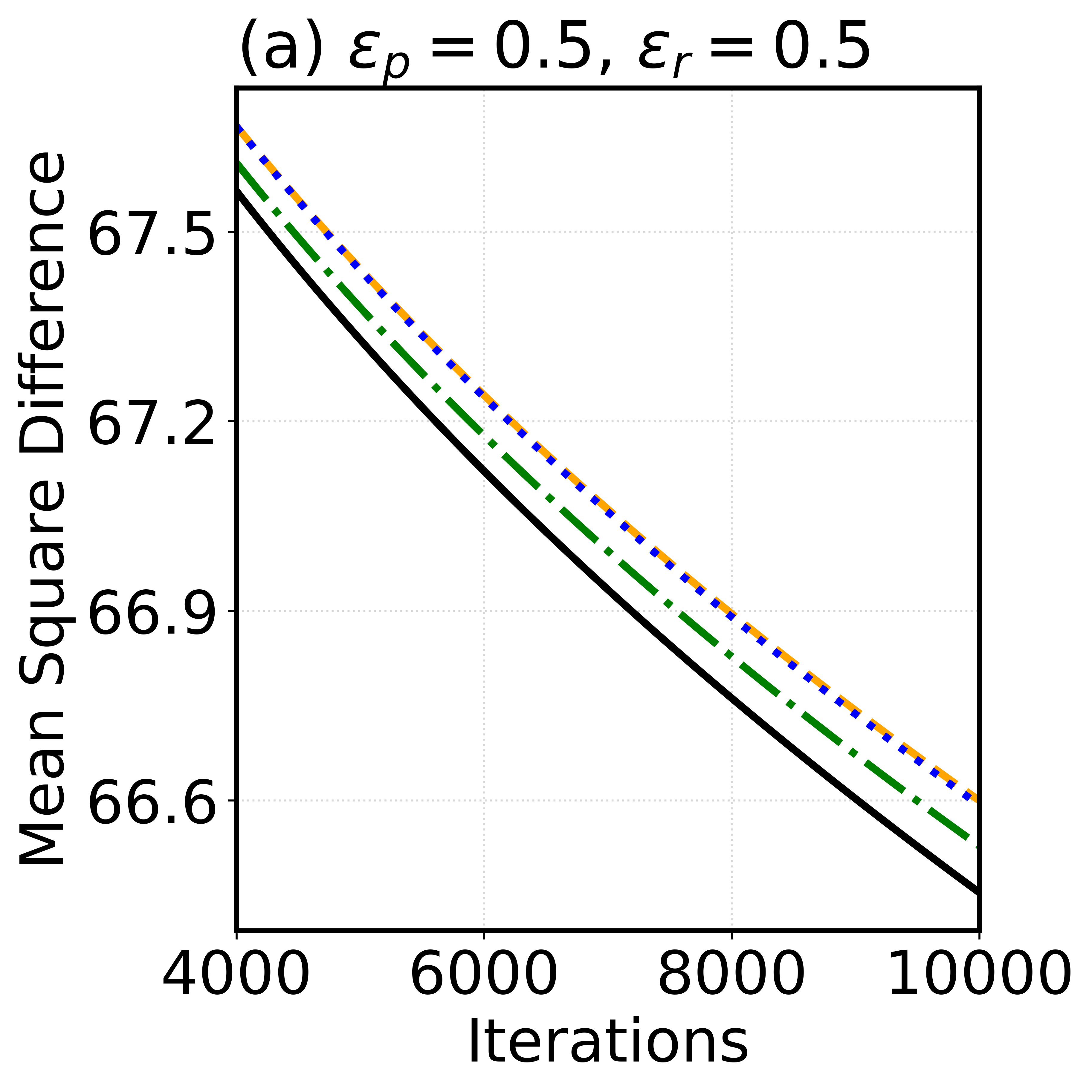} &
        \includegraphics[width=0.49\columnwidth, keepaspectratio]{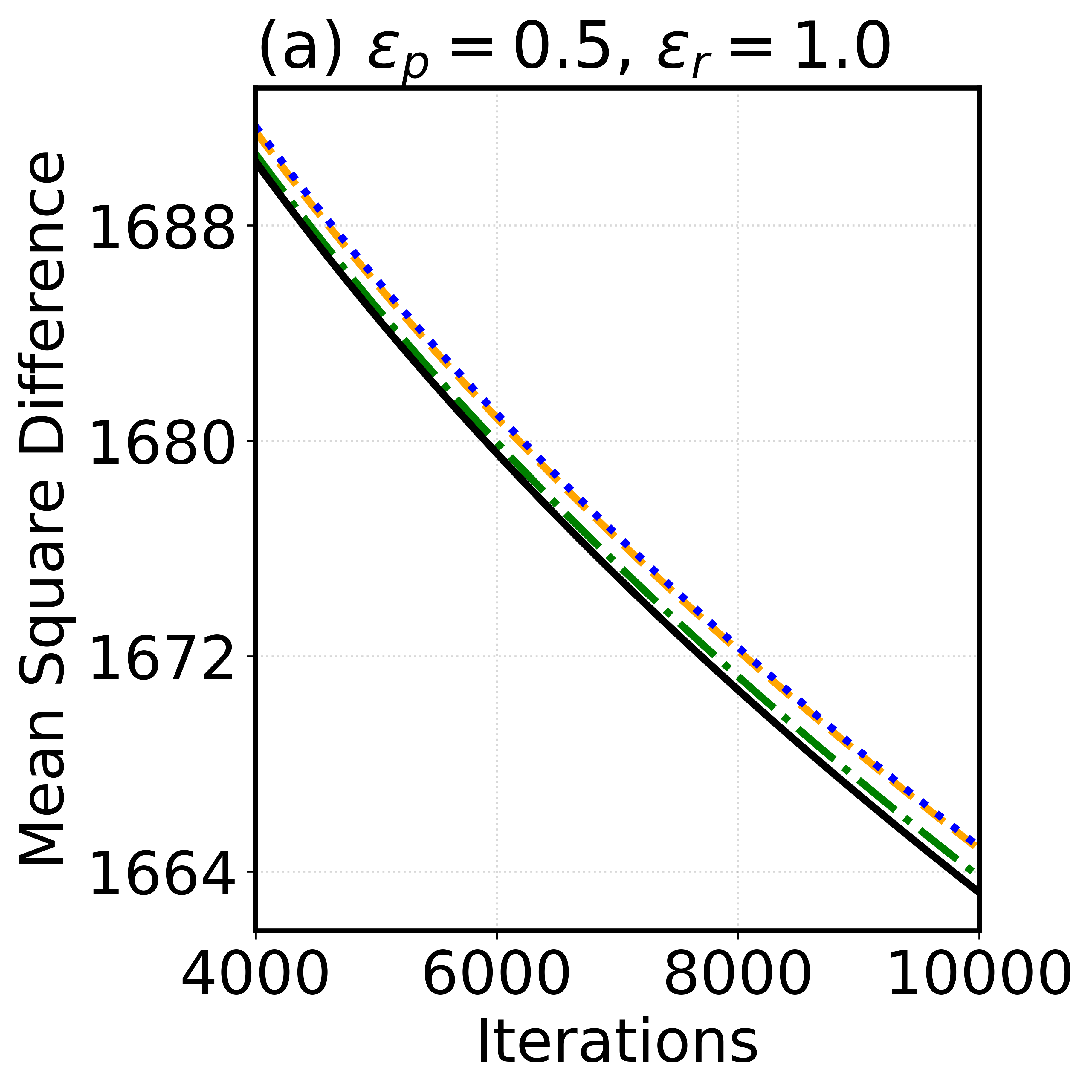} \\[0.25cm]
        
        \includegraphics[width=0.49\columnwidth, keepaspectratio]{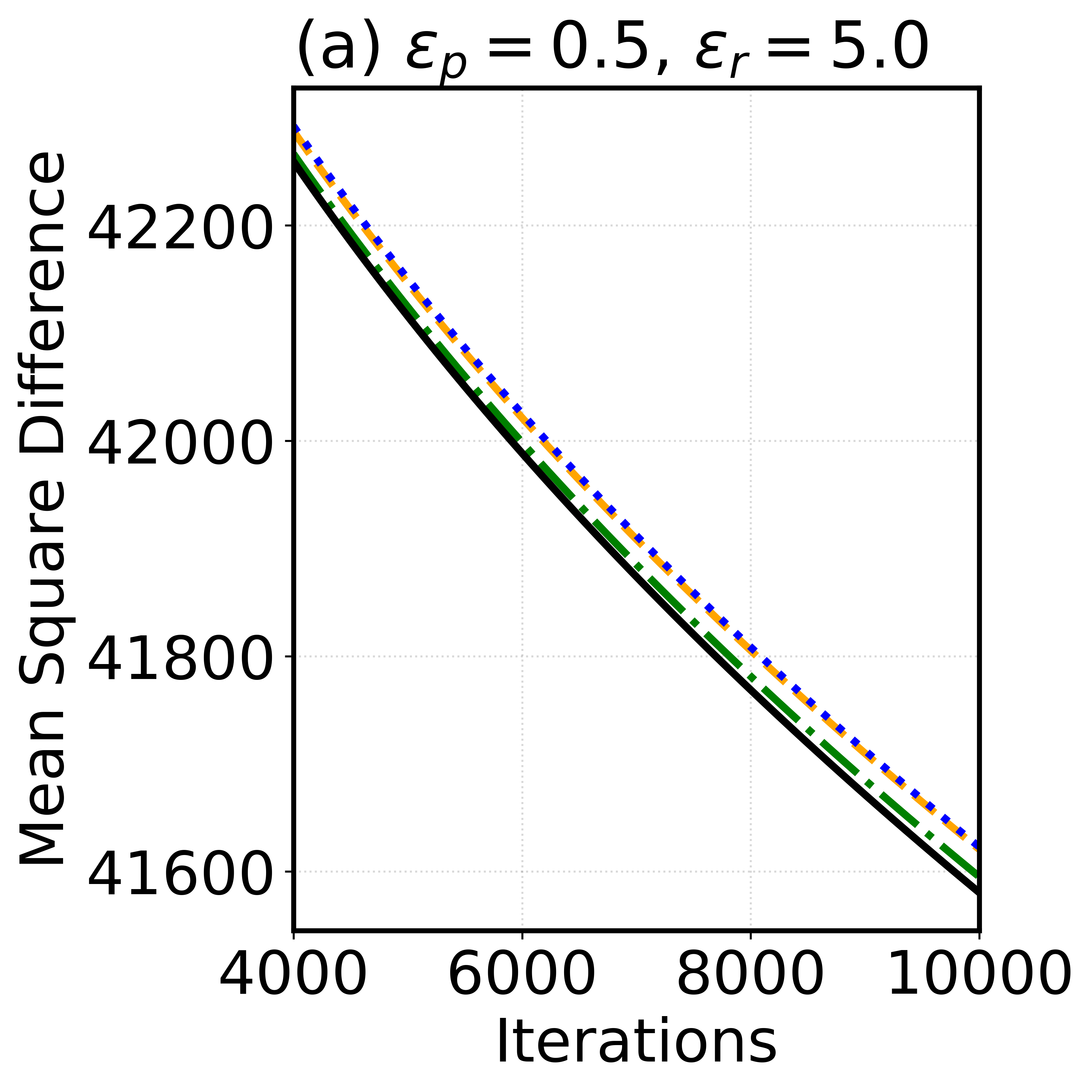} &
        \includegraphics[width=0.49\columnwidth, keepaspectratio]{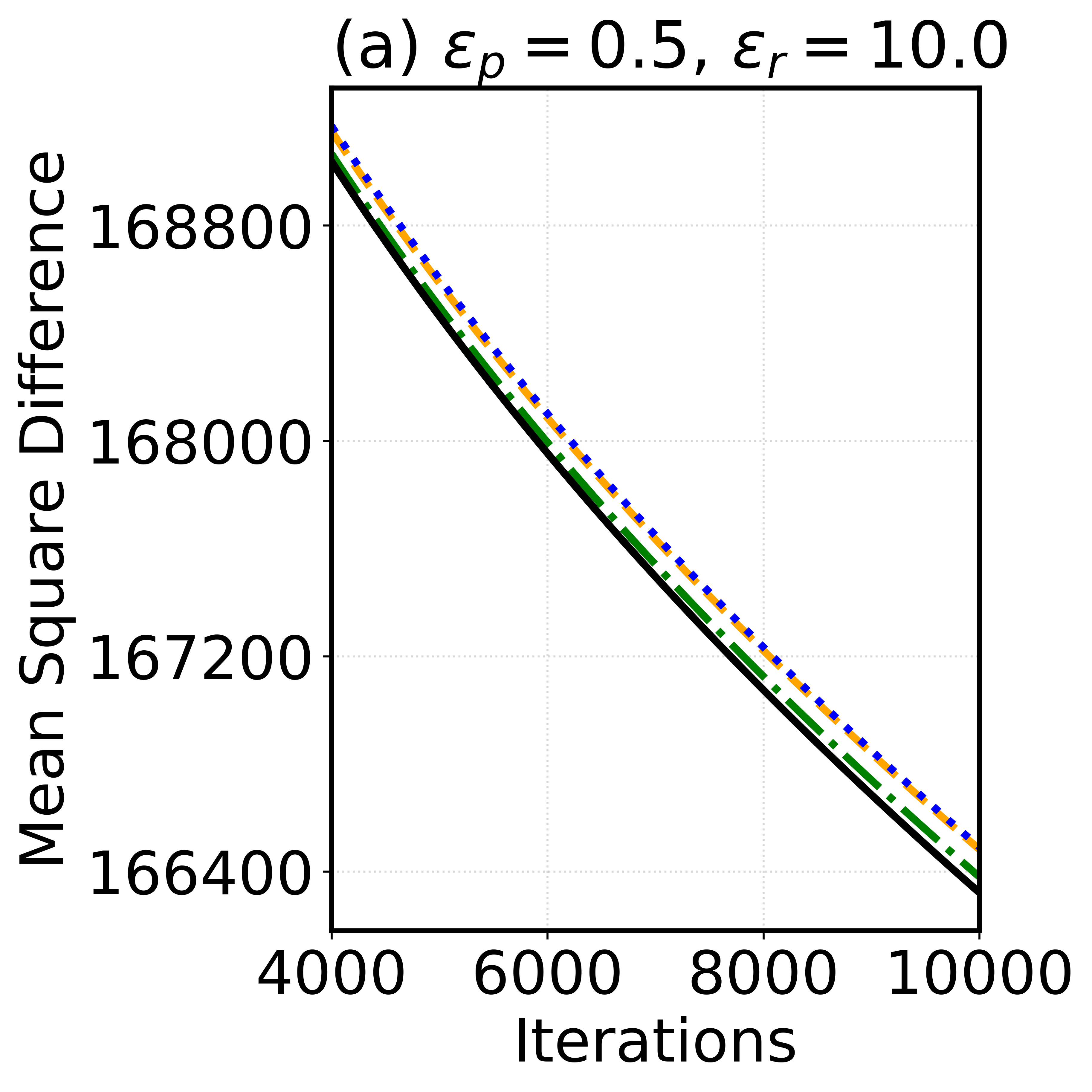} \\
    \end{tabular}

    \vspace{0.1cm}
    \includegraphics[width=\columnwidth]{IEEE-TAC-Submission/Plots/agents_legend.png}
    \caption{Comparison of simulation results executing Algorithm 2 with different values of $\er$ in a heterogeneous Markovian setting.}
    \label{fig:er.exp}
\end{figure}

The third experiment shows the effect of modifying $\er$. In these simulations, we pick $\er$ to be one of $0.5, 1, 5,$ and $10.$ We set $\Rm$ to $5\er$ when $\er\geq 1,$ and to $1$ otherwise. The other parameters are set as in previous experiments. % while the number of agents is set to $2, 5, 10,$ and $20.$ 
Figure~\ref{fig:er.avg} shows the results for AvgFedTD(0) and Figure~\ref{fig:er.exp} for ExpFedTD(0). Note that as $\er$ increases, the mean squared error's behavior is similar in shape but occurs from a higher initial value.

\begin{figure}[!htbp]
    \centering
    \begin{tabular}{cc}
        \includegraphics[width=0.48\columnwidth, keepaspectratio]{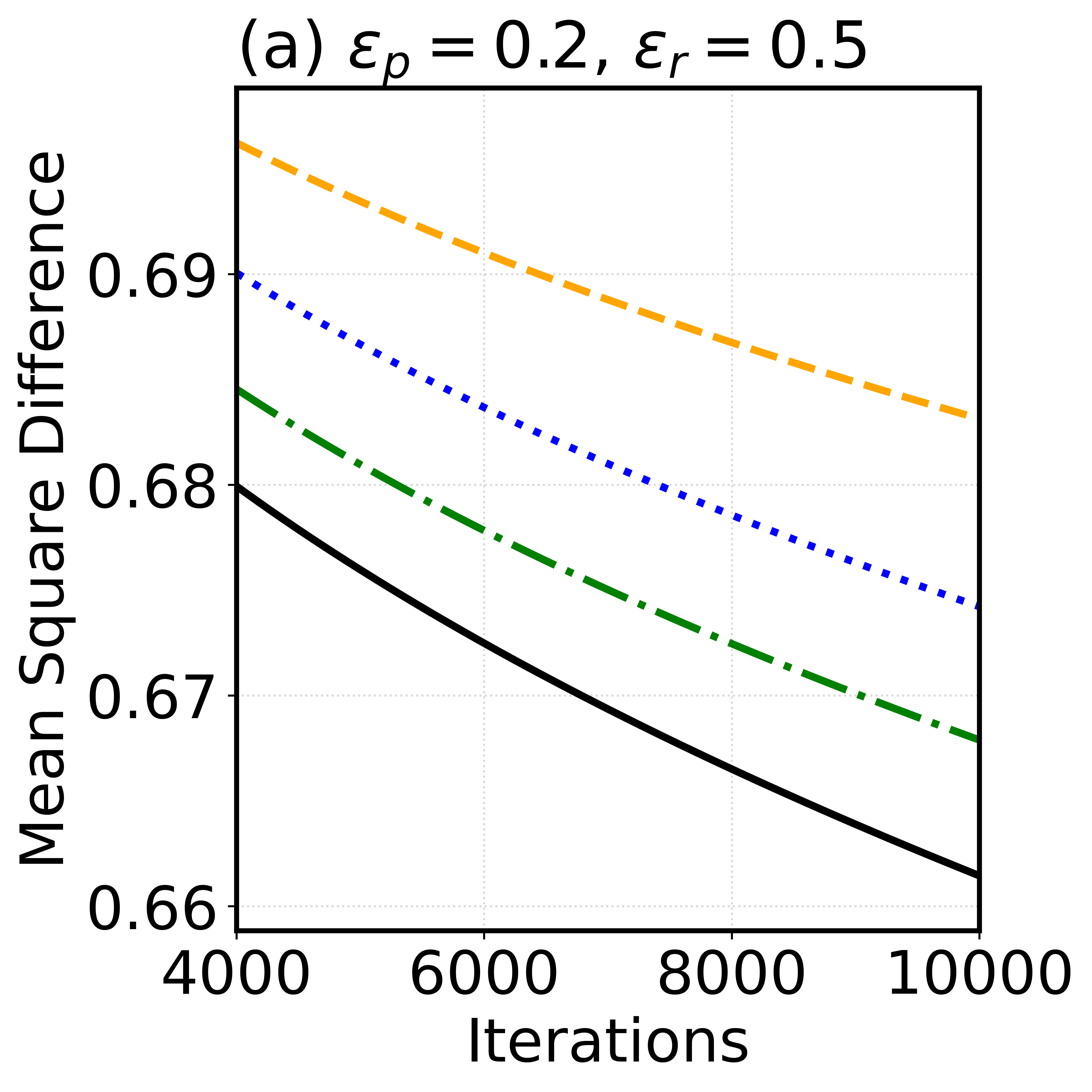} &
        \includegraphics[width=0.48\columnwidth, keepaspectratio]{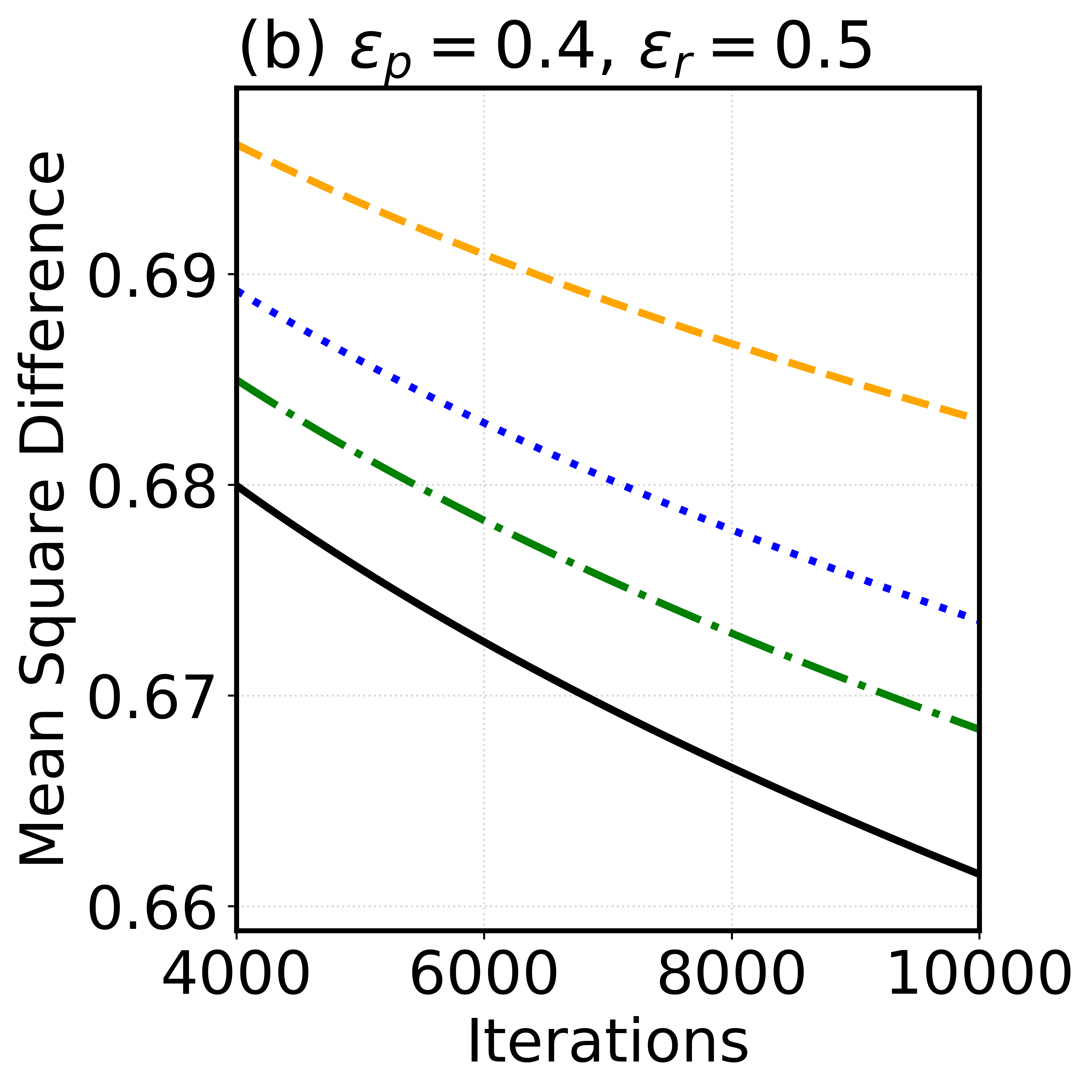} \\[0.3cm]
        
        \includegraphics[width=0.48\columnwidth, keepaspectratio]{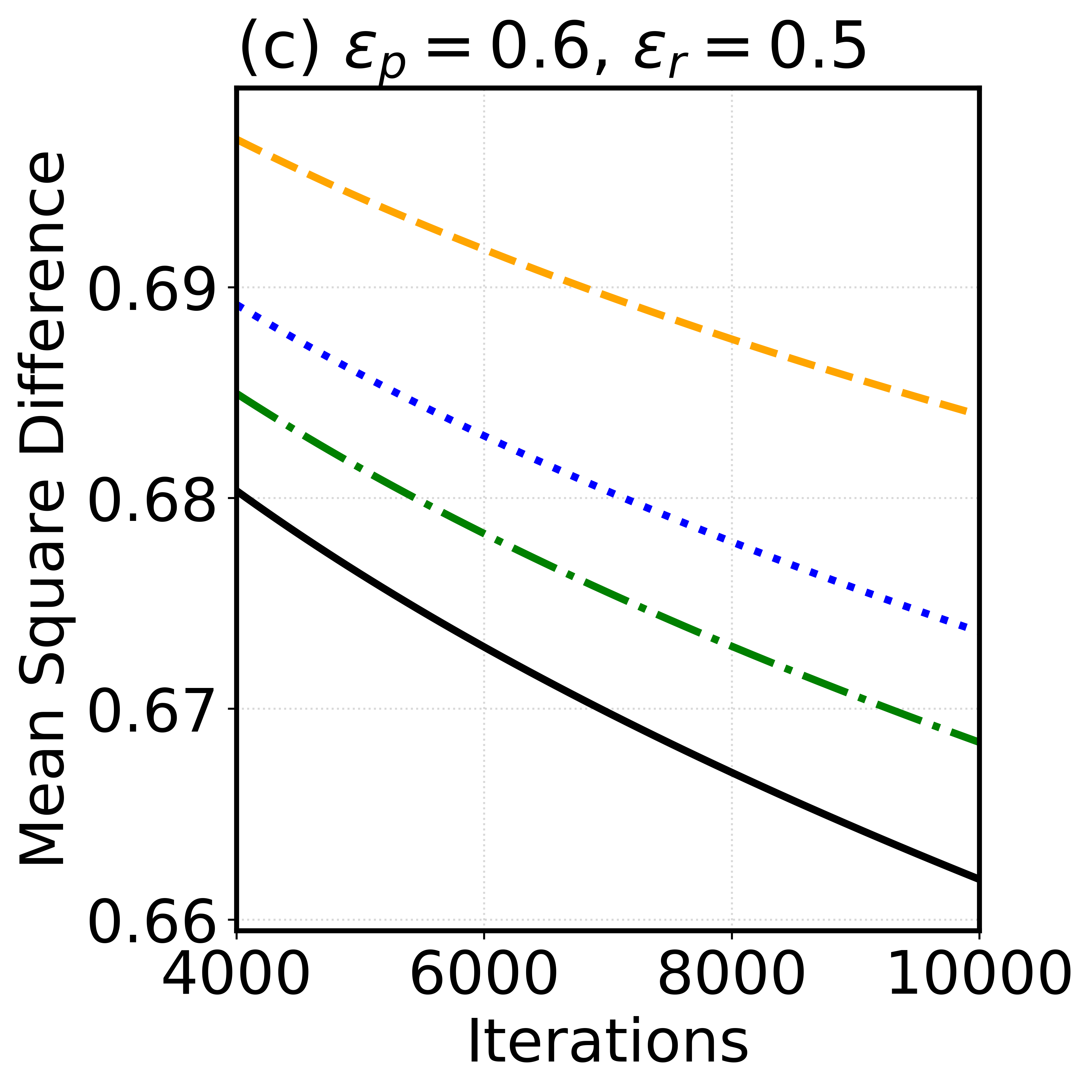} &
        \includegraphics[width=0.48\columnwidth, keepaspectratio]{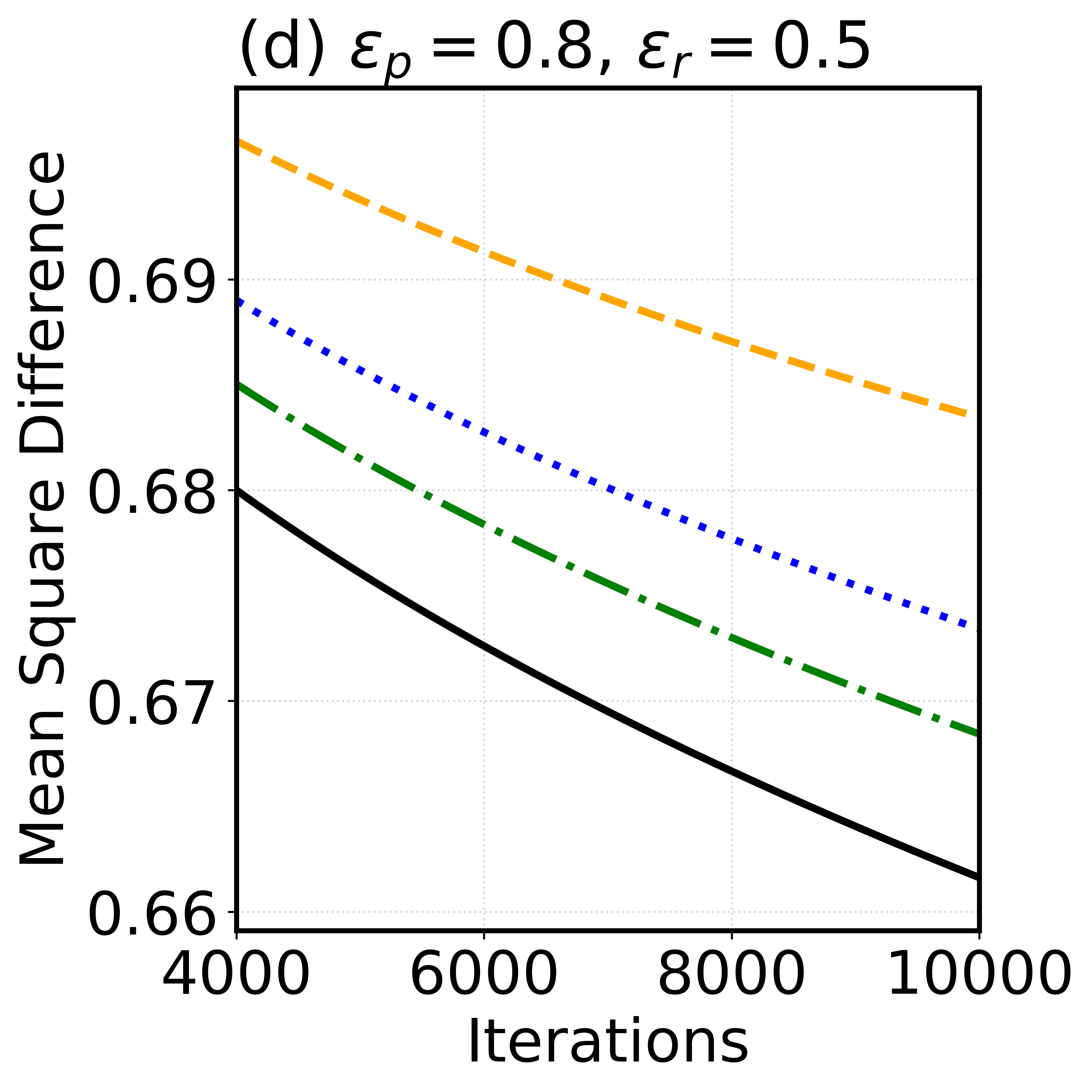} \\
    \end{tabular}

    \vspace{0.1cm}
    \includegraphics[width=\columnwidth]{IEEE-TAC-Submission/Plots/agents_legend.png}
    
    \caption{Comparison of simulation results executing Algorithm 1 with different $\epsilon_p$ values.}
    \label{fig:ep.avg}
\end{figure}

\begin{figure}
    \centering
    \setlength{\tabcolsep}{4pt}
    \begin{tabular}{cc}
        \includegraphics[width=0.49\columnwidth, keepaspectratio]{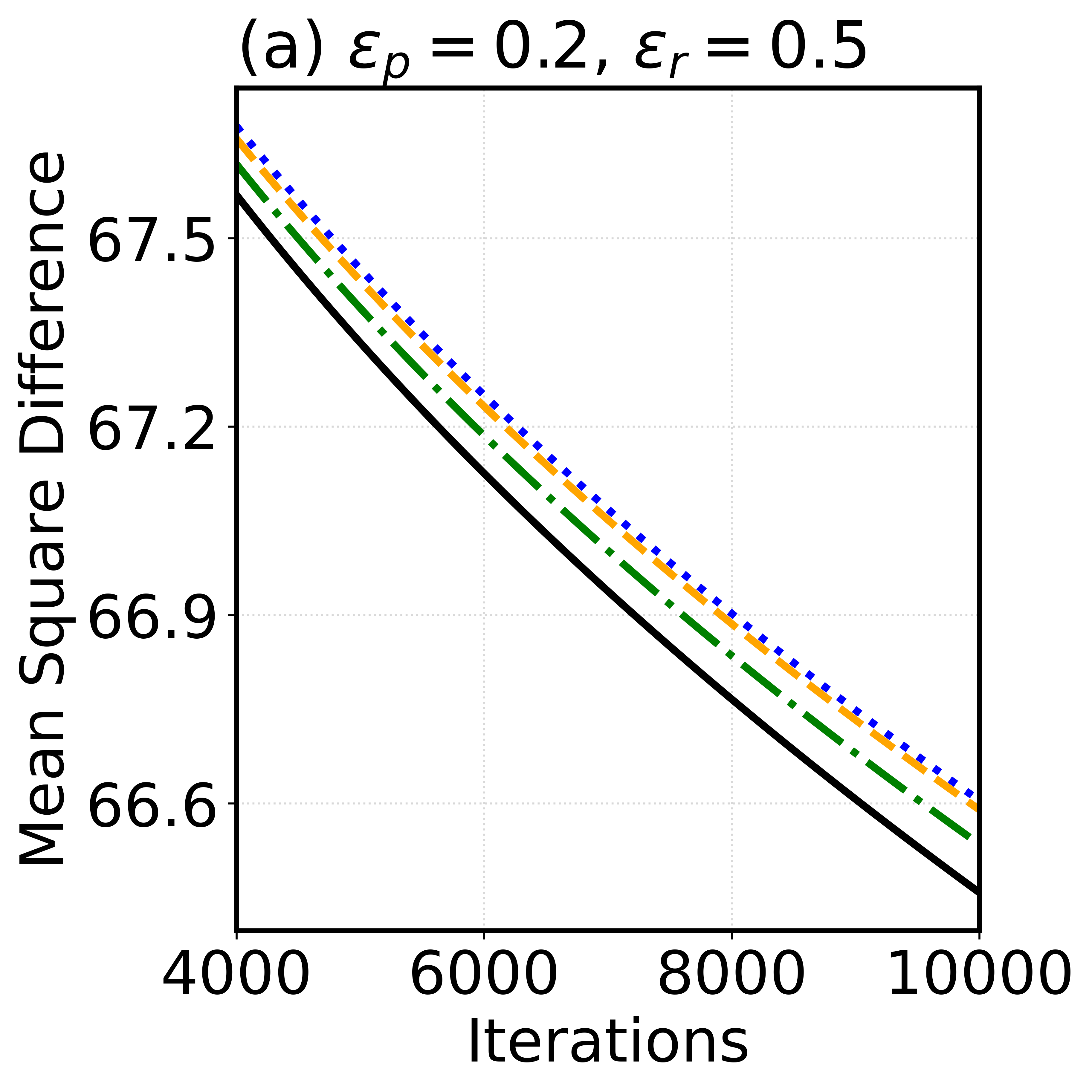} &
        \includegraphics[width=0.49\columnwidth, keepaspectratio]{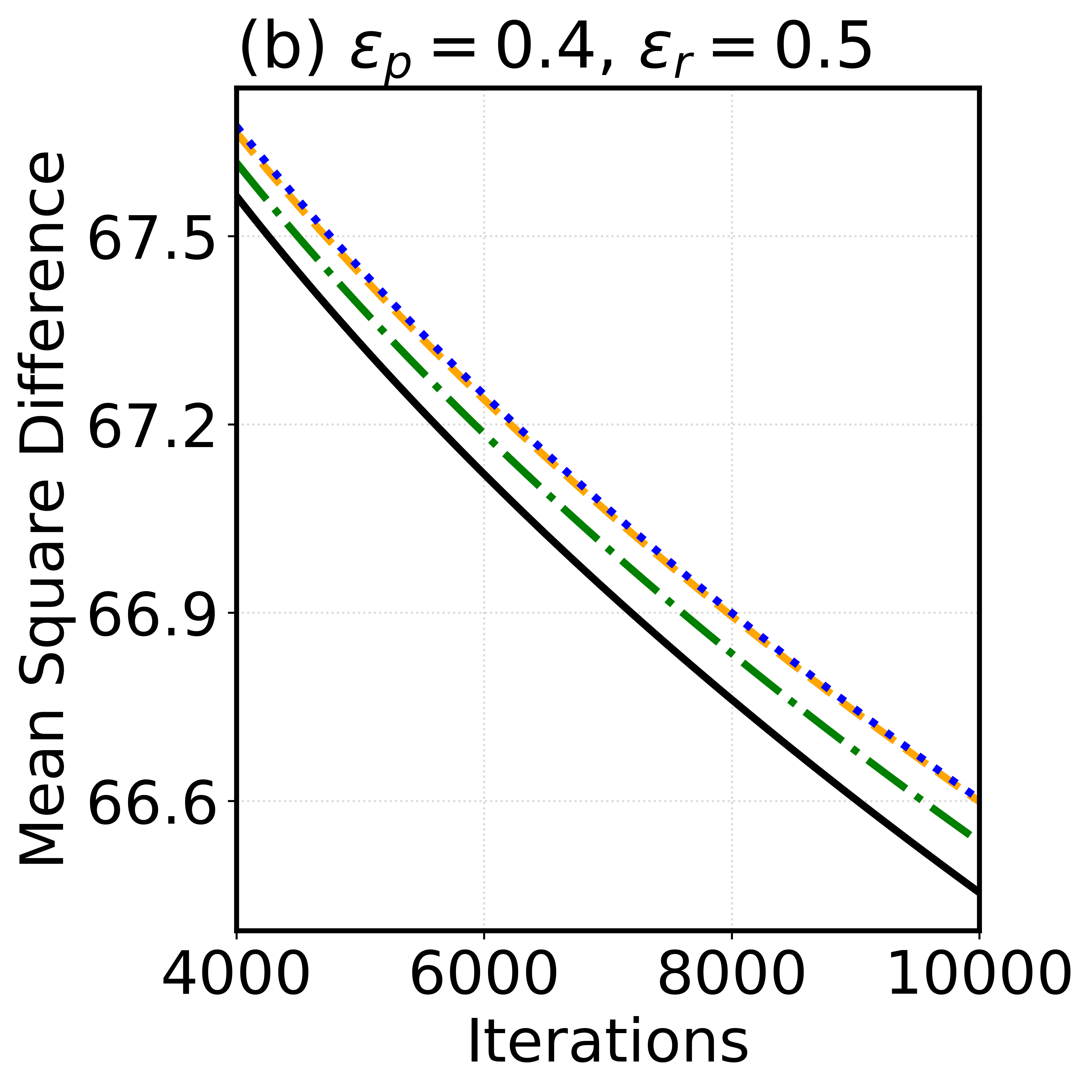} \\[0.25cm]
        \includegraphics[width=0.49\columnwidth, keepaspectratio]{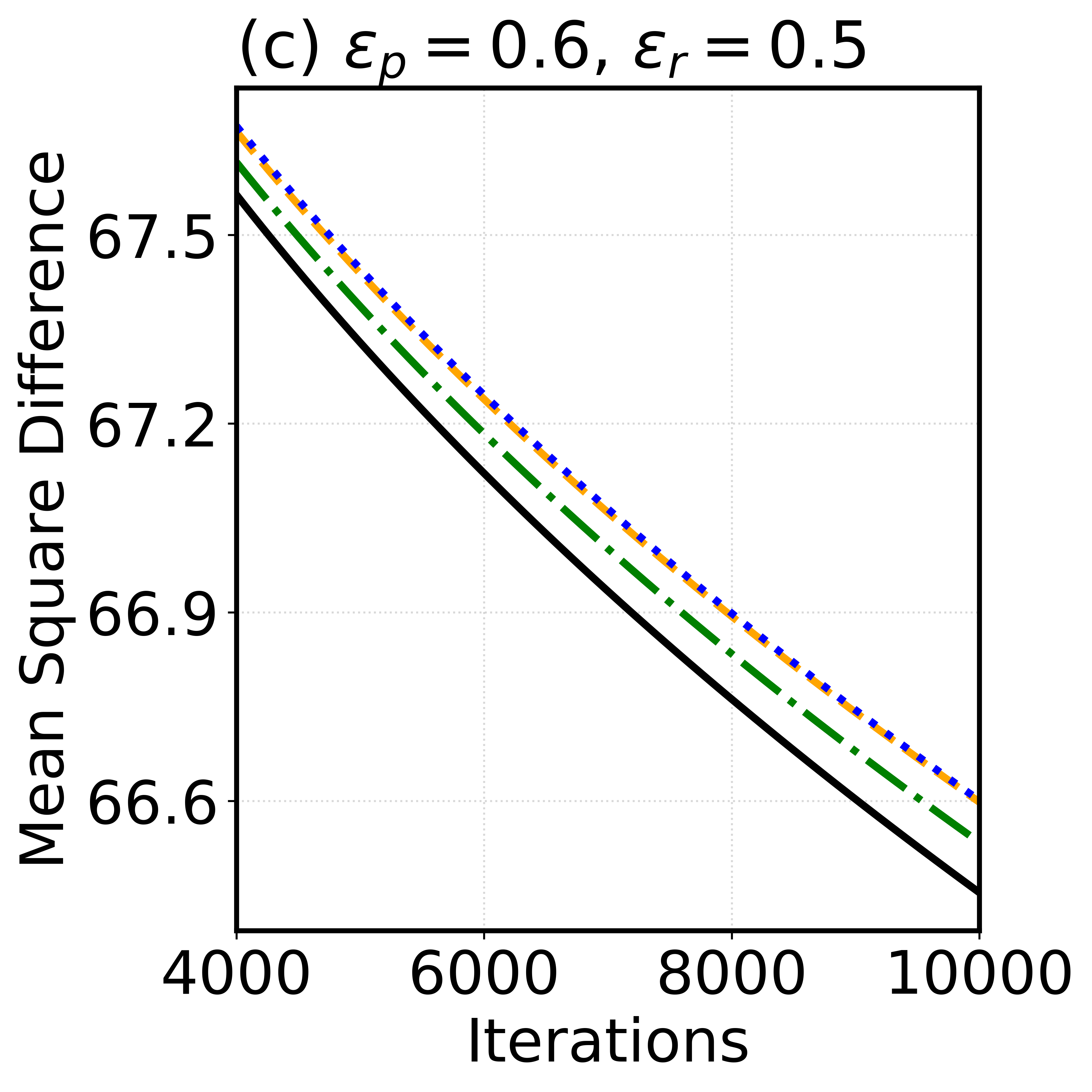} &
        \includegraphics[width=0.49\columnwidth, keepaspectratio]{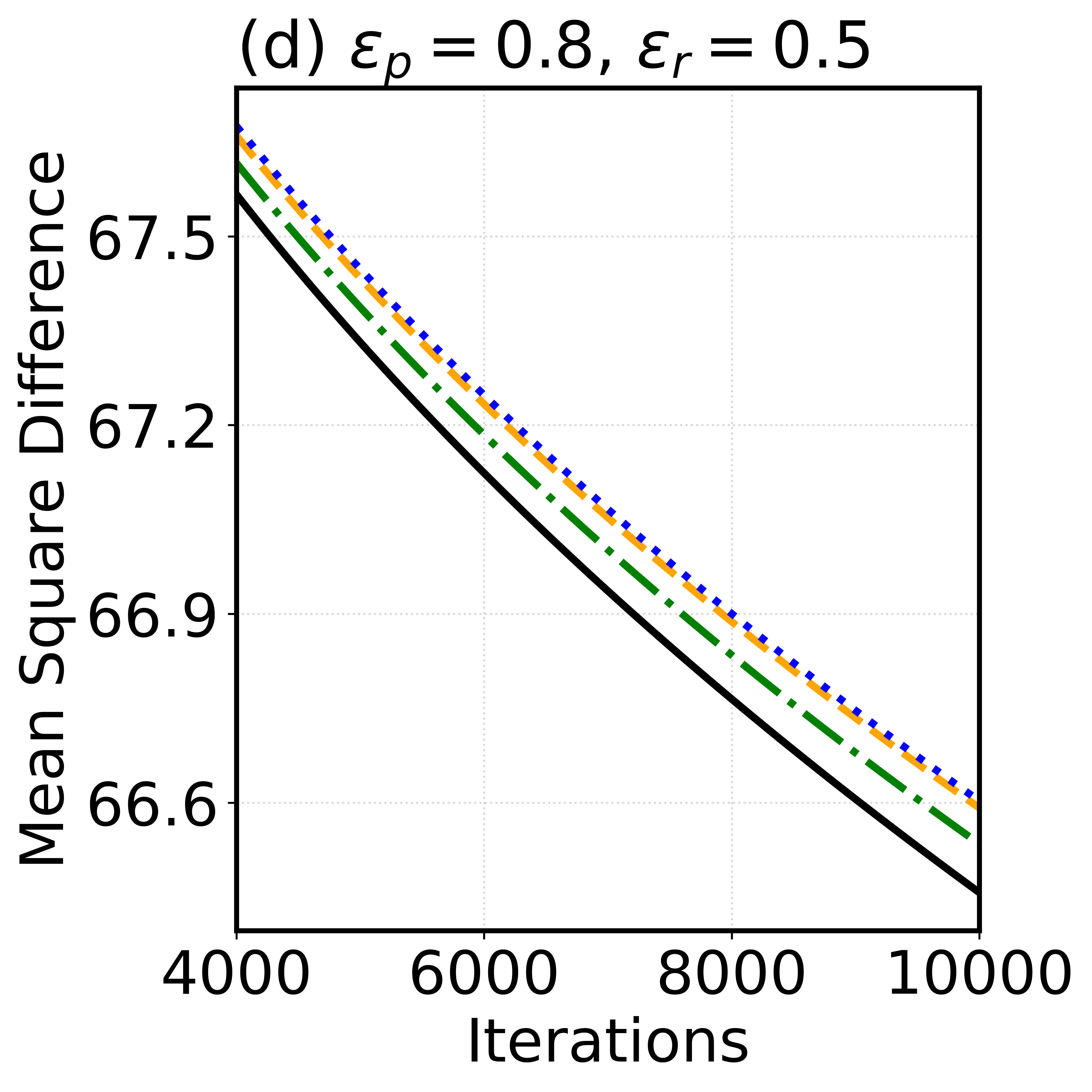} \\
    \end{tabular}

    \vspace{0.1cm}
    \includegraphics[width=\columnwidth]{IEEE-TAC-Submission/Plots/agents_legend.png}
    \caption{Comparison of simulation results executing Algorithm 2 with different $\ep$ values in a heterogeneous Markovian setting.}
    \label{fig:ep.exp}
\end{figure}

The next experiment shows the effect of $\ep$ on the two Algorithms. We set $\ep$ equal to $0.2, 0.4, 0.6,$ and $0.8.$ Figure~\ref{fig:ep.avg} shows the results for AvgFedTD(0) and Figure~\ref{fig:ep.exp} for ExpFedTD(0). We see no major differences in performance across different $\ep$ choices.

\begin{figure}
    \centering
    \includegraphics[width=\columnwidth]{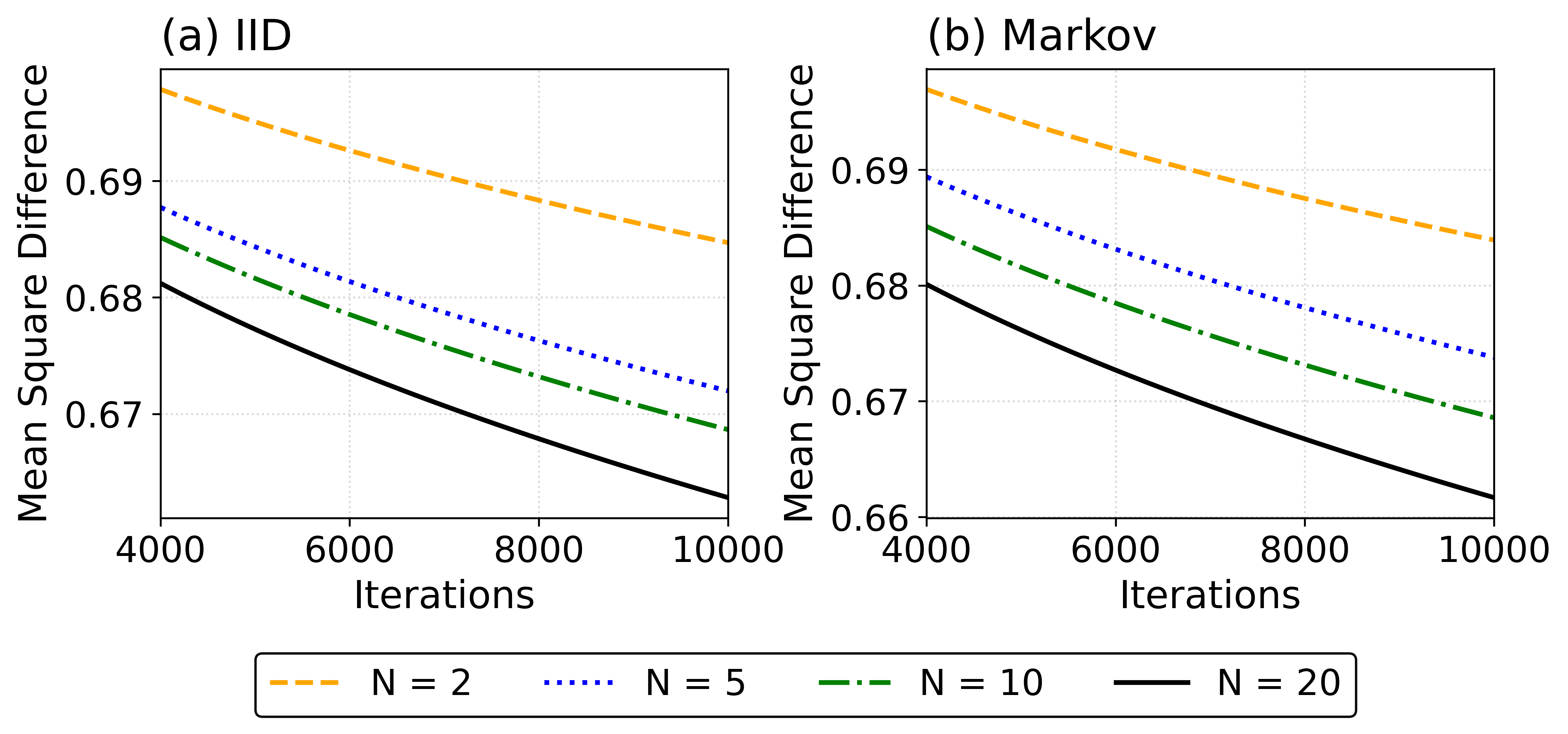}
    \caption{Algorithm 1 in an IID setting vs in a Markovian setting.}
    \label{fig:iid_vs_markov.avg}
\end{figure}

\begin{figure}
    \centering
    \includegraphics[width=\columnwidth]{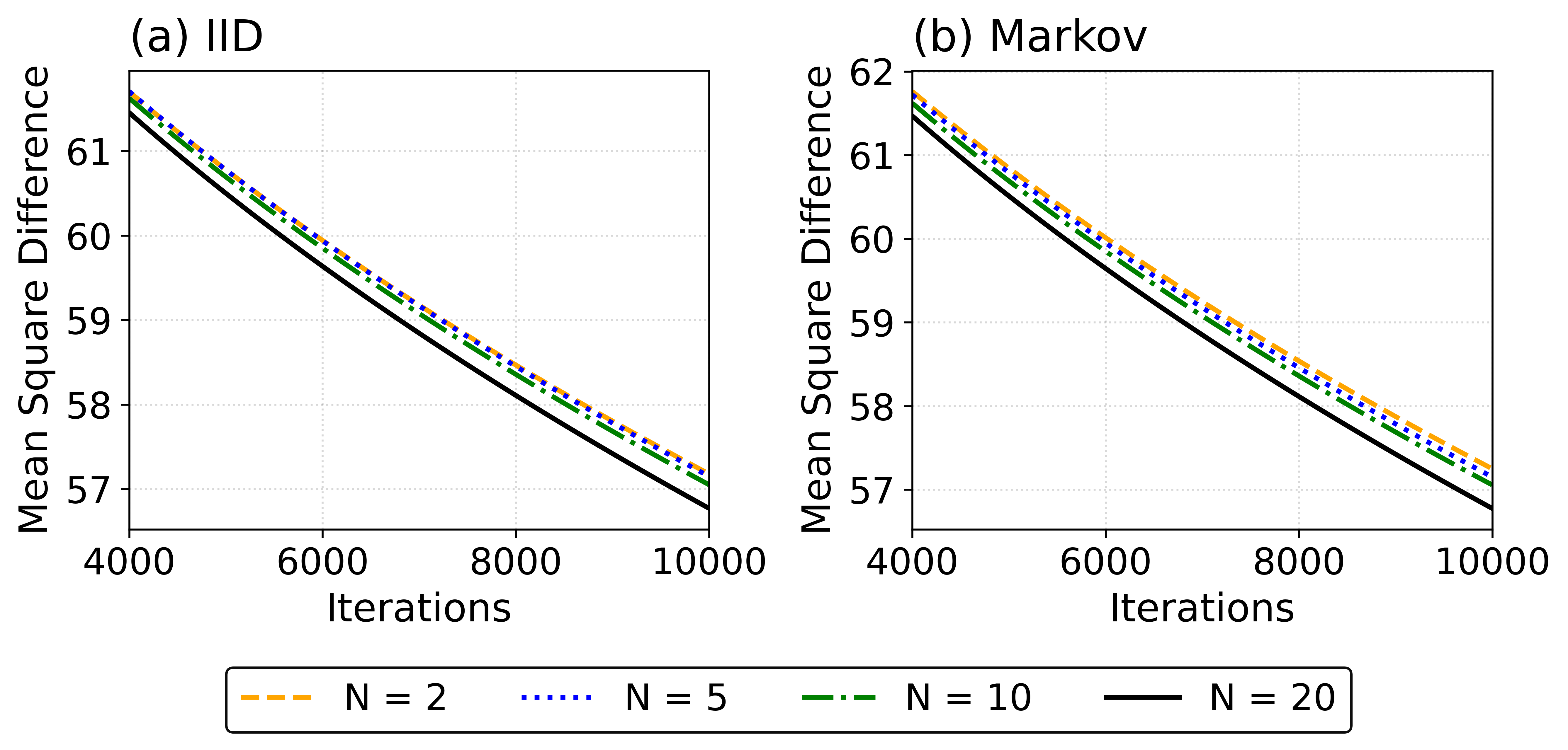}
    \caption{Algorithm 2 in an IID setting vs in a Markovian setting.}
    \label{fig:iid_vs_markov.exp}
\end{figure}

The final experiment compares the performance of the proposed algorithms in an IID setting vs in a Markovian setting. In the IID setting, at each iteration, the TD update direction is computed using a state-action-state tuple generated as follows: a current state is sampled from the stationary distribution of the Markov chain induced by the behavior policy followed by sampling an action from this policy and observing the resulting state transition. Figure~\ref{fig:iid_vs_markov.avg} plots the results for AvgFedTD(0) and~\ref{fig:iid_vs_markov.exp} for ExpFedTD(0). We see no major change in performances. 

%
% avg_markoviid_runs_350_rounds_10000_states_100_actions_100_d_21_beta_0.6_r_0.5_p_0.5.png
%

\section{Conclusion and Future Directions}

RL often faces criticism for the time it takes to explore the policy space, especially when the state and action spaces are large. FRL offers a promising solution, providing linear speedups by leveraging multiple agents, even when their MDPs are heterogeneous. In this work, we show that, by incorporating PR-averaging, optimal convergence rates  can be obtained for federated TD algorithms in both the average-reward and discounted settings without relying on problem-specific step sizes. Importantly, we show that these rates can be achieved in the realistic and more challenging scenario of asynchronous updates with Markovian sampling. 

For future work, we plan to extend these techniques to asynchronous federated Q-learning  with PR-averaging in the average-reward setting, by combining the results of this paper with those for the synchronous case in \cite{naskar2025parameter}. A key challenge is that Assumption~\ref{a: feature.matrix}---specifically, the requirement that the all-ones vector $\ones$ not lie in the column space of $\Phi$---cannot be guaranteed. Consequently, the associated Bellman operator no longer admits a unique fixed point. More broadly, we aim to address this in the function-approximation setting. However, \cite{gopalan2024should} raises concerns: Q-learning with linear function approximation and $\epsilon$-greedy exploration, if it converges, may reach a fixed point of the projected Bellman operator whose greedy policy can be suboptimal---or even the worst policy. A promising alternative is to explore model-free variants of reliable policy iteration \cite{eshwar2025reliable}, which retain the monotonicity and convergence guarantees of tabular policy iteration under arbitrary function approximation. Another exciting direction is federated RL with a small subset of adversarial workers, where our recent contributions \cite{ganesh2024global} and related work are relevant.

\bibliographystyle{ieeetr}
\bibliography{references}

\end{document}